\documentclass{article} 
\usepackage{iclr2026_conference,times}

\usepackage{amsmath,amsfonts,bm}

\def\eqref#1{equation~\ref{#1}}

\def\1{\bm{1}}

\DeclareMathAlphabet{\mathsfit}{\encodingdefault}{\sfdefault}{m}{sl}
\SetMathAlphabet{\mathsfit}{bold}{\encodingdefault}{\sfdefault}{bx}{n}

\let\ab\allowbreak

\usepackage{hyperref}
\usepackage{url}

\usepackage{booktabs}
\usepackage{multirow}
\usepackage{makecell}

\usepackage{times}
\usepackage{latexsym}
\usepackage[T1]{fontenc}

\usepackage[table]{xcolor} 
\PassOptionsToPackage{table}{xcolor}

\usepackage[utf8]{inputenc}

\usepackage{microtype}

\usepackage{inconsolata}
\usepackage{graphicx}
\usepackage{subcaption}

\usepackage{amsmath}
\usepackage{bbm}
\usepackage{amssymb}

\usepackage{booktabs} 
\usepackage{multirow} 

\usepackage{array}

\usepackage{hyphenat}     
\usepackage[most]{tcolorbox}

\usepackage{rotating}

\usepackage{pgfplots}
\usepackage{pgfplotstable}
\usepgfplotslibrary{groupplots}
\usepgfplotslibrary{external}
\usepgfplotslibrary{colorbrewer}
\pgfplotsset{compat=newest}

\definecolor{rowgray}{gray}{0.90}   
\newcommand{\ie}{\textit{i.e.}}
\newcommand{\eg}{\textit{e.g.}}

\newcommand{\curse}{\textit{curse of multilinguality}}

\newcommand{\finewebtwo}{FineWeb-2}

\definecolor{darkgreen2}{HTML}{009B55}
\definecolor{myblue2}{HTML}{29abe2}
\definecolor{myorange2}{HTML}{f7931e}

\definecolor{mypurple2}{HTML}{9823FF}

\newif\ifcomments
\commentstrue

\ifcomments
    \usepackage[normalem]{ulem}
    \definecolor{ABpurple}{rgb}{0.8,0.0,0.8}
    \newcommand\abi[1]{\textcolor{ABpurple}{#1}} 
    \newcommand\abm[1]{{\marginparwidth=2cm \marginpar{\raggedright\tiny\textcolor{ABpurple}{\textsf{{\bfseries AB\@:} #1}}}}} 
    \newcommand\abs{\bgroup\markoverwith{\textcolor{ABpurple}{\rule[.4ex]{2pt}{0.8pt}}}\ULon} 
\else
    \newcommand\ab[1]{}
    \newcommand\abi[1]{\ignorespaces}
    \newcommand\abm[1]{}
    \newcommand\abs[1]{#1}
\fi

\definecolor{myred}{HTML}{E11D3F}
\definecolor{myorange}{HTML}{F04A00}
\definecolor{myblue}{HTML}{1974D2}
\definecolor{mypurple}{HTML}{6A0DAD}
\ifcomments

    \newcommand{\todo}[1]{{\scriptsize\color{blue!80!black}[{\bf TODO: }\textsf{#1}]}}

    \newcommand{\nf}[1]{\textcolor{myorange}{\textsf{\scriptsize[\textbf{NF\@:} #1]}}}
\else
    \excludecomment{todoenv}
    \excludecomment{noteenv}
    \newcommand{\todo}[1]{}
    \newcommand{\nf}[1]{}
\fi

\def\fixedtotalbudget{\emph{Fixed Total Budget}}
\def\fixedmultilingualbudget{\emph{Fixed Multilingual Budget}}

\newcounter{resultctr}
\renewcommand{\theresultctr}{\arabic{resultctr}}

\newenvironment{assumptionbox}[1][] 
{
  \refstepcounter{resultctr}
  \tcolorbox[
    colback=white,
    colframe=black!75!white,
    sharp corners,
    boxrule=0.5pt,
    drop shadow,
    enhanced,
    #1
  ]
  \textbf{Assumption \theresultctr:}
}
{\endtcolorbox}

\title{Revisiting Multilingual Data Mixtures \\ in Language Model Pretraining}

\author{Negar Foroutan\thanks{Equal contribution}\:\:,
Paul Teiletche\footnotemark[1]\:\:,
Ayush Kumar Tarun,
Antoine Bosselut\\
EPFL, Lausanne, Switzerland\\
\small \noindent \textbf{Correspondence:} \href{mailto:negar.foroutan@epfl.ch}{\{negar.foroutan, antoine.bosselut\}@epfl.ch}\\
}

\iclrfinalcopy 
\begin{document}

\maketitle

\begin{abstract}
The impact of different multilingual data mixtures in pretraining large language models (LLMs) has been a topic of ongoing debate, often raising concerns about potential trade-offs between language coverage and model performance (\ie, the \curse{}).
In this work, we investigate these assumptions by training $1.1$B and $3$B parameter LLMs on diverse multilingual corpora, varying the number of languages from 25 to 400. Our study challenges common beliefs surrounding multilingual training.
First, we find that combining English and multilingual data does not necessarily degrade the in-language performance of either group, provided that languages have a sufficient number of tokens included in the pretraining corpus.
Second, we observe that using English as a pivot language (\ie, a high-resource language that serves as a catalyst for multilingual generalization) yields benefits across language families, and contrary to expectations, selecting a pivot language from within a specific family does not consistently improve performance for languages within that family. Lastly, we do not observe a significant ``\curse{}'' as the number of training languages increases in models at this scale.
Our findings suggest that multilingual data, when balanced appropriately, can enhance language model capabilities without compromising performance, even in low-resource settings.
\end{abstract}
\section{Introduction}
\label{introduction}

Recent advances in large language models (LLMs) have demonstrated impressive performance across a wide range of non-English languages, including many that are considered low-resource \citep{yang2025qwen3, gemmateam2025gemma3technicalreport, grattafiori2024llama3herdmodels, ustun-etal-2024-aya, achiam2023gpt}. These models are typically pretrained on data from over 100 high- and mid-resource languages, leveraging the broad availability of multilingual content on the web. Despite this progress, the impact of multilingual data composition on model training remains a subject of active debate, particularly regarding potential trade-offs between total language coverage and model performance in different languages \citep{alastruey2025interference}.
Practitioners often face difficult trade-offs: Should they include more languages in the pretraining data mixture or concentrate resources to prioritize performance in fewer languages? For greater multilingual generalization, should they include pivot languages from different language families or merely from high-resource global languages? Could curriculum learning among pivot languages also lead to greater multilingual generalization? 

While previous studies tried to address these questions, they have generally been limited in scope, either by the number of languages considered or by the scale of the models used. For instance, one study investigates the so-called \curse{} using relatively small models with $45$M parameters~\citep{chang-etal-2024-multilinguality}. Another recent work explores scaling laws for multilingual language models and proposes an optimal sampling ratio for multilingual data \citep{he2024scalinglawsmultilinguallanguage}. However, this work focuses on only 23 languages and similarly small models (85M parameters). Other studies have discussed multilingual data mixtures for task training \citep{wang-etal-2020-negative} or instruction-tuning \citep{ustun-etal-2024-aya}, but it is unknown to what extent their intuitions would extend to pretraining.

In this work, we study the impact of multilingual data composition in training large-scale LLMs. 
Specifically, we train a series of $1.1$B and $3$B parameter models on corpora of $100$B tokens containing up to 400 languages, allowing us to systematically explore the effects of language count, diversity, and token distribution. 
Our experiments challenge several prevailing propositions about multilingual training. We summarize our key findings as follows:

\textbf{Findings \#1: More English data does not necessarily hurt multilingual performance.}
We show that varying the proportion and absolute amount of English data in the training mix does not harm multilingual performance, as long as a sufficient number of multilingual tokens are included in the pretraining mixture. The reverse is also true, as increasing the number of multilingual tokens does not harm English performance as long as there are sufficient English tokens in the pretraining mixture.

\textbf{Findings \#2: Language family boundaries are not barriers to transfer.}
Contrary to the prevailing wisdom that family-specific pivots are most effective for cross-lingual transfer~\citep{he2024scalinglawsmultilinguallanguage}, we find that using English as a pivot language\footnote{Historically, \textit{pivot} languages are used as intermediary languages for \textit{many-to-many} translation. In the context of this work we refer to pivot languages as those that are highly represented in pretraining data and whose presence serves as a catalyst for multilingual generalization.}
provides benefits across language families. Selecting a high-resource pivot language from within a specific family (\eg, Russian for Slavic languages) does not consistently enhance performance across languages in that family.
Given that English has the most diverse and highest-quality data on the web, this evidence shows the potential for leveraging a high-resource language to improve performance in other languages, regardless of their family.

\textbf{Findings \#3: Curriculum learning fails to mitigate negative interference.}
Prior work has shown training on multiple languages simultaneously can degrade performance in both high- and low-resource languages, a phenomenon coined as negative interference \citep{wang-etal-2020-negative}. 
Although curriculum learning has been proposed as a potential solution to this problem~\citep{zhang-etal-2021-competence-based, kumar2021learning, choi2023order}, our results show that staging the introduction of languages during training neither reduces negative interference nor improves performance on non-English languages.

\textbf{Findings \#4: Increasing the number of training languages does not always lead to performance degradation.}
The \curse{} suggests that expanding language coverage
reduces model performance in both monolingual and cross-lingual settings \citep{chang-etal-2024-multilinguality, blevins-etal-2024-breaking, pfeiffer-etal-2022-lifting, conneau-etal-2020-unsupervised}.
We find the \curse{} arises not from simply adding more languages, but from the finite capacity of models and data distributions that amplify the impact of noisy, low-resource languages.

Collectively, our findings offer practical guidance for designing more effective multilingual pretraining strategies and contribute to the development of stronger, more inclusive multilingual LLMs.

\section{Experimental Setup}
\label{experimental_setup}

\textbf{Model.} We train decoder-only Transformer models~\citep{vaswani2017attention} based on the LLaMA architecture~\citep{touvron2023llama}, in two sizes: $1.1$ and $3$ billion parameters ($1.1$B and $3$B). The model sizes are determined by varying the number of layers, hidden dimensions, and attention heads. Detailed configuration and training parameters are provided in Appendix~\ref{appendix:training_setup}.

\textbf{Pretraining Data.} We use two corpora in our experiments. For experiments involving 30 languages, we use the multilingual version of the C4 corpus (mC4;~\citealp{xue-etal-2021-mt5, 2019t5}).\footnote{https://huggingface.co/datasets/allenai/c4} For experiments involving a larger set of up to 1,834 languages, we use the FineWeb2 corpus~\citep{penedo2024fineweb-2}. All data are tokenized using the \texttt{Mistral-Nemo-Base-2407} tokenizer,\footnote{https://huggingface.co/mistralai/Mistral-Nemo-Base-2407} which has a vocabulary size of $|\mathcal{V}| = 131{,}000$ tokens. Models are trained on $D = 100$ to $D = 225$ billion tokens.

\textbf{Evaluation.} We evaluate our models by measuring their language modeling loss on a held-out validation set that is distinct from the pretraining data. In addition, we perform \textit{downstream task evaluations} using a suite of multilingual benchmarks. For each model, we aggregate results by language to obtain a comprehensive score for every model-language pair. Details of the benchmark suite and the aggregation procedure are provided in Appendix~\ref{appendix:benchmark_setup}.

\begin{figure*}[t]
    \centering
    \begin{subfigure}[b]{0.49\textwidth}
        \includegraphics[width=\textwidth]{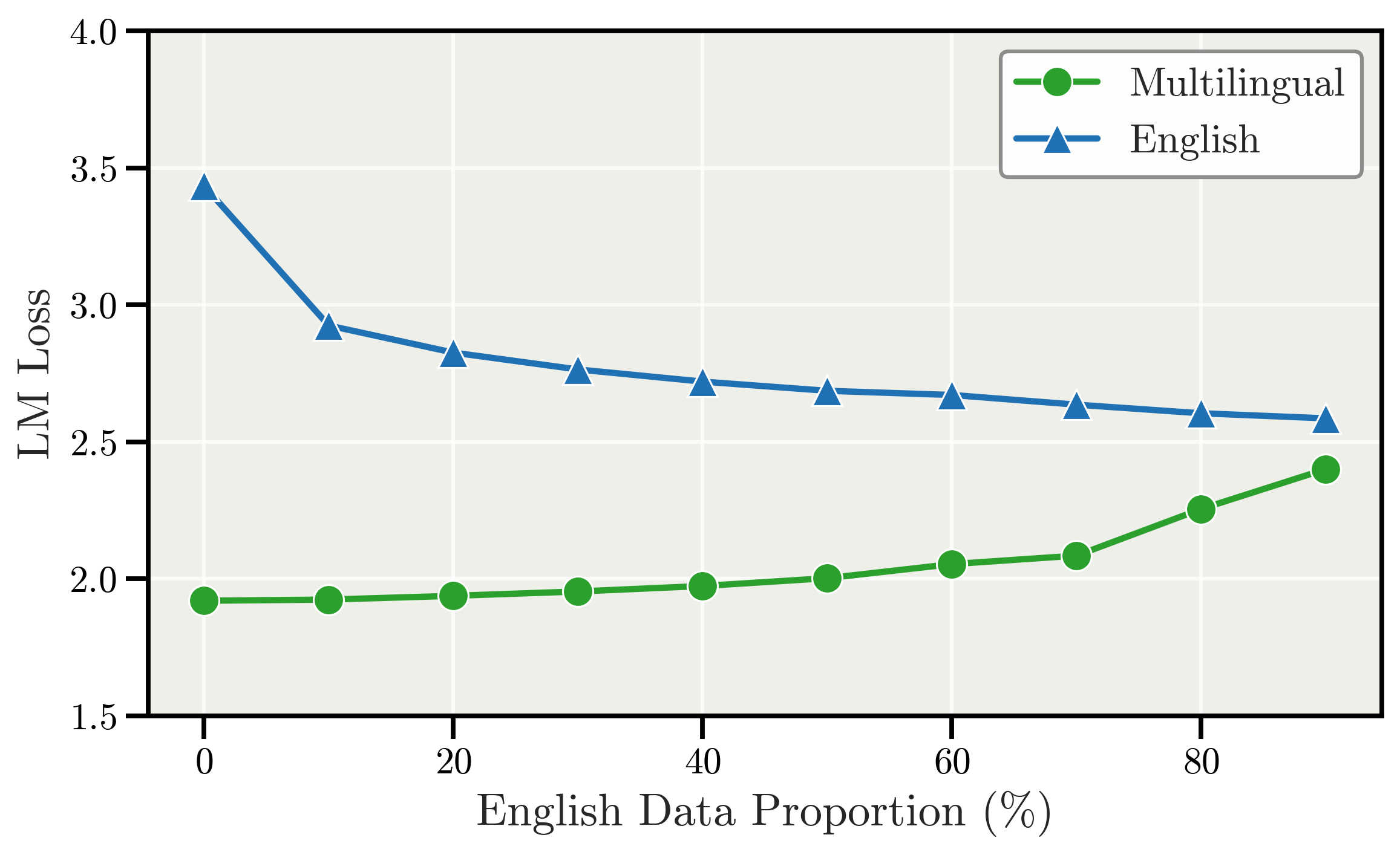}
        \caption{Fixed Total Budget}
        \label{fig:fixed_total}
    \end{subfigure}
    \begin{subfigure}[b]{0.49\textwidth}
        \includegraphics[width=\textwidth]{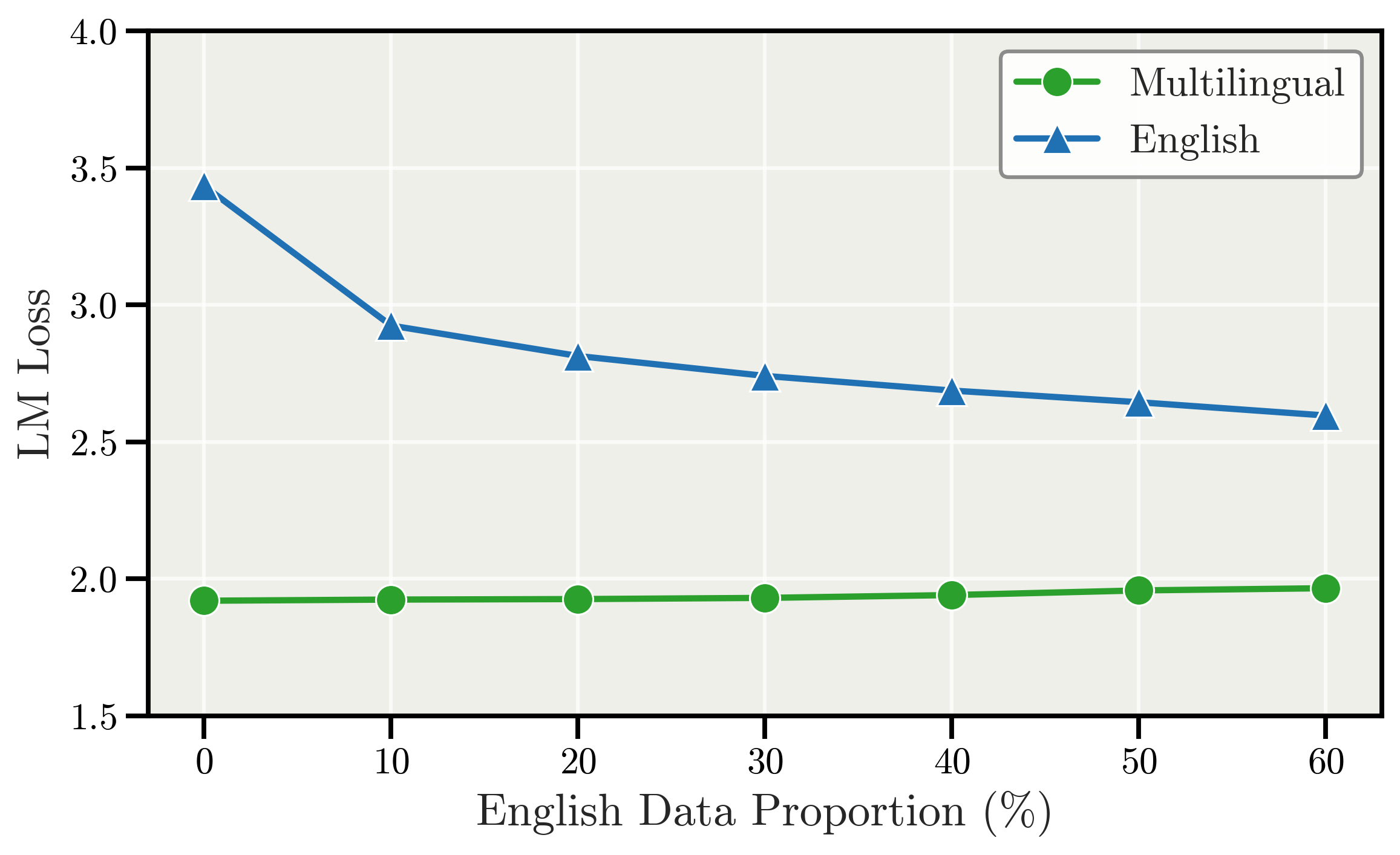}
        
        \caption{Fixed Multilingual Budget}
        \label{fig:fixed_multilingual}
    \end{subfigure}
    \caption{Validation {LM loss} for \textbf{English} and weighted average {LM loss} of non-English languages (\textbf{Multilingual}) across different proportions of English in the pretraining data for \textbf{$1.1$B} models. \textbf{(a)} In a \textbf{Fixed Total Budget}, increasing English data ($\geq$50\%) leads to a performance drop in other languages.  \textbf{(b)} In a \textbf{Fixed Multilingual Budget}, increasing English data (up to 60\%) does not have a negative effect on other languages.}
    \label{fig:multilingual_effect}
\end{figure*}
\begin{figure*}[t]
    \centering
    \begin{subfigure}[b]{0.49\textwidth}
        \includegraphics[width=\textwidth]{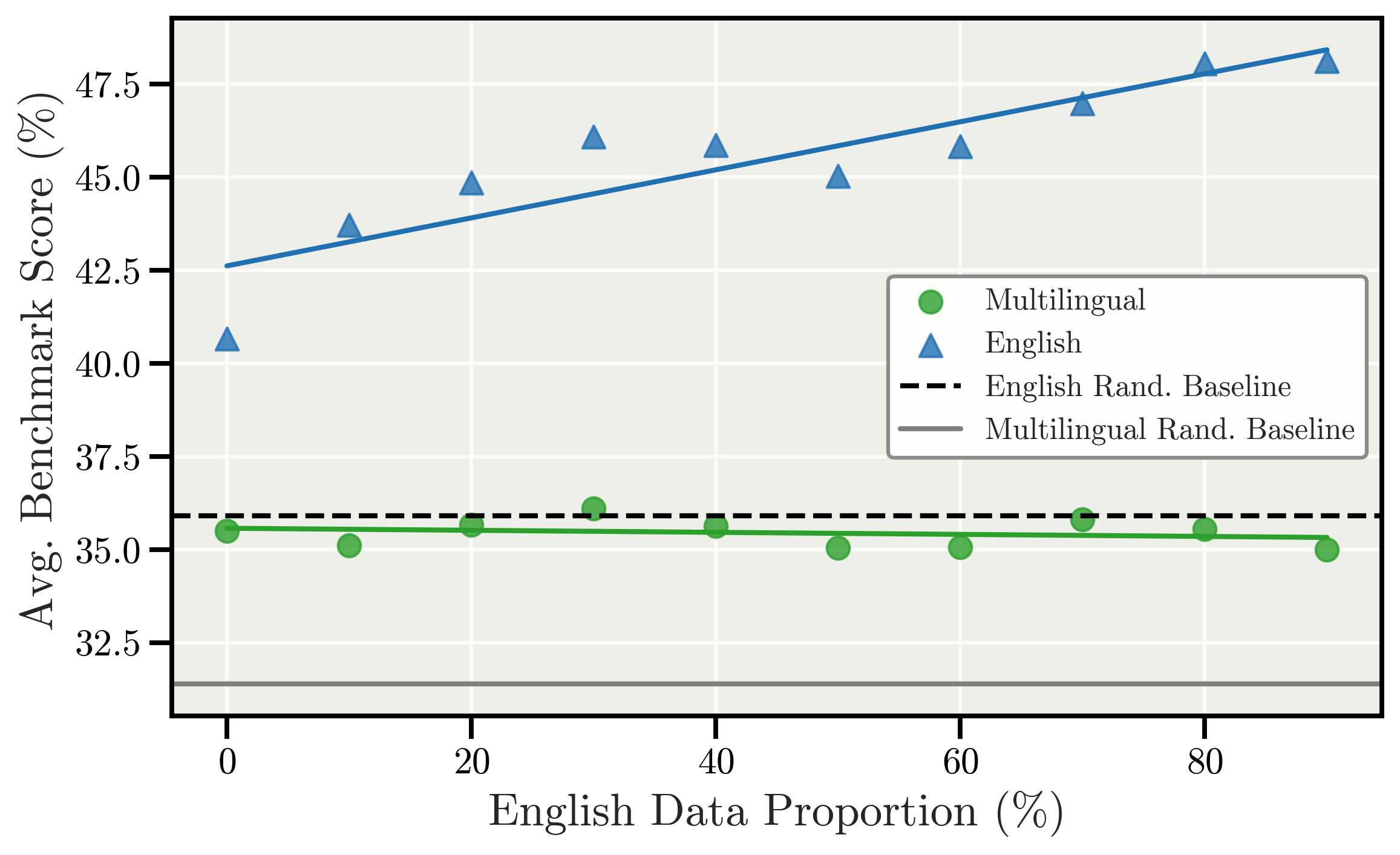}
        \caption{Fixed Total Budget}
        \label{fig:fixed_total_benchmark_1.1B}
    \end{subfigure}
    \begin{subfigure}[b]{0.49\textwidth}
        \includegraphics[width=\textwidth]{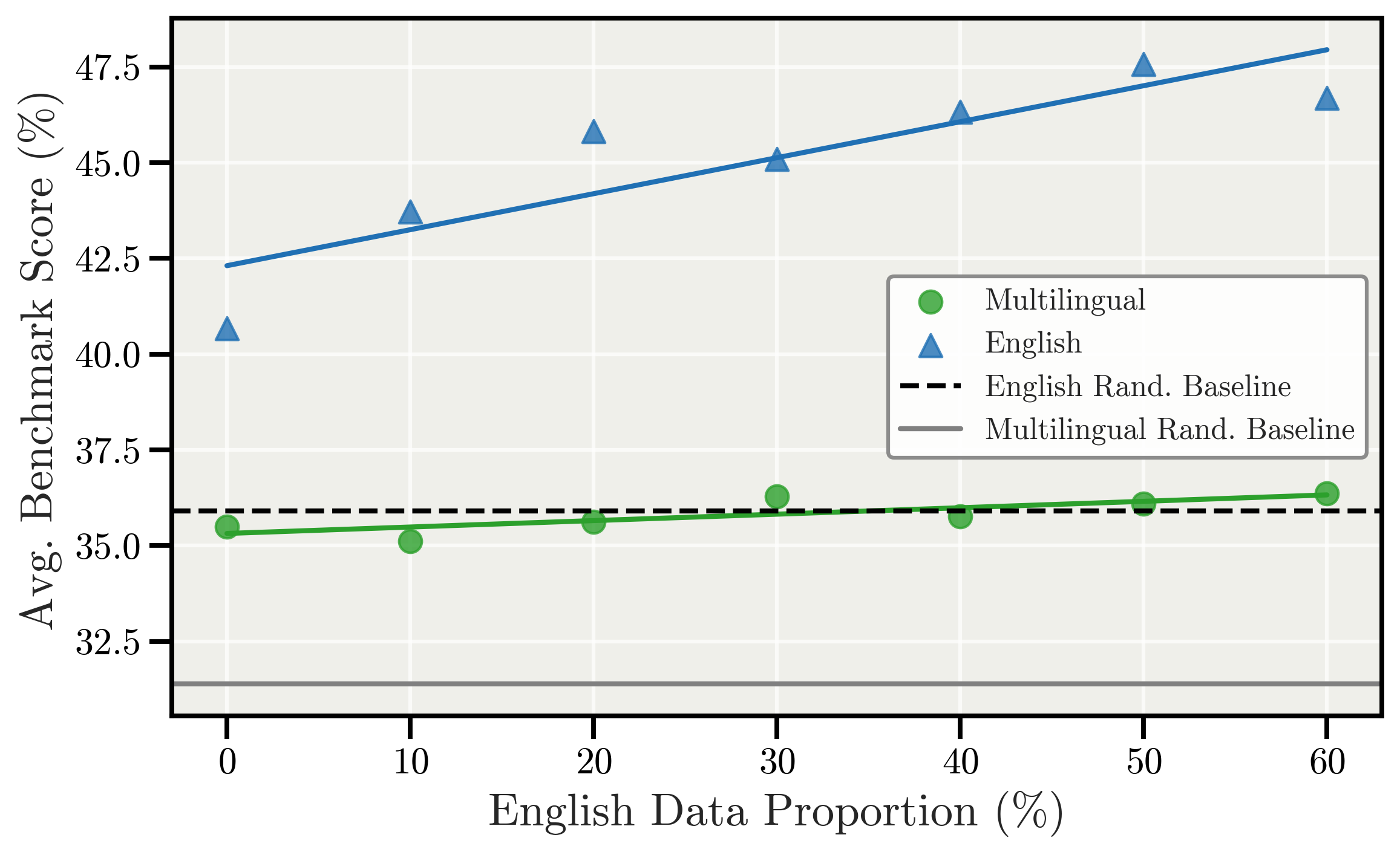}
        \caption{Fixed Multilingual Budget}
        \label{fig:fixed_multilingual_benchmark_1.1B}
    \end{subfigure}
    \caption{Aggregated \textit{benchmark performance} for \textbf{English} and weighted average of non-English (\textbf{Multilingual}) across different proportions of English in the training data for \textbf{$1.1$B} models. 
    The dashed lines represent the random baselines for each language group. 
    \textbf{(a)} In a \textbf{Fixed Total Budget}, increasing English data ($\geq$50\%), does not hurt downstream performance on the \textbf{Multilingual} group. \textbf{(b)} In a \textbf{Fixed Multilingual Budget}, we see that increasing English data has a negligible impact on the \textbf{Multilingual} group's performance.}
    \label{fig:multilingual_effect_benchmark_1.1B}
\end{figure*}

\section{Assumption \#1: English Hurts Multilinguality}
\label{sec:english}

English serves as the dominant pivot language for LLMs due to the abundance, diversity, and quality of English data available on the web.
Simultaneously, due to the prevalence of LLM applications in English, maintaining English performance is often prioritized when training multilingual models by increasing the total proportion of English data, potentially at the expense of multilingual performance.

\begin{assumptionbox}[label={assumption1}]
More English data comes at the cost of performance in other languages.
\end{assumptionbox}

In this experiment, we investigate how the amount of English pretraining data influences performance in non-English languages. We train models of $1.1$B and 3B parameters using data in 30 languages from the mC4 corpus, systematically varying the proportion of English data from 0\% to 100\%.
The selected languages represent diverse language families and data resource levels (Table~\ref{tab:en_pivot_lang_metadata}). We use temperature sampling with $\tau = 3.3$ (details in Appendix~\ref{appendix:sampling_methods}).
When deciding on the data budget for these experiments, we consider two settings to disentangle the impact of data composition from the total amount of data seen during training:

\fixedtotalbudget{}: The total pretraining budget is held constant at \underline{100B} tokens. Increasing the proportion of English reduces the amount of non-English (multilingual) data. This setup explores the trade-off between English and multilingual data under a constrained data regime.
    
\fixedmultilingualbudget{}: The amount of non-English data is fixed at \underline{90B} tokens with English data added on top, leading to a growing total data size (up to $225$B tokens). 
This setup explores the effect of increasing English data without reducing multilingual coverage, simulating an unconstrained data regime (where multilingual data may be available in smaller quantities in web data than English data).

\paragraph{Results.} Figure~\ref{fig:fixed_total} shows the final validation loss for English and non-English languages for the \textbf{$1.1$B} model for the \fixedtotalbudget{} setting. As expected, increasing the proportion of English data leads to a lower validation loss for English. For non-English languages, validation loss remains relatively stable up to approximately 40\% English data. Beyond this point, performance begins to degrade, indicating that allocating more capacity to English at the expense of other languages negatively impacts multilingual learning.

In contrast, under the \fixedmultilingualbudget{} setting (Figure~\ref{fig:fixed_multilingual}), we observe that multilingual performance remains largely unaffected—even when English comprises up to 60\% of the dataset. These results suggest that, provided there is sufficient data to support learning robust multilingual representations, adding more English data does not interfere with performance on other languages.
A similar pattern holds for the 3B models, as shown by the results in Appendix Figure~\ref{fig:multilingual_effect_3B}.

Figure~\ref{fig:fixed_multilingual_benchmark_1.1B} presents the benchmark results for this experiment. In both the \fixedtotalbudget{} and \fixedmultilingualbudget{} settings, we observe that increasing the proportion of English data consistently improves downstream task a in English. Mirroring the same patterns as for the loss, this increase does not degrade performance on other languages.
Furthermore, Figure~\ref{fig:multilingual_effect_benchmark_3B} shows results for the 3B models (see Appendix~\ref{appendix:pivot_ablation}), which exhibit a similar trend.

\textbf{Takeaway:} Contrary to common belief, increasing the amount of English data in the training of LLMs does not necessarily degrade their multilingual capabilities, provided that the training set also contains a sufficient quantity of multilingual tokens. In other words, it is possible to support additional languages while still maintaining strong performance in English.

\section{Assumption 2: ``Stay in the Family''} 
\label{sec:family}

Previous research suggests that cross-lingual transfer is generally more effective between languages that belong to the same language family \citep{he2024scalinglawsmultilinguallanguage, bagheri-nezhad-agrawal-2024-drives}. This implies that, if the pattern holds consistently, selecting a pivot language from within the same family is likely to yield greater transfer benefits than choosing one from a different family.

\begin{assumptionbox}[label={assumption2}]
Languages within the same family offer the strongest boost to multilingual generalization.
\end{assumptionbox}

\begin{figure*}[t]
    \centering
    \begin{subfigure}[b]{0.49\textwidth}
        \includegraphics[width=\textwidth]{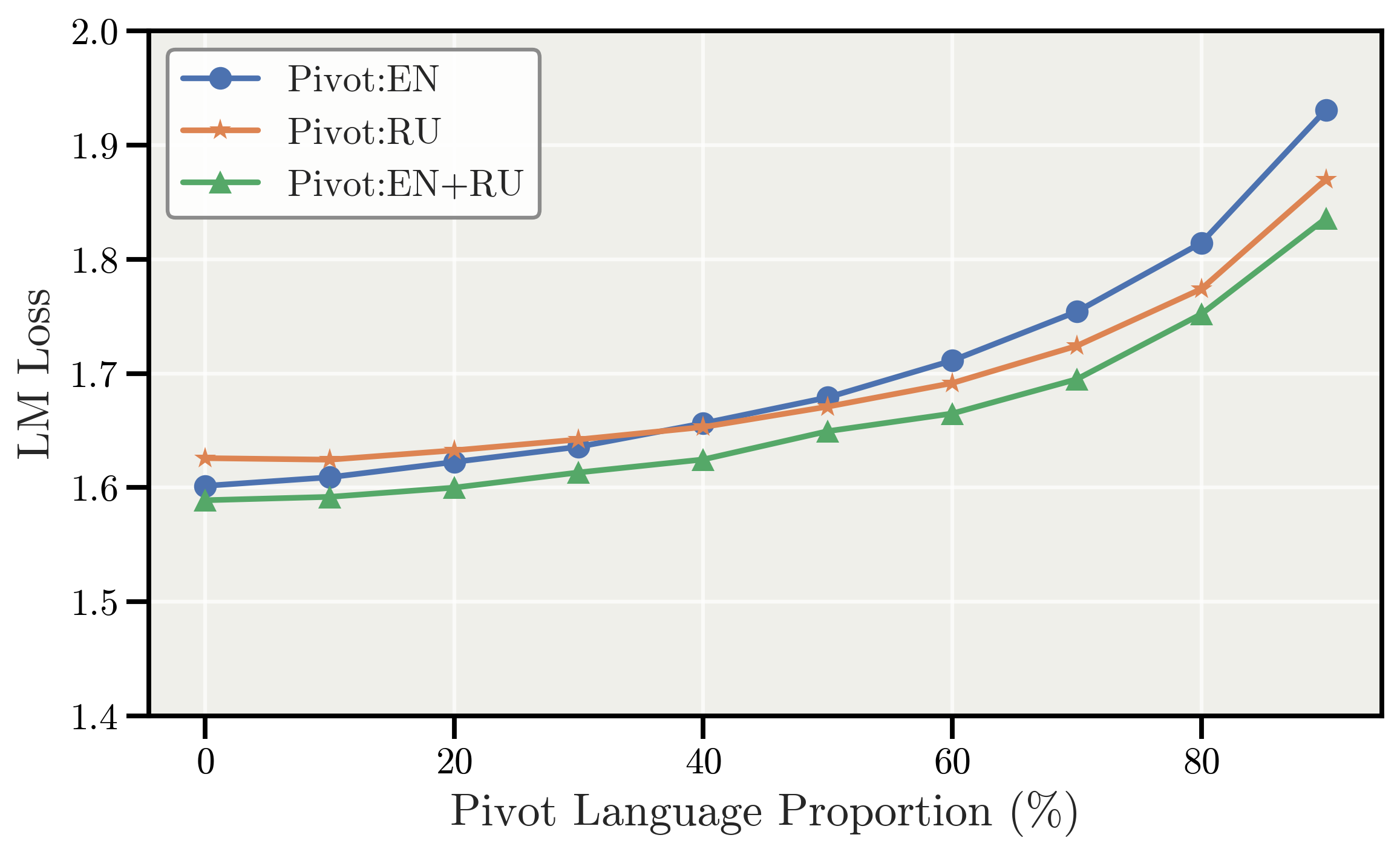}
        \caption{Slavic Languages}
    \end{subfigure}
    \begin{subfigure}[b]{0.49\textwidth}
        \includegraphics[width=\textwidth]{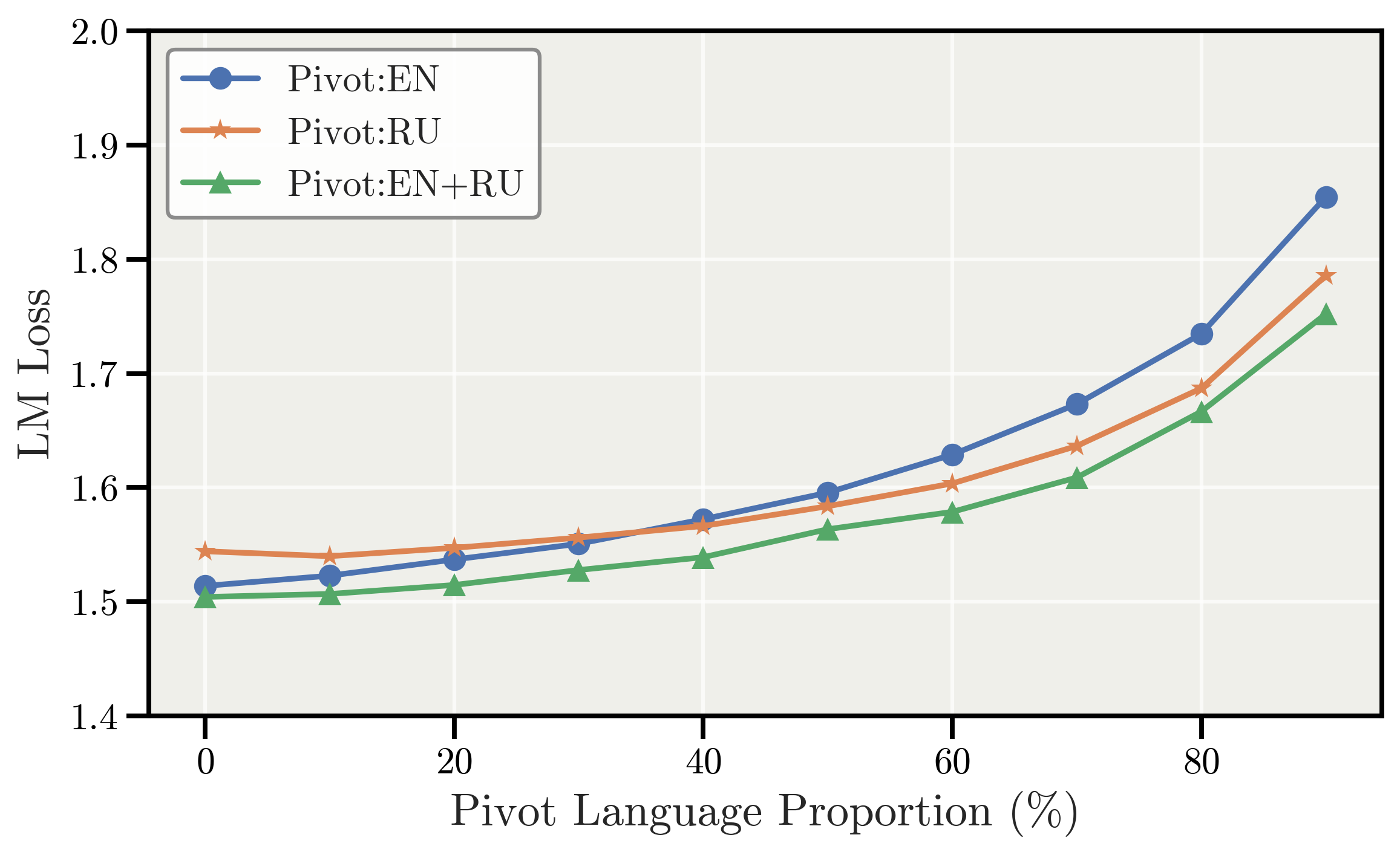}
        \caption{Cyrillic-Script Languages}
    \end{subfigure}
    \caption{Weighted average of validation LM loss for (a) \textit{Slavic} and (b) \textit{Cyrillic-script} languages when we have English, Russian, or English+Russian as a pivot language in the training data mix. Having a combination of Russian and English as pivots leads to the best performance for both groups of languages (Model size = $1.1$B).}
    \label{fig:pivot_ablation}
\end{figure*}

In this experiment, we investigate the impact of using various types of pivot languages in a training corpus with multiple language families. A pivot language is defined as an intermediary language in a pretraining set for more effectively learning languages with less available data.

We compare using English as a pivot language for all languages, and selecting a pivot language from within the same language family for certain languages. Specifically, we train a $1.1$B model on a subset of \textit{Slavic} and \textit{Cyrillic-script} languages under three different conditions: 
(1) English as the pivot language, 
(2) Russian as the pivot language, and 
(3) a uniform combination of English and Russian as pivots. 
The \textit{Slavic} set includes Belarusian, Ukrainian, Macedonian, Bulgarian, Mongolian, Serbian, Polish, Czech, and Slovak. The \textit{Cyrillic-script} set comprises Belarusian, Ukrainian, Macedonian, Bulgarian, Kyrgyz, Tajik, Kazakh, Mongolian, Serbian, and Uzbek (see Table~\ref{tab:cyrillic_pivot_lang_metadata} for details).

\paragraph{Results.} Figure~\ref{fig:pivot_ablation} presents the weighted average loss across both language groups.
We observe that as the proportion of training data assigned to the pivot language increases (and the complement proportion for non-pivot languages decreases), the loss for non-pivot languages remains relatively stable at first. However, as less data is allocated to them, their loss eventually rises, as expected.
Up to a 50\% allocation to the pivot language, English and Russian perform comparably. However, beyond this threshold---particularly at 60\% or more, Russian proves slightly more effective as a pivot, yielding lower loss for the remaining languages.
One possible explanation is that when pivot allocation is relatively low, non-pivot languages still benefit from having access to their own training data. But in extremely low-resource conditions, these languages gain more from leveraging similarities with a strong pivot language.
Another factor is that English training data is often more diverse and standardized, with broad domain coverage. This richness may make English a strong pivot up to a certain point, after which typological proximity favors Russian.
Notably, combining \textit{both} English and Russian as joint pivots yields the lowest overall loss, suggesting a complementary effect: English contributes wide coverage, while Russian offers closer linguistic ties to many of the target languages.
The detailed per-language loss values are provided in Figure~\ref{fig:pivot_ablation_per_lang_loss} in Appendix~\ref{appendix:pivot_ablation}.

\textbf{Takeaway:} English can serve as a broadly effective pivot language, but in very low-resource settings, typological similarity becomes increasingly important. Using multiple pivots that balance breadth and proximity provides the most consistent benefits across language families.

\section{Assumption 3: Multilingual Curriculum Learning Reduces Negative Interference}
\label{sec:curriculm}

Previous work suggests that the order in which languages are introduced during training can influence model performance and potentially reduce competition between languages~\citep{choi2023order, ranaldi-etal-2024-language}.

\begin{assumptionbox}[label={assumption3}]
Curriculum-based language introduction mitigates negative interference.
\end{assumptionbox}

To investigate the dynamics of cross-lingual competition and knowledge transfer in multilingual language models, we designed a series of controlled \textit{curriculum learning} experiments. Our goal is to understand how the timing and order of language inclusion during training influence model performance.
We design four experimental setups:

    \textit{All-at-once (baseline):} The model is trained on the full multilingual dataset from the outset. This setup, common in many multilingual LLMs (\eg, \citealp{swissai2025apertus}) serves as a control to benchmark the effects of curriculum-based training strategies.
    
    \textit{English-all}: For the first 25\% of training, the model is exposed only to English. After this phase, training proceeds on the full multilingual dataset. This allows us to isolate the impact of early single-language pretraining on subsequent multilingual generalization and interference.
    
    \textit{English-Pivots-all}: Training is divided into three phases
    (1) 0–25\%: Only English data is used.
    (2) 25–50\%: We introduce three additional high---resource languages---Arabic, Chinese, and Russian---alongside English as pivot languages. These four languages were chosen to represent four distinct scripts: Latin, Arabic, Han, and Cyrillic, respectively. This intermediate stage allows us to explore early competition between strong languages with differing orthographic and typological properties.
    (3) 50–100\%: The model is trained on the full multilingual dataset.
    This progressive inclusion strategy enables a controlled examination of cross-lingual interactions and competition under varying degrees of language diversity. 
    
    \textit{Pivots-all}:
    For the first 25\% of training, the model is trained using our 4 pivot languages. After this phase, training continues on the full multilingual dataset. This allows us to isolate the impact of early high-resource pretraining on subsequent multilingual generalization and interference.

\begin{figure}[t]
    \centering
    \begin{subfigure}[t]{0.32\linewidth}
        \includegraphics[width=\textwidth]{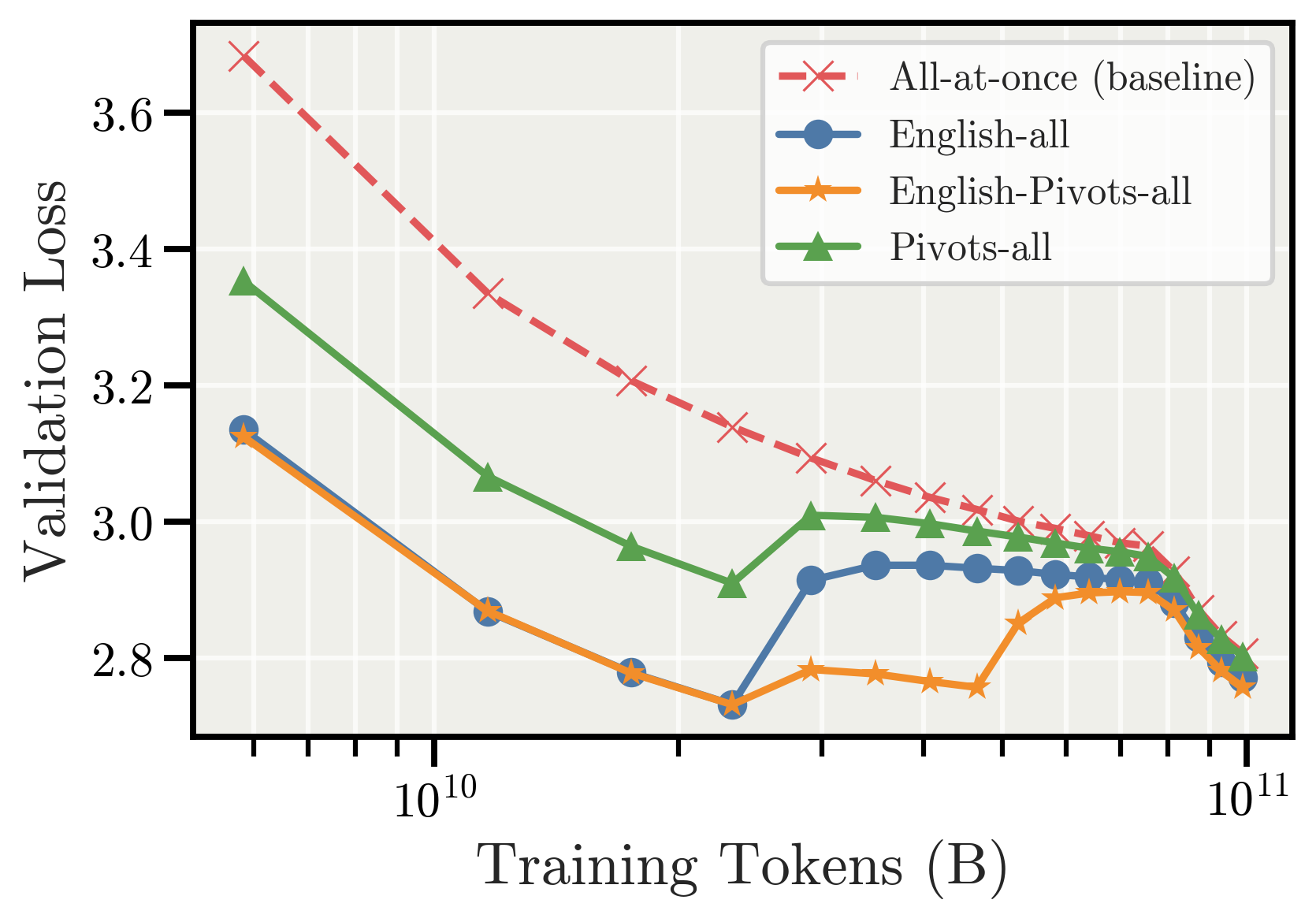}
        \caption{Loss on English-language validation data.}
        \label{fig:curriculum_english}
    \end{subfigure}
    \hfill 
        \begin{subfigure}[t]{0.31\linewidth}
        \includegraphics[width=\textwidth]{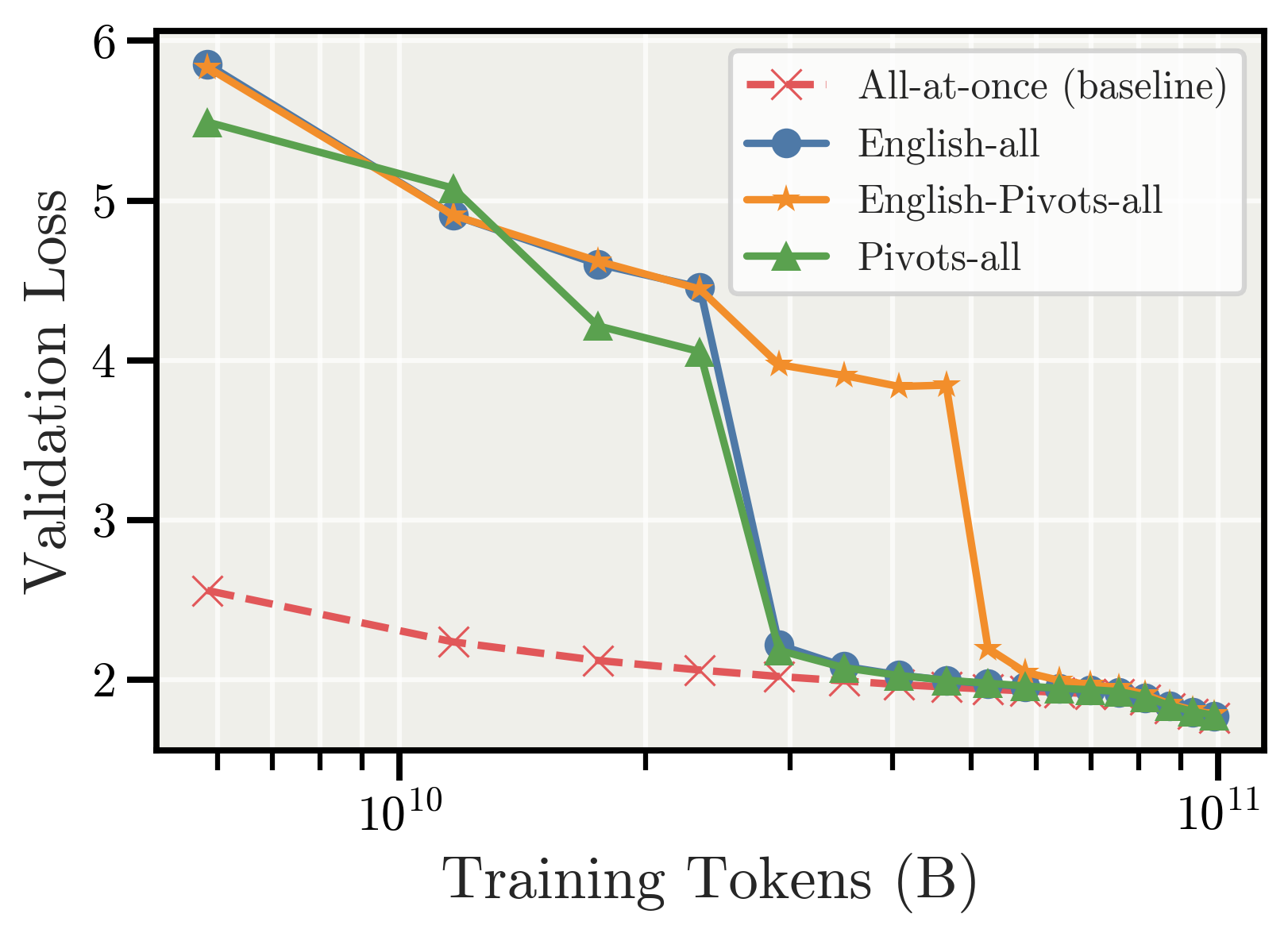}
        \caption{Weighted average loss for non-pivot languages.}
        \label{fig:curriculum_others}
    \end{subfigure}
    \hfill
    \begin{subfigure}[t]{0.31\linewidth}
        \includegraphics[width=\textwidth]{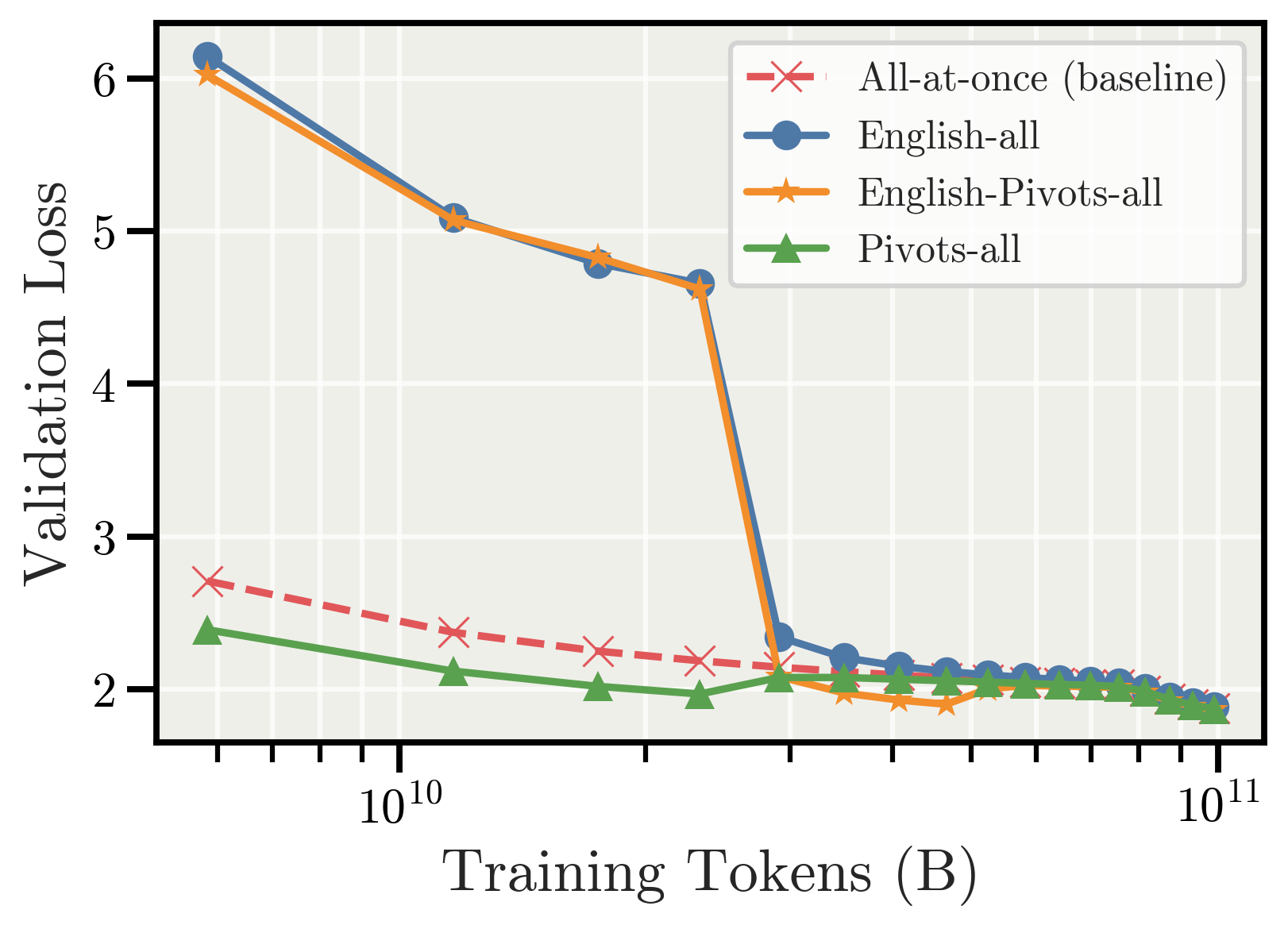}
        \caption{Weighted average Loss for Arabic, Chinese, and Russian.}
        \label{fig:curriculum_pivots}
    \end{subfigure}
    \caption{LM loss on the validation set for $3$B models as a function of consumed training tokens, shown separately for (a) English, (b) non-English, and (c) pivot languages under different curriculum strategies.}
    \label{fig:curriculum}
\end{figure}
\paragraph{Results.} Figure~\ref{fig:curriculum} presents the results of our curriculum learning experiments.
When examining English loss, we find that introducing English early in training---either alone (\textit{English-all}) or alongside pivot languages (\textit{English-pivots-all})---leads to lower final loss for English.
Notably, transitioning between curriculum stages (\ie, adding new languages in successive phases) temporarily increases the loss for previously seen languages. This suggests a short-term ``forgetting'' effect, where the model learns new languages at the cost of temporarily degrading performance on earlier ones, before eventually recovering and integrating all knowledge across the languages.

For the other three pivot languages (Figure~\ref{fig:curriculum_pivots}), the curriculum that begins with English and subsequently introduces the pivots (\textit{Pivots-all}) achieves the lowest average loss midway through training. However, as additional languages are introduced, the loss increases, ultimately converging to the same level as other runs. As with English, we observe a forgetting effect at each transition. 

When analyzing the average loss across other non-English languages (Figure~\ref{fig:curriculum_others}), we observe that while different curriculum regimes begin at different starting points and follow distinct learning trajectories, they all converge to a similar final loss by the end of training. This consistency indicates that curriculum order primarily affects learning dynamics, but not final multilingual performance.

Although curriculum learning appears to benefit English, further analysis reveals that this improvement is largely attributable to data quantity. Specifically, we find a strong correlation between the number of English tokens in the training mix and the model’s performance on English. In other words, models exposed to more English data achieve lower loss. Consequently, the \textit{English-pivots-all} setup attains the lowest English loss primarily because it includes the largest proportion of English data in its curriculum.

\textbf{Takeaway}: 
Curriculum learning shapes the trajectory of multilingual training but does not reduce interference or improve final performance. Observed gains for English under certain curricula are explained by the data distribution rather than curriculum structure.

\section{Assumption 4: The ``Curse of Multiliguality''}
\label{sec:com}
Prior work has shown that, for a fixed model capacity, adding more languages during pretraining initially improves cross-lingual transfer, particularly for low-resource languages.
However, beyond a certain point, both monolingual and cross-lingual performance begin to degrade. This trade-off is commonly referred to as the \curse{} \citep{conneau-etal-2020-unsupervised, pfeiffer-etal-2022-lifting, chang-etal-2024-multilinguality}.

\begin{assumptionbox}[label={assumption4}]
Adding more languages to a pretraining mixture reduces performance.
\end{assumptionbox}

We revisit this assumption by training language models with varying numbers of languages and analyzing the impact on both high- and low-resource languages.

\begin{table}[h]
\centering
\begin{tabular}{ccccc}
\toprule
\multirow{2}{*}{\textbf{\# Languages}} & \multicolumn{2}{c}{\textbf{LM Loss $\downarrow$}} & \multicolumn{2}{c}{\textbf{Benchmark Performance $\uparrow$}} \\
\cmidrule(lr){2-3} \cmidrule(lr){4-5}
 & \textbf{ \scriptsize  Natural Dist.} & \scriptsize  \textbf{Temp. Sampling} &  \scriptsize \textbf{Natural Dist.} &  \scriptsize \textbf{Temp. Sampling}\\
\toprule
25  & 2.678 & 2.675 & 50.13 \scriptsize $\pm$ 1.868 &  43.24 \scriptsize $\pm$ 1.874 \\
50  & 2.678 & 2.681 & 49.41 \scriptsize $\pm$ 1.868 &  43.80 \scriptsize $\pm$ 1.878\\
100 & 2.682 & 2.687 & 49.29 \scriptsize $\pm$ 1.865 &  43.76 \scriptsize $\pm$ 1.872\\
200 & 2.680 & 2.696 & 49.11 \scriptsize $\pm$ 1.864 &  42.38 \scriptsize $\pm$ 1.870\\
400 & 2.678 & 2.707 & 49.64 \scriptsize $\pm$ 1.871 &  42.12 \scriptsize $\pm$ 1.854\\
\bottomrule
\end{tabular}
\caption{English validation LM loss and benchmark performance (\%) when increasing languages coverage from 25 to 400 ($3$B model). English represents 40\% of the training data in all runs ($40$B tokens). Increasing the number of languages, while keeping English data fixed, does not hurt English performance. Per-benchmark values are provided in Table~\ref{tab:english_benchmark}.}
\label{tab:english_com_natural}
\end{table}

\begin{figure}[ht]
    \begin{subfigure}[b]{0.49\linewidth}
        \includegraphics[width=\textwidth]{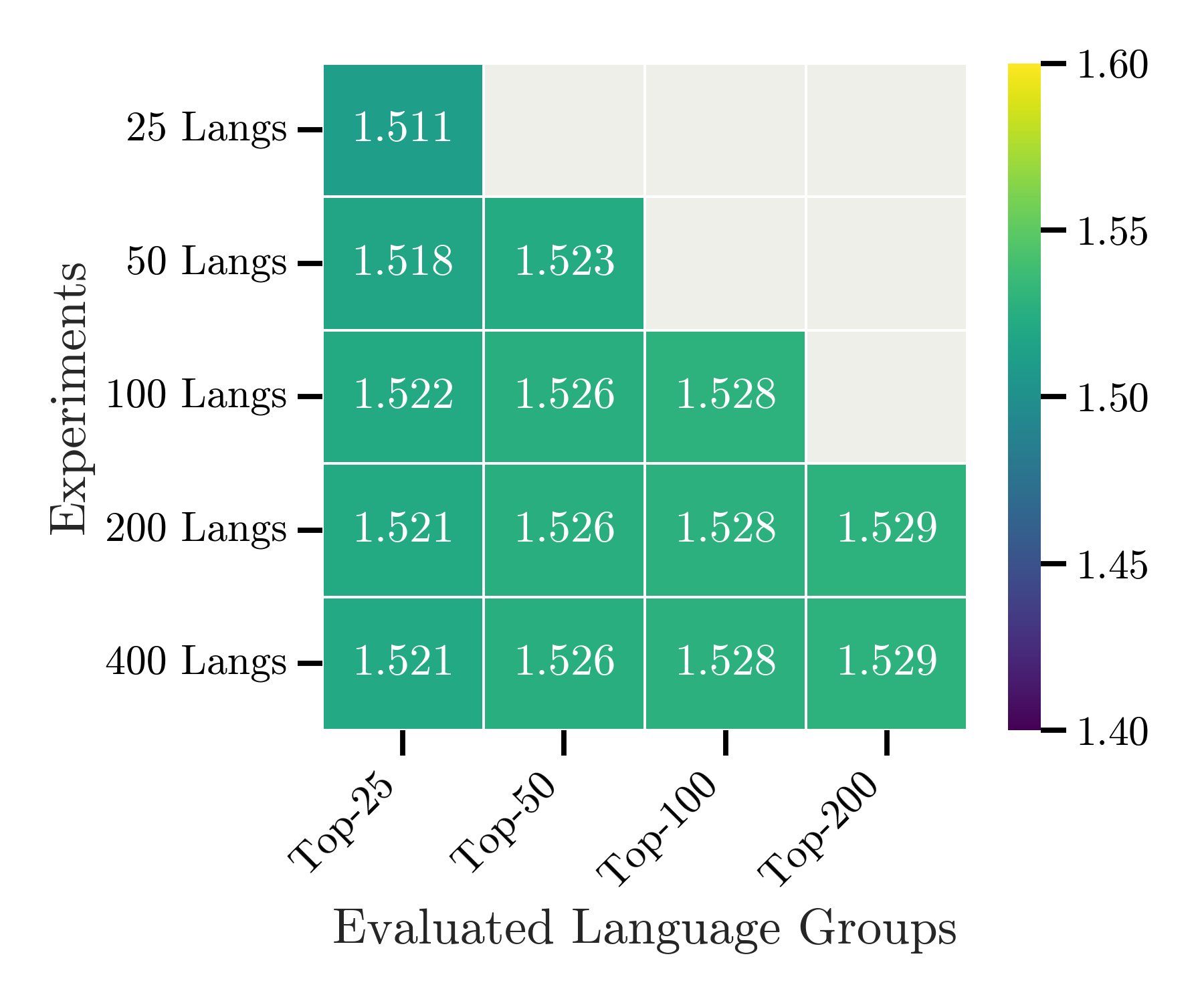}
        \caption{Fixed Total Budget (Natural Distribution)}
        \label{fig:com_avg_natural}
    \end{subfigure}
        \begin{subfigure}[b]{0.49\linewidth}
        \includegraphics[width=\textwidth]{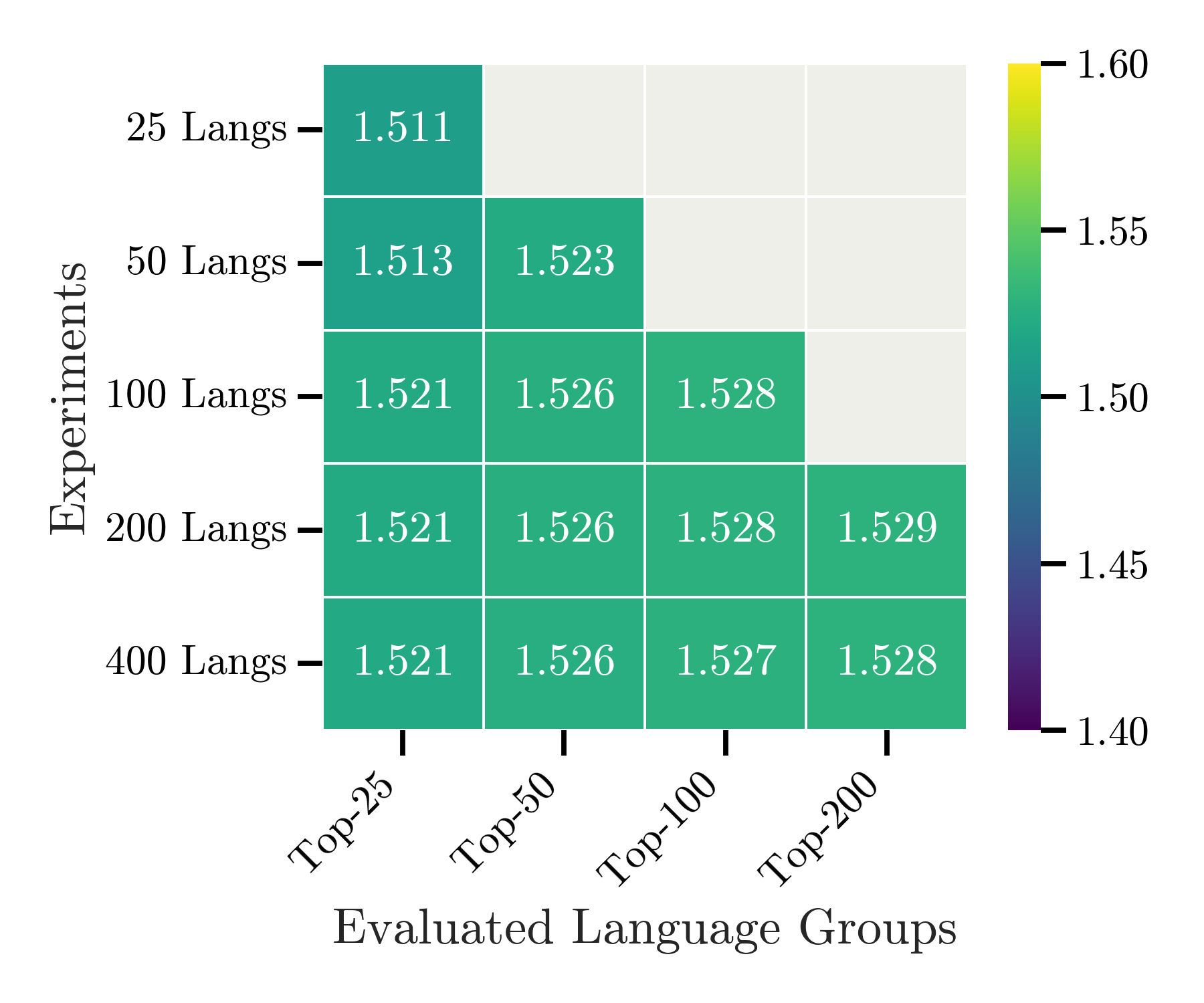}
        \caption{Controlled Growth Setting (Natural Distribution)}
         \label{fig:com_avg_natural_extended}
    \end{subfigure}

\begin{subfigure}[b]{0.49\linewidth}
        \includegraphics[width=\textwidth]{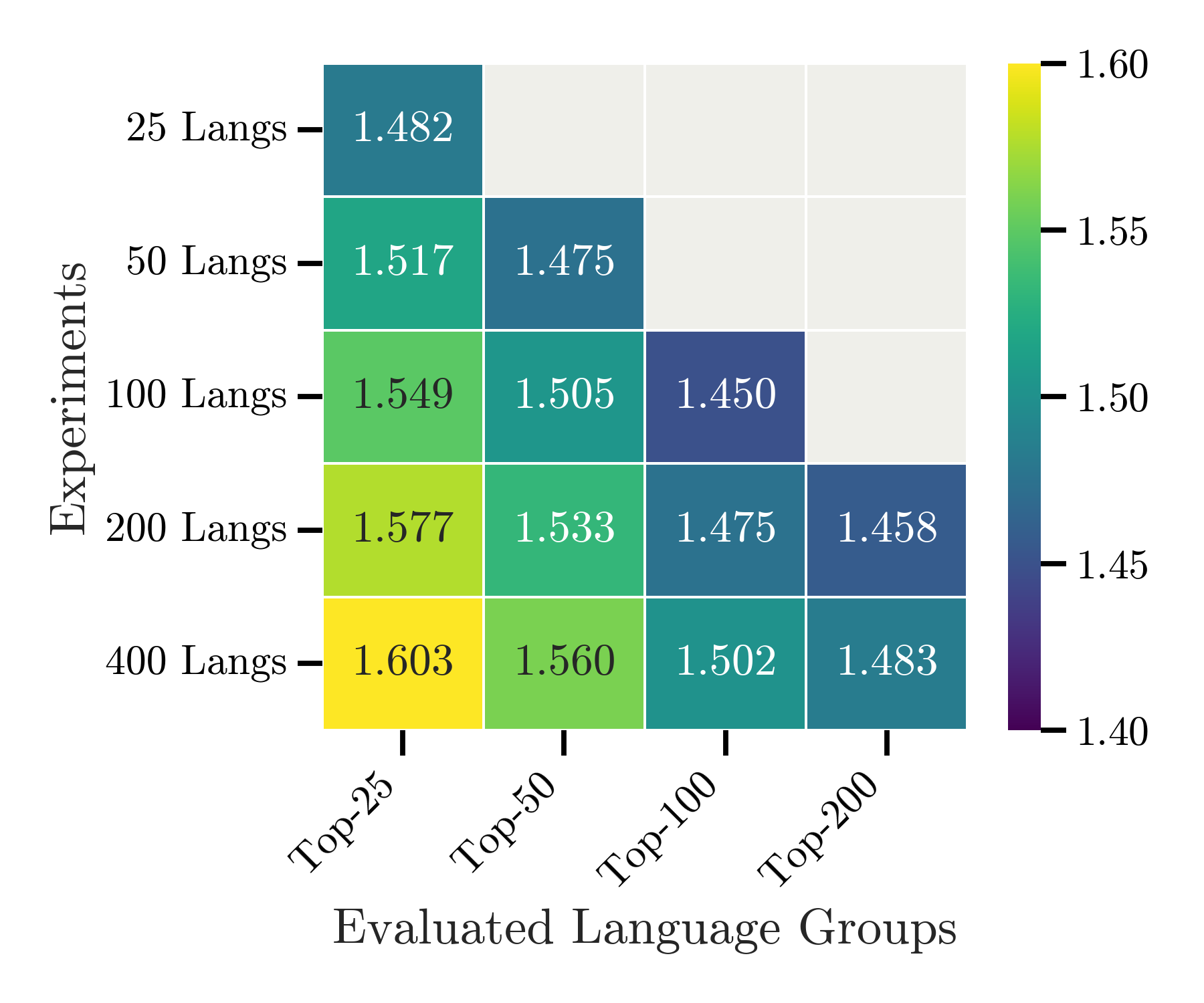}
        \caption{Fixed Total Budget (Temperature Sampling)}
        \label{fig:com_avg_temp}
    \end{subfigure}
        \begin{subfigure}[b]{0.49\linewidth}
        \includegraphics[width=\textwidth]{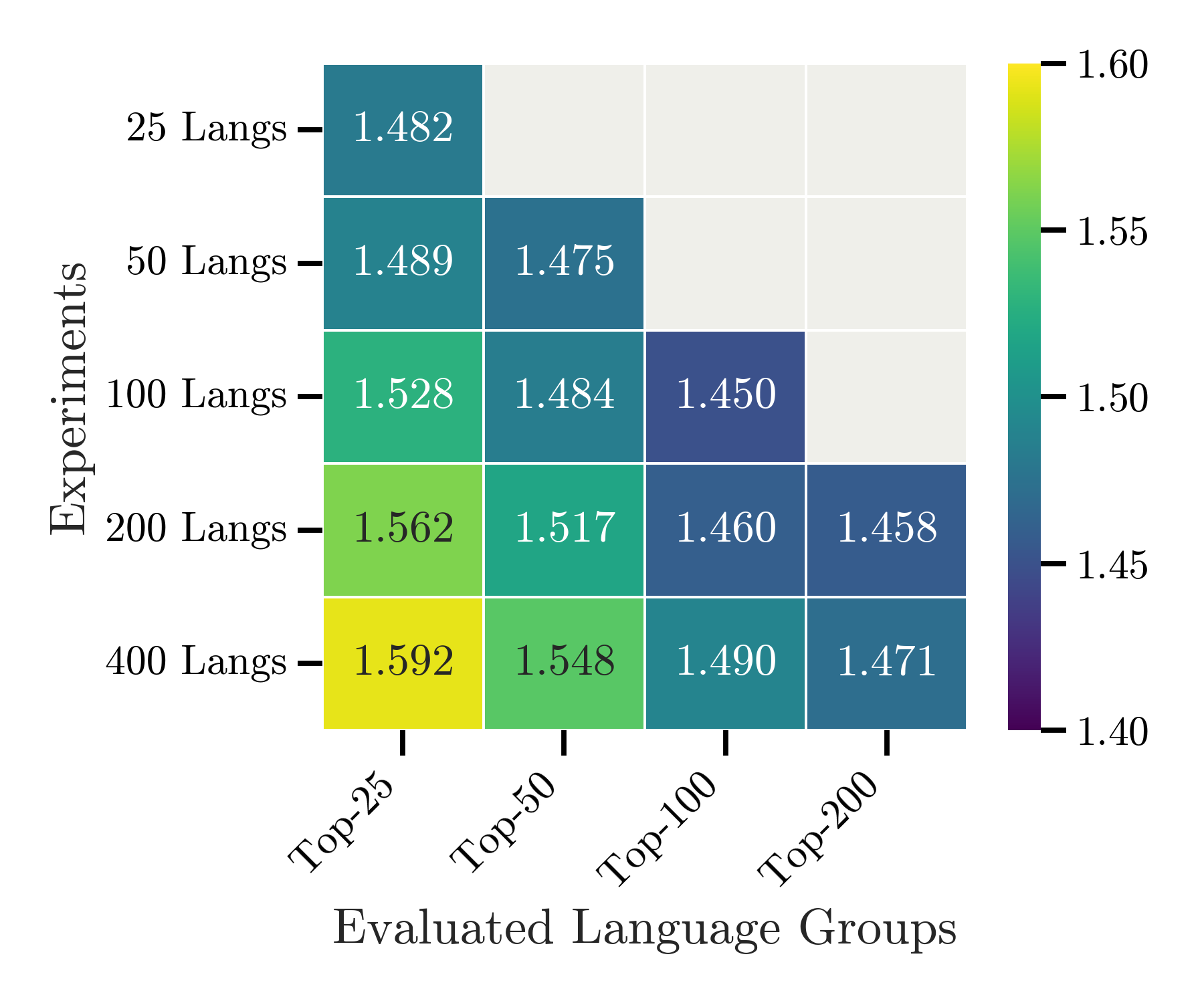}
        \caption{Controlled Growth Setting (Temperature Sampling)}
         \label{fig:com_avg_temp_extended}
    \end{subfigure}

    \caption{Average validation LM loss for different language groups ($x$-axis) across various \curse{} experiments that include more languages in the pretraining mixture ($y$-axis). Increasing the number of languages does not necessarily degrade the performance of languages included in previous experiments, provided that the amount of training data (in tokens) for those languages remains the same. (English is excluded from these evaluations)}
    \label{fig:com_avg_group}
\end{figure}

Practically, we train 3B parameter models on 100B tokens from the \finewebtwo{} corpus. In all settings, English accounts for 40\% of the training data, while the number of non-English languages is systematically increased---from 25 to 400.
We experiment with the top-25, 50, 100, 200, and 400 most frequent languages in \finewebtwo{} under two distributions: (1) the \textit{natural distribution} and (2) \textit{temperature sampling} with $\tau=3.3$.
We then evaluate how increasing linguistic diversity in the non-English data subset affects English and non-English performance. Details of the training data distribution are provided in Table~\ref{tab:token_distribution}.

\paragraph{Results.} Table~\ref{tab:english_com_natural} summarizes English validation loss and average downstream performance across these configurations.
Two main observations emerge. 
First, for a fixed number of languages and a fixed English share, English performance is consistently stronger under the natural distribution than under temperature sampling. In this case, English benefits from cross-lingual transfer with high-resource, typologically related languages (\eg, German, French), which receive more data under the natural distribution (we further investigate this effect in Appendix~\ref{sec:cross_lingual_transfer}).
Second, even when scaling up to 400 languages, English performance remains largely stable---particularly under the natural distribution, suggesting that English performance is not determined by the sheer number of languages included in the training process. In other words, the key factor is not how many languages are present, but how the training data is distributed among them.

Building on this insight, we show in Figures~\ref{fig:com_avg_natural} and \ref{fig:com_avg_temp} the weighted average LM validation loss for the top-25, 50, 100, and 200 language groups (excluding English) under the two distributions.
The $x$-axis denotes language groups used for evaluation, while the $y$-axis indicates language groups used for training. 
Because the total data budget is fixed at 100B tokens, adding more languages necessarily reduces the relative share of data for previously included ones.
Under the natural distribution, however, performance remains stable as languages are added. In contrast, under temperature sampling we observe up to a $\sim$0.1 increase in loss when expanding from 25 to 400 languages.
This effect is expected, since temperature sampling reduces the allocation of mid- and high-resource languages more aggressively, amplifying the effect of including low-resource ones.

To disentangle the effect of adding new languages from the effect of reducing data for existing ones, we also run a controlled setting where the data for the original set of languages remains fixed across two consecutive runs.
For example, when increasing from 25 to 50 languages, the first 25 languages receive the same amount of training data as before; the same approach is applied when scaling from 50 to 100, 100 to 200, and 200 to 400 languages.
Figures~\ref{fig:com_avg_natural_extended} and \ref{fig:com_avg_temp_extended} report the results. Once again, under the natural distribution, performance remains stable, and we also observe a smaller relative degradation for the temperature sampling setting.

Taken together, these results suggest that the \curse{} is not primarily about the number of languages added, but instead reflects limitations in model capacity and the quality and distribution of multilingual data.
Under the natural distribution, the phenomenon is better described as a \textit{curse of capacity}: models have a finite ability to absorb tokens, and beyond a certain point, additional data yields diminishing or even negative returns, a constraint not unique to multilingual models.
Under temperature sampling, the issue more closely resembles a \textit{curse of data quality}: oversampling very low-resource languages introduces more noisy data into training, which hurts performance.

\textbf{Takeaway: } The \curse{}, while measurable, likely arises not from simply adding more languages, but from (1) the finite capacity of models and (2) data distributions that too strongly amplify the impact of languages represented by lower-quality data.
\section{Related Work}
\label{related_work}

\paragraph{Pretraining Data Mixture. }
Prior work has explored the impact of pretraining data composition on the performance of large language models (LLMs) \citep{gu-etal-2024-cmr, zhao-etal-2024-deciphering, xie2023data, albalak2312efficient, held2025optimizing,swissai2025apertus}. 
Several studies have proposed algorithms to optimize domain weights using proxy models, thereby improving the generalization ability of LLMs \citep{xie2024doremi, fan2023doge}.
Another approach formulates the identification of high-performing data mixtures as a regression problem~\citep{liu2024regmix}.

In the multilingual setting, temperature-based sampling has traditionally been used to balance representation across languages~\citep{devlin-etal-2019-bert, xue-etal-2021-mt5}. However, this heuristic method can lead to overfitting on low-resource (tail) languages. To address this, \cite{chung2023unimax} proposes a sampling method that ensures more uniform coverage of high-resource (head) languages while capping repetition on low-resource languages. Additionally, \cite{he2024scalinglawsmultilinguallanguage} investigates scaling laws specific to multilingual LLMs, providing further insight into optimal data mixture strategies.

\textbf{Curse of Multilinguality \& Negative Interference.}
The \curse{}, introduced by~\cite{conneau-etal-2020-unsupervised}, describes the phenomenon where, under a fixed model capacity, adding more languages initially improves cross-lingual performance---especially for low-resource languages---but eventually leads to degradation in both monolingual and cross-lingual performance.
Most previous investigations into this phenomenon have been limited in scale, in terms of both model size and language coverage. For instance,~\cite{pfeiffer-etal-2022-lifting} studies this trade-off using a $270$M parameter bidirectional model trained on 75 languages, proposing a modular architecture to mitigate interference. Recently,~\cite{chuang2025metaclip2worldwide} shows that the \curse{} breaks with larger count of parameters for multimodal embedding tasks. 

\cite{blevins-etal-2024-breaking} introduces a cross-lingual expert language model, in which separate models are trained on subsets of the multilingual corpus to reduce competition among languages. Similarly, \cite{chang-etal-2024-multilinguality} explores this effect using monolingual and multilingual models (up to 45M parameters) trained across 250 languages and derives optimal sampling ratios. \cite{wang-etal-2020-negative} examines the phenomenon of negative interference in multilingual LMs and introduces a meta-learning algorithm that improves cross-lingual transfer and alleviates interference effects. \cite{alastruey2025interference} challenge the prevailing assumption that cross-lingual interference depends on language family, showing instead that it is primarily related to script.

\textbf{Impact of Pivot Languages. }
The role of \textit{pivot} languages in improving monolingual and cross-lingual performance of multilingual LLMs has been studied before. Several works have demonstrated the benefits of using a pivot language for machine translation~\citep{kim-etal-2019-pivot, zou2022investigating, gaikwad2024effective, mohammadshahi-etal-2024-investigating}.
\cite{zhang-etal-2024-plug} shows that using English as a pivot for cross-lingual instruction tuning, by first interpreting instructions in English before generating responses in the target language, can be highly effective. Pivot languages have also been used to improve alignment in multilingual representation spaces~\citep{zhao2024lens}.

\textbf{Curriculum Learning (CL) for LLMs. }
Curriculum Learning (CL), a data-centric training strategy inspired by human learning processes, has been studied for improving the performance of LLMs~\citep{nair-etal-2024-curriculum, kim2024strategic, li2021curriculum}. Several studies have demonstrated the effectiveness of CL in multilingual machine translation~\citep{zhang-etal-2021-competence-based, kumar2021learning, zhou-etal-2021-self, choi2023order}.
\cite{ranaldi-etal-2024-language} applies the CL paradigm during the instruction-tuning phase of multilingual LLMs and reports notable improvements. Additionally, \cite{yoo2024code} proposes a code-switching-based CL strategy to enhance cross-lingual transfer capabilities in LLMs.

\section{Discussion \& Conclusion}
\label{sec:discussion}

Our study investigates the influence of data mixture composition on multilingual large language model (LLM) pretraining, leveraging $1.1$B and $3$B models across up to $400$ languages. Our findings challenge prevailing assumptions about multilingual pretraining, offering direct guidance for multilingual data mixture design. First, we demonstrated that the quantity and proportion of high-resource English data do not inherently compromise multilingual performance, provided a sufficient number of non-English tokens are present. This finding suggests practitioners should prioritize ensuring an adequate absolute volume of diverse, high-quality multilingual content over strictly reducing the high-resource component. Second, and contrary to suggestions that family-specific pivots are most effective, we established that English consistently serves as a high-quality pivot language, providing cross-lingual transfer benefits across language families.

We also provided new insights into core challenges of multilingual scaling: negative interference and the ``\curse{}.'' Our results showed that staging the introduction of languages through curriculum learning does not mitigate negative interference \citep{wang-etal-2020-negative} or improve non-English performance, suggesting that interference is a fundamental problem related to the fixed capacity of models, and not merely one that can be fixed through different data curricula. Furthermore, our findings refine the understanding of the ``\curse{}''\citep{chang-etal-2024-multilinguality}, demonstrating that performance degradation arises not from the simple count of languages added, but from the finite capacity of models and data distributions that amplify the impact of noisy, low-resource languages.

Collectively, these findings translate into the following practices for training multilingual LLMs.
First, adopting regimes taking into account the order of languages such as curriculum learning offer no demonstrable benefit over a well-mixed approach.
Second, given the resilience to high English proportions, focus resource investment on scaling and cleaning low-resource data rather than on costly data balancing operations.
Third, do not limit language coverage arbitrarily, as the curse is primarily a function of quality and quantity of the multilingual data, not language count. Our evidence implies that future efforts to break the curse should focus on including adequate high-quality data for each language.
While these principles were established on $1.1$B and $3$B parameter models, future work must validate these trade-offs on larger models models to explore how increased model capacity potentially alters the non-linear relationship between data composition, interference, and performance.
\section*{Limitations}
Despite employing larger models and more data than prior work, our study remains far below the scale of frontier models such as \cite{meta2024llama4, guo2025deepseek}, as operating at that scale would have prevented us from running the number of experiments necessary to draw reasonable conclusions within our computational constraints. Furthermore, we were unable to explore the impact of post-training and the effects of various data sampling strategies for the same reason. Lastly, the choice of our tokenizer may limit performance on lower-resource languages. We selected a pre-existing tokenizer that supported the greatest number of languages in our study, as training a tokenizer to support 1,834 languages is practically infeasible without substantially increasing the model’s vocabulary size and the associated GPU memory requirements.

\section*{Acknowledgment}
This work was supported as part of the Swiss AI Initiative by a grant from the Swiss National Supercomputing Centre (CSCS) under project ID a06 on Alps. We also gratefully acknowledge the support of the Swiss National Science Foundation (No. 215390), Innosuisse (PFFS-21-29), the EPFL Center for Imaging, Sony Group Corporation, and a Meta LLM Evaluation Research Grant.

\bibliography{iclr2026_conference}

\begin{thebibliography}{66}
\providecommand{\natexlab}[1]{#1}
\providecommand{\url}[1]{\texttt{#1}}
\expandafter\ifx\csname urlstyle\endcsname\relax
  \providecommand{\doi}[1]{doi: #1}\else
  \providecommand{\doi}{doi: \begingroup \urlstyle{rm}\Url}\fi

\bibitem[Alastruey et~al.(2025)Alastruey, Janeiro, Allauzen, Elbayad, Barrault,
  and Costa-juss{\`a}]{alastruey2025interference}
Belen Alastruey, Jo{\~a}o~Maria Janeiro, Alexandre Allauzen, Maha Elbayad,
  Lo{\"\i}c Barrault, and Marta~R Costa-juss{\`a}.
\newblock Interference matrix: Quantifying cross-lingual interference in
  transformer encoders.
\newblock \emph{arXiv preprint arXiv:2508.02256}, 2025.

\bibitem[Albalak et~al.(2023)Albalak, Pan, Raffel, and
  Wang]{albalak2312efficient}
Alon Albalak, Liangming Pan, Colin Raffel, and William~Yang Wang.
\newblock Efficient online data mixing for language model pre-training.
\newblock \emph{URL https://arxiv. org/abs/2312.02406}, 2023.

\bibitem[Apertus~Project et~al.(2025)Apertus~Project, Hägele, Huang, Romanou,
  Solergibert, Pasztor, Messmer, Garbaya, Ďurech, Hakimi, Giraldo,
  Ismayilzada, Foroutan, Moalla, Chen, Sabolčec, Xu, Aerni, AlKhamissi,
  Marinas, Amani, Ansaripour, Badanin, Benoit, Boros, Browning, Bösch,
  Böther, Canova, Challier, Charmillot, Coles, Deriu, Devos, Drescher,
  Dzenhaliou, Ehrmann, Fan, Fan, Gao, Gila, Grandury, Hashemi, Hoyle, Jiang,
  Klein, Kucharavy, Kucherenko, Lübeck, Machacek, Manitaras, Marfurt, Matoba,
  Matrenok, Mendoncça, Mohamed, Montariol, Mouchel, Najem-Meyer, Ni, Oliva,
  Pagliardini, Palme, Panferov, Paoletti, Passerini, Pavlov, Poiroux, Ponkshe,
  Ranchin, Rando, Sauser, Saydaliev, Sayfiddinov, Schneider, Schuppli,
  Scialanga, Semenov, Shridhar, Singhal, Sotnikova, Sternfeld, Tarun,
  Teiletche, Vamvas, Yao, Ilic, Klimovic, Krause, Gulcehre, Rosenthal, Ash,
  Tramèr, VandeVondele, Veraldi, Rajman, Schulthess, Hoefler, Bosselut, Jaggi,
  and Schlag]{swissai2025apertus}
Alejandro Hernández-Cano Apertus~Project, Alexander Hägele, Allen~Hao Huang,
  Angelika Romanou, Antoni-Joan Solergibert, Barna Pasztor, Bettina Messmer,
  Dhia Garbaya, Eduard~Frank Ďurech, Ido Hakimi, Juan~García Giraldo, Mete
  Ismayilzada, Negar Foroutan, Skander Moalla, Tiancheng Chen, Vinko Sabolčec,
  Yixuan Xu, Michael Aerni, Badr AlKhamissi, Ines~Altemir Marinas,
  Mohammad~Hossein Amani, Matin Ansaripour, Ilia Badanin, Harold Benoit,
  Emanuela Boros, Nicholas Browning, Fabian Bösch, Maximilian Böther, Niklas
  Canova, Camille Challier, Clement Charmillot, Jonathan Coles, Jan Deriu,
  Arnout Devos, Lukas Drescher, Daniil Dzenhaliou, Maud Ehrmann, Dongyang Fan,
  Simin Fan, Silin Gao, Miguel Gila, María Grandury, Diba Hashemi, Alexander
  Hoyle, Jiaming Jiang, Mark Klein, Andrei Kucharavy, Anastasiia Kucherenko,
  Frederike Lübeck, Roman Machacek, Theofilos Manitaras, Andreas Marfurt, Kyle
  Matoba, Simon Matrenok, Henrique Mendoncça, Fawzi~Roberto Mohamed, Syrielle
  Montariol, Luca Mouchel, Sven Najem-Meyer, Jingwei Ni, Gennaro Oliva, Matteo
  Pagliardini, Elia Palme, Andrei Panferov, Léo Paoletti, Marco Passerini,
  Ivan Pavlov, Auguste Poiroux, Kaustubh Ponkshe, Nathan Ranchin, Javi Rando,
  Mathieu Sauser, Jakhongir Saydaliev, Muhammad~Ali Sayfiddinov, Marian
  Schneider, Stefano Schuppli, Marco Scialanga, Andrei Semenov, Kumar Shridhar,
  Raghav Singhal, Anna Sotnikova, Alexander Sternfeld, Ayush~Kumar Tarun, Paul
  Teiletche, Jannis Vamvas, Xiaozhe Yao, Hao Zhao~Alexander Ilic, Ana Klimovic,
  Andreas Krause, Caglar Gulcehre, David Rosenthal, Elliott Ash, Florian
  Tramèr, Joost VandeVondele, Livio Veraldi, Martin Rajman, Thomas Schulthess,
  Torsten Hoefler, Antoine Bosselut, Martin Jaggi, and Imanol Schlag.
\newblock Apertus: Democratizing open and compliant llms for global language
  environments, 2025.

\bibitem[Bagheri~Nezhad \& Agrawal(2024)Bagheri~Nezhad and
  Agrawal]{bagheri-nezhad-agrawal-2024-drives}
Sina Bagheri~Nezhad and Ameeta Agrawal.
\newblock What drives performance in multilingual language models?
\newblock In Yves Scherrer, Tommi Jauhiainen, Nikola Ljube{\v{s}}i{\'c}, Marcos
  Zampieri, Preslav Nakov, and J{\"o}rg Tiedemann (eds.), \emph{Proceedings of
  the Eleventh Workshop on NLP for Similar Languages, Varieties, and Dialects
  (VarDial 2024)}, pp.\  16--27, Mexico City, Mexico, June 2024. Association
  for Computational Linguistics.
\newblock \doi{10.18653/v1/2024.vardial-1.2}.
\newblock URL \url{https://aclanthology.org/2024.vardial-1.2/}.

\bibitem[Bandarkar et~al.(2024)Bandarkar, Liang, Muller, Artetxe, Shukla, Husa,
  Goyal, Krishnan, Zettlemoyer, and Khabsa]{bandarkar-etal-2024-belebele}
Lucas Bandarkar, Davis Liang, Benjamin Muller, Mikel Artetxe, Satya~Narayan
  Shukla, Donald Husa, Naman Goyal, Abhinandan Krishnan, Luke Zettlemoyer, and
  Madian Khabsa.
\newblock The belebele benchmark: a parallel reading comprehension dataset in
  122 language variants.
\newblock In Lun-Wei Ku, Andre Martins, and Vivek Srikumar (eds.),
  \emph{Proceedings of the 62nd Annual Meeting of the Association for
  Computational Linguistics (Volume 1: Long Papers)}, pp.\  749--775, Bangkok,
  Thailand, August 2024. Association for Computational Linguistics.
\newblock \doi{10.18653/v1/2024.acl-long.44}.
\newblock URL \url{https://aclanthology.org/2024.acl-long.44/}.

\bibitem[Blevins et~al.(2024)Blevins, Limisiewicz, Gururangan, Li, Gonen,
  Smith, and Zettlemoyer]{blevins-etal-2024-breaking}
Terra Blevins, Tomasz Limisiewicz, Suchin Gururangan, Margaret Li, Hila Gonen,
  Noah~A. Smith, and Luke Zettlemoyer.
\newblock Breaking the curse of multilinguality with cross-lingual expert
  language models.
\newblock In Yaser Al-Onaizan, Mohit Bansal, and Yun-Nung Chen (eds.),
  \emph{Proceedings of the 2024 Conference on Empirical Methods in Natural
  Language Processing}, pp.\  10822--10837, Miami, Florida, USA, November 2024.
  Association for Computational Linguistics.
\newblock \doi{10.18653/v1/2024.emnlp-main.604}.
\newblock URL \url{https://aclanthology.org/2024.emnlp-main.604/}.

\bibitem[Chang et~al.(2024)Chang, Arnett, Tu, and
  Bergen]{chang-etal-2024-multilinguality}
Tyler~A. Chang, Catherine Arnett, Zhuowen Tu, and Ben Bergen.
\newblock When is multilinguality a curse? language modeling for 250 high- and
  low-resource languages.
\newblock In Yaser Al-Onaizan, Mohit Bansal, and Yun-Nung Chen (eds.),
  \emph{Proceedings of the 2024 Conference on Empirical Methods in Natural
  Language Processing}, pp.\  4074--4096, Miami, Florida, USA, November 2024.
  Association for Computational Linguistics.
\newblock \doi{10.18653/v1/2024.emnlp-main.236}.
\newblock URL \url{https://aclanthology.org/2024.emnlp-main.236/}.

\bibitem[Chen et~al.(2019)Chen, D{'}Arcy, Liu, Fernandez, and
  Downey]{Chen2019CODAHAA}
Michael Chen, Mike D{'}Arcy, Alisa Liu, Jared Fernandez, and Doug Downey.
\newblock Codah: An adversarially-authored question answering dataset for
  common sense.
\newblock In \emph{Proceedings of the 3rd Workshop on Evaluating Vector Space
  Representations for {NLP}}, pp.\  63--69, Minneapolis, USA, 2019. Association
  for Computational Linguistics.
\newblock \doi{10.18653/v1/W19-2008}.
\newblock URL \url{https://www.aclweb.org/anthology/W19-2008}.

\bibitem[Choi et~al.(2023)Choi, Xin, Dadkhahi, Gilmer, Garg, Firat, Yeh, Dai,
  and Ghorbani]{choi2023order}
Dami Choi, Derrick Xin, Hamid Dadkhahi, Justin Gilmer, Ankush Garg, Orhan
  Firat, Chih-Kuan Yeh, Andrew~M Dai, and Behrooz Ghorbani.
\newblock Order matters in the presence of dataset imbalance for multilingual
  learning.
\newblock \emph{Advances in Neural Information Processing Systems},
  36:\penalty0 66902--66922, 2023.

\bibitem[Chuang et~al.(2025)Chuang, Li, Wang, Yeh, Lyu, Raghavendra, Glass,
  Huang, Weston, Zettlemoyer, Chen, Liu, Xie, tau Yih, Li, and
  Xu]{chuang2025metaclip2worldwide}
Yung-Sung Chuang, Yang Li, Dong Wang, Ching-Feng Yeh, Kehan Lyu, Ramya
  Raghavendra, James Glass, Lifei Huang, Jason Weston, Luke Zettlemoyer, Xinlei
  Chen, Zhuang Liu, Saining Xie, Wen tau Yih, Shang-Wen Li, and Hu~Xu.
\newblock Meta clip 2: A worldwide scaling recipe, 2025.
\newblock URL \url{https://arxiv.org/abs/2507.22062}.

\bibitem[Chung et~al.(2023)Chung, Constant, Garcia, Roberts, Tay, Narang, and
  Firat]{chung2023unimax}
Hyung~Won Chung, Noah Constant, Xavier Garcia, Adam Roberts, Yi~Tay, Sharan
  Narang, and Orhan Firat.
\newblock Unimax: Fairer and more effective language sampling for large-scale
  multilingual pretraining.
\newblock \emph{arXiv preprint arXiv:2304.09151}, 2023.

\bibitem[Conneau et~al.(2020)Conneau, Khandelwal, Goyal, Chaudhary, Wenzek,
  Guzm{\'a}n, Grave, Ott, Zettlemoyer, and
  Stoyanov]{conneau-etal-2020-unsupervised}
Alexis Conneau, Kartikay Khandelwal, Naman Goyal, Vishrav Chaudhary, Guillaume
  Wenzek, Francisco Guzm{\'a}n, Edouard Grave, Myle Ott, Luke Zettlemoyer, and
  Veselin Stoyanov.
\newblock Unsupervised cross-lingual representation learning at scale.
\newblock In Dan Jurafsky, Joyce Chai, Natalie Schluter, and Joel Tetreault
  (eds.), \emph{Proceedings of the 58th Annual Meeting of the Association for
  Computational Linguistics}, pp.\  8440--8451, Online, July 2020. Association
  for Computational Linguistics.
\newblock \doi{10.18653/v1/2020.acl-main.747}.
\newblock URL \url{https://aclanthology.org/2020.acl-main.747/}.

\bibitem[Dac~Lai et~al.(2023)Dac~Lai, Van~Nguyen, Ngo, Nguyen, Dernoncourt,
  Rossi, and Nguyen]{dac2023okapi}
Viet Dac~Lai, Chien Van~Nguyen, Nghia~Trung Ngo, Thuat Nguyen, Franck
  Dernoncourt, Ryan~A Rossi, and Thien~Huu Nguyen.
\newblock Okapi: Instruction-tuned large language models in multiple languages
  with reinforcement learning from human feedback.
\newblock \emph{arXiv e-prints}, pp.\  arXiv--2307, 2023.

\bibitem[Devlin et~al.(2019)Devlin, Chang, Lee, and
  Toutanova]{devlin-etal-2019-bert}
Jacob Devlin, Ming-Wei Chang, Kenton Lee, and Kristina Toutanova.
\newblock {BERT}: Pre-training of deep bidirectional transformers for language
  understanding.
\newblock In Jill Burstein, Christy Doran, and Thamar Solorio (eds.),
  \emph{Proceedings of the 2019 Conference of the North {A}merican Chapter of
  the Association for Computational Linguistics: Human Language Technologies,
  Volume 1 (Long and Short Papers)}, pp.\  4171--4186, Minneapolis, Minnesota,
  June 2019. Association for Computational Linguistics.
\newblock \doi{10.18653/v1/N19-1423}.
\newblock URL \url{https://aclanthology.org/N19-1423/}.

\bibitem[Dryer \& Haspelmath(2013)Dryer and Haspelmath]{wals}
Matthew~S. Dryer and Martin Haspelmath (eds.).
\newblock \emph{WALS Online (v2020.4)}.
\newblock Zenodo, 2013.
\newblock \doi{10.5281/zenodo.13950591}.
\newblock URL \url{https://doi.org/10.5281/zenodo.13950591}.

\bibitem[Fan et~al.(2023)Fan, Pagliardini, and Jaggi]{fan2023doge}
Simin Fan, Matteo Pagliardini, and Martin Jaggi.
\newblock Doge: Domain reweighting with generalization estimation.
\newblock \emph{arXiv preprint arXiv:2310.15393}, 2023.

\bibitem[Gaikwad et~al.(2024)Gaikwad, Doshi, Dabre, and
  Bhattacharyya]{gaikwad2024effective}
Pranav Gaikwad, Meet Doshi, Raj Dabre, and Pushpak Bhattacharyya.
\newblock How effective is multi-source pivoting for translation of low
  resource indian languages?
\newblock \emph{arXiv preprint arXiv:2406.13332}, 2024.

\bibitem[Gpt-4~Team et~al.(2023)Gpt-4~Team, Adler, Agarwal, Ahmad, Akkaya,
  Aleman, Almeida, Altenschmidt, Altman, Anadkat, et~al.]{achiam2023gpt}
Josh Gpt-4~Team, Achiam, Steven Adler, Sandhini Agarwal, Lama Ahmad, Ilge
  Akkaya, Florencia~Leoni Aleman, Diogo Almeida, Janko Altenschmidt, Sam
  Altman, Shyamal Anadkat, et~al.
\newblock Gpt-4 technical report.
\newblock \emph{arXiv preprint arXiv:2303.08774}, 2023.

\bibitem[Grattafiori et~al.(2024)Grattafiori, Dubey, Jauhri, Pandey, Kadian,
  Al-Dahle, Letman, Mathur, Schelten, Vaughan, Yang, Fan, Goyal, Hartshorn,
  Yang, Mitra, Sravankumar, Korenev, Hinsvark, Rao, Zhang, Rodriguez,
  Gregerson, Spataru, Roziere, Biron, Tang, Chern, Caucheteux, Nayak, Bi,
  Marra, McConnell, Keller, Touret, Wu, Wong, Ferrer, Nikolaidis, Allonsius,
  Song, Pintz, Livshits, Wyatt, Esiobu, Choudhary, Mahajan, Garcia-Olano,
  Perino, Hupkes, Lakomkin, AlBadawy, Lobanova, Dinan, Smith, Radenovic,
  Guzmán, Zhang, Synnaeve, Lee, Anderson, Thattai, Nail, Mialon, Pang,
  Cucurell, Nguyen, Korevaar, Xu, Touvron, Zarov, Ibarra, Kloumann, Misra,
  Evtimov, Zhang, Copet, Lee, Geffert, Vranes, Park, Mahadeokar, Shah, van~der
  Linde, Billock, Hong, Lee, Fu, Chi, Huang, Liu, Wang, Yu, Bitton, Spisak,
  Park, Rocca, Johnstun, Saxe, Jia, Alwala, Prasad, Upasani, Plawiak, Li,
  Heafield, Stone, El-Arini, Iyer, Malik, Chiu, Bhalla, Lakhotia,
  Rantala-Yeary, van~der Maaten, Chen, Tan, Jenkins, Martin, Madaan, Malo,
  Blecher, Landzaat, de~Oliveira, Muzzi, Pasupuleti, Singh, Paluri, Kardas,
  Tsimpoukelli, Oldham, Rita, Pavlova, Kambadur, Lewis, Si, Singh, Hassan,
  Goyal, Torabi, Bashlykov, Bogoychev, Chatterji, Zhang, Duchenne, Çelebi,
  Alrassy, Zhang, Li, Vasic, Weng, Bhargava, Dubal, Krishnan, Koura, Xu, He,
  Dong, Srinivasan, Ganapathy, Calderer, Cabral, Stojnic, Raileanu, Maheswari,
  Girdhar, Patel, Sauvestre, Polidoro, Sumbaly, Taylor, Silva, Hou, Wang,
  Hosseini, Chennabasappa, Singh, Bell, Kim, Edunov, Nie, Narang, Raparthy,
  Shen, Wan, Bhosale, Zhang, Vandenhende, Batra, Whitman, Sootla, Collot,
  Gururangan, Borodinsky, Herman, Fowler, Sheasha, Georgiou, Scialom,
  Speckbacher, Mihaylov, Xiao, Karn, Goswami, Gupta, Ramanathan, Kerkez,
  Gonguet, Do, Vogeti, Albiero, Petrovic, Chu, Xiong, Fu, Meers, Martinet,
  Wang, Wang, Tan, Xia, Xie, Jia, Wang, Goldschlag, Gaur, Babaei, Wen, Song,
  Zhang, Li, Mao, Coudert, Yan, Chen, Papakipos, Singh, Srivastava, Jain,
  Kelsey, Shajnfeld, Gangidi, Victoria, Goldstand, Menon, Sharma, Boesenberg,
  Baevski, Feinstein, Kallet, Sangani, Teo, Yunus, Lupu, Alvarado, Caples, Gu,
  Ho, Poulton, Ryan, Ramchandani, Dong, Franco, Goyal, Saraf, Chowdhury,
  Gabriel, Bharambe, Eisenman, Yazdan, James, Maurer, Leonhardi, Huang, Loyd,
  Paola, Paranjape, Liu, Wu, Ni, Hancock, Wasti, Spence, Stojkovic, Gamido,
  Montalvo, Parker, Burton, Mejia, Liu, Wang, Kim, Zhou, Hu, Chu, Cai, Tindal,
  Feichtenhofer, Gao, Civin, Beaty, Kreymer, Li, Adkins, Xu, Testuggine, David,
  Parikh, Liskovich, Foss, Wang, Le, Holland, Dowling, Jamil, Montgomery,
  Presani, Hahn, Wood, Le, Brinkman, Arcaute, Dunbar, Smothers, Sun, Kreuk,
  Tian, Kokkinos, Ozgenel, Caggioni, Kanayet, Seide, Florez, Schwarz, Badeer,
  Swee, Halpern, Herman, Sizov, Guangyi, Zhang, Lakshminarayanan, Inan,
  Shojanazeri, Zou, Wang, Zha, Habeeb, Rudolph, Suk, Aspegren, Goldman, Zhan,
  Damlaj, Molybog, Tufanov, Leontiadis, Veliche, Gat, Weissman, Geboski, Kohli,
  Lam, Asher, Gaya, Marcus, Tang, Chan, Zhen, Reizenstein, Teboul, Zhong, Jin,
  Yang, Cummings, Carvill, Shepard, McPhie, Torres, Ginsburg, Wang, Wu, U,
  Saxena, Khandelwal, Zand, Matosich, Veeraraghavan, Michelena, Li, Jagadeesh,
  Huang, Chawla, Huang, Chen, Garg, A, Silva, Bell, Zhang, Guo, Yu, Moshkovich,
  Wehrstedt, Khabsa, Avalani, Bhatt, Mankus, Hasson, Lennie, Reso, Groshev,
  Naumov, Lathi, Keneally, Liu, Seltzer, Valko, Restrepo, Patel, Vyatskov,
  Samvelyan, Clark, Macey, Wang, Hermoso, Metanat, Rastegari, Bansal,
  Santhanam, Parks, White, Bawa, Singhal, Egebo, Usunier, Mehta, Laptev, Dong,
  Cheng, Chernoguz, Hart, Salpekar, Kalinli, Kent, Parekh, Saab, Balaji,
  Rittner, Bontrager, Roux, Dollar, Zvyagina, Ratanchandani, Yuvraj, Liang,
  Alao, Rodriguez, Ayub, Murthy, Nayani, Mitra, Parthasarathy, Li, Hogan,
  Battey, Wang, Howes, Rinott, Mehta, Siby, Bondu, Datta, Chugh, Hunt, Dhillon,
  Sidorov, Pan, Mahajan, Verma, Yamamoto, Ramaswamy, Lindsay, Lindsay, Feng,
  Lin, Zha, Patil, Shankar, Zhang, Zhang, Wang, Agarwal, Sajuyigbe, Chintala,
  Max, Chen, Kehoe, Satterfield, Govindaprasad, Gupta, Deng, Cho, Virk,
  Subramanian, Choudhury, Goldman, Remez, Glaser, Best, Koehler, Robinson, Li,
  Zhang, Matthews, Chou, Shaked, Vontimitta, Ajayi, Montanez, Mohan, Kumar,
  Mangla, Ionescu, Poenaru, Mihailescu, Ivanov, Li, Wang, Jiang, Bouaziz,
  Constable, Tang, Wu, Wang, Wu, Gao, Kleinman, Chen, Hu, Jia, Qi, Li, Zhang,
  Zhang, Adi, Nam, Yu, Wang, Zhao, Hao, Qian, Li, He, Rait, DeVito, Rosnbrick,
  Wen, Yang, Zhao, and Ma]{grattafiori2024llama3herdmodels}
Aaron Grattafiori, Abhimanyu Dubey, Abhinav Jauhri, Abhinav Pandey, Abhishek
  Kadian, Ahmad Al-Dahle, Aiesha Letman, Akhil Mathur, Alan Schelten, Alex
  Vaughan, Amy Yang, Angela Fan, Anirudh Goyal, Anthony Hartshorn, Aobo Yang,
  Archi Mitra, Archie Sravankumar, Artem Korenev, Arthur Hinsvark, Arun Rao,
  Aston Zhang, Aurelien Rodriguez, Austen Gregerson, Ava Spataru, Baptiste
  Roziere, Bethany Biron, Binh Tang, Bobbie Chern, Charlotte Caucheteux, Chaya
  Nayak, Chloe Bi, Chris Marra, Chris McConnell, Christian Keller, Christophe
  Touret, Chunyang Wu, Corinne Wong, Cristian~Canton Ferrer, Cyrus Nikolaidis,
  Damien Allonsius, Daniel Song, Danielle Pintz, Danny Livshits, Danny Wyatt,
  David Esiobu, Dhruv Choudhary, Dhruv Mahajan, Diego Garcia-Olano, Diego
  Perino, Dieuwke Hupkes, Egor Lakomkin, Ehab AlBadawy, Elina Lobanova, Emily
  Dinan, Eric~Michael Smith, Filip Radenovic, Francisco Guzmán, Frank Zhang,
  Gabriel Synnaeve, Gabrielle Lee, Georgia~Lewis Anderson, Govind Thattai,
  Graeme Nail, Gregoire Mialon, Guan Pang, Guillem Cucurell, Hailey Nguyen,
  Hannah Korevaar, Hu~Xu, Hugo Touvron, Iliyan Zarov, Imanol~Arrieta Ibarra,
  Isabel Kloumann, Ishan Misra, Ivan Evtimov, Jack Zhang, Jade Copet, Jaewon
  Lee, Jan Geffert, Jana Vranes, Jason Park, Jay Mahadeokar, Jeet Shah, Jelmer
  van~der Linde, Jennifer Billock, Jenny Hong, Jenya Lee, Jeremy Fu, Jianfeng
  Chi, Jianyu Huang, Jiawen Liu, Jie Wang, Jiecao Yu, Joanna Bitton, Joe
  Spisak, Jongsoo Park, Joseph Rocca, Joshua Johnstun, Joshua Saxe, Junteng
  Jia, Kalyan~Vasuden Alwala, Karthik Prasad, Kartikeya Upasani, Kate Plawiak,
  Ke~Li, Kenneth Heafield, Kevin Stone, Khalid El-Arini, Krithika Iyer, Kshitiz
  Malik, Kuenley Chiu, Kunal Bhalla, Kushal Lakhotia, Lauren Rantala-Yeary,
  Laurens van~der Maaten, Lawrence Chen, Liang Tan, Liz Jenkins, Louis Martin,
  Lovish Madaan, Lubo Malo, Lukas Blecher, Lukas Landzaat, Luke de~Oliveira,
  Madeline Muzzi, Mahesh Pasupuleti, Mannat Singh, Manohar Paluri, Marcin
  Kardas, Maria Tsimpoukelli, Mathew Oldham, Mathieu Rita, Maya Pavlova,
  Melanie Kambadur, Mike Lewis, Min Si, Mitesh~Kumar Singh, Mona Hassan, Naman
  Goyal, Narjes Torabi, Nikolay Bashlykov, Nikolay Bogoychev, Niladri
  Chatterji, Ning Zhang, Olivier Duchenne, Onur Çelebi, Patrick Alrassy,
  Pengchuan Zhang, Pengwei Li, Petar Vasic, Peter Weng, Prajjwal Bhargava,
  Pratik Dubal, Praveen Krishnan, Punit~Singh Koura, Puxin Xu, Qing He,
  Qingxiao Dong, Ragavan Srinivasan, Raj Ganapathy, Ramon Calderer,
  Ricardo~Silveira Cabral, Robert Stojnic, Roberta Raileanu, Rohan Maheswari,
  Rohit Girdhar, Rohit Patel, Romain Sauvestre, Ronnie Polidoro, Roshan
  Sumbaly, Ross Taylor, Ruan Silva, Rui Hou, Rui Wang, Saghar Hosseini, Sahana
  Chennabasappa, Sanjay Singh, Sean Bell, Seohyun~Sonia Kim, Sergey Edunov,
  Shaoliang Nie, Sharan Narang, Sharath Raparthy, Sheng Shen, Shengye Wan,
  Shruti Bhosale, Shun Zhang, Simon Vandenhende, Soumya Batra, Spencer Whitman,
  Sten Sootla, Stephane Collot, Suchin Gururangan, Sydney Borodinsky, Tamar
  Herman, Tara Fowler, Tarek Sheasha, Thomas Georgiou, Thomas Scialom, Tobias
  Speckbacher, Todor Mihaylov, Tong Xiao, Ujjwal Karn, Vedanuj Goswami, Vibhor
  Gupta, Vignesh Ramanathan, Viktor Kerkez, Vincent Gonguet, Virginie Do, Vish
  Vogeti, Vítor Albiero, Vladan Petrovic, Weiwei Chu, Wenhan Xiong, Wenyin Fu,
  Whitney Meers, Xavier Martinet, Xiaodong Wang, Xiaofang Wang, Xiaoqing~Ellen
  Tan, Xide Xia, Xinfeng Xie, Xuchao Jia, Xuewei Wang, Yaelle Goldschlag,
  Yashesh Gaur, Yasmine Babaei, Yi~Wen, Yiwen Song, Yuchen Zhang, Yue Li,
  Yuning Mao, Zacharie~Delpierre Coudert, Zheng Yan, Zhengxing Chen, Zoe
  Papakipos, Aaditya Singh, Aayushi Srivastava, Abha Jain, Adam Kelsey, Adam
  Shajnfeld, Adithya Gangidi, Adolfo Victoria, Ahuva Goldstand, Ajay Menon,
  Ajay Sharma, Alex Boesenberg, Alexei Baevski, Allie Feinstein, Amanda Kallet,
  Amit Sangani, Amos Teo, Anam Yunus, Andrei Lupu, Andres Alvarado, Andrew
  Caples, Andrew Gu, Andrew Ho, Andrew Poulton, Andrew Ryan, Ankit Ramchandani,
  Annie Dong, Annie Franco, Anuj Goyal, Aparajita Saraf, Arkabandhu Chowdhury,
  Ashley Gabriel, Ashwin Bharambe, Assaf Eisenman, Azadeh Yazdan, Beau James,
  Ben Maurer, Benjamin Leonhardi, Bernie Huang, Beth Loyd, Beto~De Paola,
  Bhargavi Paranjape, Bing Liu, Bo~Wu, Boyu Ni, Braden Hancock, Bram Wasti,
  Brandon Spence, Brani Stojkovic, Brian Gamido, Britt Montalvo, Carl Parker,
  Carly Burton, Catalina Mejia, Ce~Liu, Changhan Wang, Changkyu Kim, Chao Zhou,
  Chester Hu, Ching-Hsiang Chu, Chris Cai, Chris Tindal, Christoph
  Feichtenhofer, Cynthia Gao, Damon Civin, Dana Beaty, Daniel Kreymer, Daniel
  Li, David Adkins, David Xu, Davide Testuggine, Delia David, Devi Parikh,
  Diana Liskovich, Didem Foss, Dingkang Wang, Duc Le, Dustin Holland, Edward
  Dowling, Eissa Jamil, Elaine Montgomery, Eleonora Presani, Emily Hahn, Emily
  Wood, Eric-Tuan Le, Erik Brinkman, Esteban Arcaute, Evan Dunbar, Evan
  Smothers, Fei Sun, Felix Kreuk, Feng Tian, Filippos Kokkinos, Firat Ozgenel,
  Francesco Caggioni, Frank Kanayet, Frank Seide, Gabriela~Medina Florez,
  Gabriella Schwarz, Gada Badeer, Georgia Swee, Gil Halpern, Grant Herman,
  Grigory Sizov, Guangyi, Zhang, Guna Lakshminarayanan, Hakan Inan, Hamid
  Shojanazeri, Han Zou, Hannah Wang, Hanwen Zha, Haroun Habeeb, Harrison
  Rudolph, Helen Suk, Henry Aspegren, Hunter Goldman, Hongyuan Zhan, Ibrahim
  Damlaj, Igor Molybog, Igor Tufanov, Ilias Leontiadis, Irina-Elena Veliche,
  Itai Gat, Jake Weissman, James Geboski, James Kohli, Janice Lam, Japhet
  Asher, Jean-Baptiste Gaya, Jeff Marcus, Jeff Tang, Jennifer Chan, Jenny Zhen,
  Jeremy Reizenstein, Jeremy Teboul, Jessica Zhong, Jian Jin, Jingyi Yang, Joe
  Cummings, Jon Carvill, Jon Shepard, Jonathan McPhie, Jonathan Torres, Josh
  Ginsburg, Junjie Wang, Kai Wu, Kam~Hou U, Karan Saxena, Kartikay Khandelwal,
  Katayoun Zand, Kathy Matosich, Kaushik Veeraraghavan, Kelly Michelena, Keqian
  Li, Kiran Jagadeesh, Kun Huang, Kunal Chawla, Kyle Huang, Lailin Chen,
  Lakshya Garg, Lavender A, Leandro Silva, Lee Bell, Lei Zhang, Liangpeng Guo,
  Licheng Yu, Liron Moshkovich, Luca Wehrstedt, Madian Khabsa, Manav Avalani,
  Manish Bhatt, Martynas Mankus, Matan Hasson, Matthew Lennie, Matthias Reso,
  Maxim Groshev, Maxim Naumov, Maya Lathi, Meghan Keneally, Miao Liu,
  Michael~L. Seltzer, Michal Valko, Michelle Restrepo, Mihir Patel, Mik
  Vyatskov, Mikayel Samvelyan, Mike Clark, Mike Macey, Mike Wang, Miquel~Jubert
  Hermoso, Mo~Metanat, Mohammad Rastegari, Munish Bansal, Nandhini Santhanam,
  Natascha Parks, Natasha White, Navyata Bawa, Nayan Singhal, Nick Egebo,
  Nicolas Usunier, Nikhil Mehta, Nikolay~Pavlovich Laptev, Ning Dong, Norman
  Cheng, Oleg Chernoguz, Olivia Hart, Omkar Salpekar, Ozlem Kalinli, Parkin
  Kent, Parth Parekh, Paul Saab, Pavan Balaji, Pedro Rittner, Philip Bontrager,
  Pierre Roux, Piotr Dollar, Polina Zvyagina, Prashant Ratanchandani, Pritish
  Yuvraj, Qian Liang, Rachad Alao, Rachel Rodriguez, Rafi Ayub, Raghotham
  Murthy, Raghu Nayani, Rahul Mitra, Rangaprabhu Parthasarathy, Raymond Li,
  Rebekkah Hogan, Robin Battey, Rocky Wang, Russ Howes, Ruty Rinott, Sachin
  Mehta, Sachin Siby, Sai~Jayesh Bondu, Samyak Datta, Sara Chugh, Sara Hunt,
  Sargun Dhillon, Sasha Sidorov, Satadru Pan, Saurabh Mahajan, Saurabh Verma,
  Seiji Yamamoto, Sharadh Ramaswamy, Shaun Lindsay, Shaun Lindsay, Sheng Feng,
  Shenghao Lin, Shengxin~Cindy Zha, Shishir Patil, Shiva Shankar, Shuqiang
  Zhang, Shuqiang Zhang, Sinong Wang, Sneha Agarwal, Soji Sajuyigbe, Soumith
  Chintala, Stephanie Max, Stephen Chen, Steve Kehoe, Steve Satterfield,
  Sudarshan Govindaprasad, Sumit Gupta, Summer Deng, Sungmin Cho, Sunny Virk,
  Suraj Subramanian, Sy~Choudhury, Sydney Goldman, Tal Remez, Tamar Glaser,
  Tamara Best, Thilo Koehler, Thomas Robinson, Tianhe Li, Tianjun Zhang, Tim
  Matthews, Timothy Chou, Tzook Shaked, Varun Vontimitta, Victoria Ajayi,
  Victoria Montanez, Vijai Mohan, Vinay~Satish Kumar, Vishal Mangla, Vlad
  Ionescu, Vlad Poenaru, Vlad~Tiberiu Mihailescu, Vladimir Ivanov, Wei Li,
  Wenchen Wang, Wenwen Jiang, Wes Bouaziz, Will Constable, Xiaocheng Tang,
  Xiaojian Wu, Xiaolan Wang, Xilun Wu, Xinbo Gao, Yaniv Kleinman, Yanjun Chen,
  Ye~Hu, Ye~Jia, Ye~Qi, Yenda Li, Yilin Zhang, Ying Zhang, Yossi Adi, Youngjin
  Nam, Yu, Wang, Yu~Zhao, Yuchen Hao, Yundi Qian, Yunlu Li, Yuzi He, Zach Rait,
  Zachary DeVito, Zef Rosnbrick, Zhaoduo Wen, Zhenyu Yang, Zhiwei Zhao, and
  Zhiyu Ma.
\newblock The llama 3 herd of models, 2024.
\newblock URL \url{https://arxiv.org/abs/2407.21783}.

\bibitem[Gu et~al.(2024)Gu, Yang, Ding, Zhao, and Tan]{gu-etal-2024-cmr}
Jiawei Gu, Zacc Yang, Chuanghao Ding, Rui Zhao, and Fei Tan.
\newblock {CMR} scaling law: Predicting critical mixture ratios for continual
  pre-training of language models.
\newblock In Yaser Al-Onaizan, Mohit Bansal, and Yun-Nung Chen (eds.),
  \emph{Proceedings of the 2024 Conference on Empirical Methods in Natural
  Language Processing}, pp.\  16143--16162, Miami, Florida, USA, November 2024.
  Association for Computational Linguistics.
\newblock \doi{10.18653/v1/2024.emnlp-main.903}.
\newblock URL \url{https://aclanthology.org/2024.emnlp-main.903/}.

\bibitem[Guo et~al.(2025)Guo, Yang, Zhang, Song, Zhang, Xu, Zhu, Ma, Wang, Bi,
  et~al.]{guo2025deepseek}
Daya Guo, Dejian Yang, Haowei Zhang, Junxiao Song, Ruoyu Zhang, Runxin Xu,
  Qihao Zhu, Shirong Ma, Peiyi Wang, Xiao Bi, et~al.
\newblock Deepseek-r1: Incentivizing reasoning capability in llms via
  reinforcement learning.
\newblock \emph{arXiv preprint arXiv:2501.12948}, 2025.

\bibitem[Habib et~al.(2023)Habib, Fourrier, Kydlíček, Wolf, and
  Tunstall]{lighteval}
Nathan Habib, Clémentine Fourrier, Hynek Kydlíček, Thomas Wolf, and Lewis
  Tunstall.
\newblock Lighteval: A lightweight framework for llm evaluation, 2023.
\newblock URL \url{https://github.com/huggingface/lighteval}.

\bibitem[H{\"a}gele et~al.(2024)H{\"a}gele, Bakouch, Kosson, Allal, Von~Werra,
  and Jaggi]{hagele2024scaling}
Alexander H{\"a}gele, Elie Bakouch, Atli Kosson, Loubna~Ben Allal, Leandro
  Von~Werra, and Martin Jaggi.
\newblock Scaling laws and compute-optimal training beyond fixed training
  durations.
\newblock \emph{arXiv preprint arXiv:2405.18392}, 2024.

\bibitem[Hardalov et~al.(2020)Hardalov, Mihaylov, Zlatkova, Dinkov, Koychev,
  and Nakov]{hardalov-etal-2020-exams}
Momchil Hardalov, Todor Mihaylov, Dimitrina Zlatkova, Yoan Dinkov, Ivan
  Koychev, and Preslav Nakov.
\newblock {EXAMS}: A multi-subject high school examinations dataset for
  cross-lingual and multilingual question answering.
\newblock In Bonnie Webber, Trevor Cohn, Yulan He, and Yang Liu (eds.),
  \emph{Proceedings of the 2020 Conference on Empirical Methods in Natural
  Language Processing (EMNLP)}, pp.\  5427--5444, Online, November 2020.
  Association for Computational Linguistics.
\newblock \doi{10.18653/v1/2020.emnlp-main.438}.
\newblock URL \url{https://aclanthology.org/2020.emnlp-main.438/}.

\bibitem[He et~al.(2024)He, Benhaim, Patra, Vaddamanu, Ahuja, Chopra,
  Chaudhary, Zhao, and Song]{he2024scalinglawsmultilinguallanguage}
Yifei He, Alon Benhaim, Barun Patra, Praneetha Vaddamanu, Sanchit Ahuja, Parul
  Chopra, Vishrav Chaudhary, Han Zhao, and Xia Song.
\newblock Scaling laws for multilingual language models, 2024.
\newblock URL \url{https://arxiv.org/abs/2410.12883}.

\bibitem[Held et~al.(2025)Held, Paranjape, Koura, Lewis, Zhang, and
  Mihaylov]{held2025optimizing}
William Held, Bhargavi Paranjape, Punit~Singh Koura, Mike Lewis, Frank Zhang,
  and Todor Mihaylov.
\newblock Optimizing pretraining data mixtures with llm-estimated utility.
\newblock \emph{arXiv preprint arXiv:2501.11747}, 2025.

\bibitem[Hendrycks et~al.(2021)Hendrycks, Burns, Basart, Zou, Mazeika, Song,
  and Steinhardt]{hendryckstest2021}
Dan Hendrycks, Collin Burns, Steven Basart, Andy Zou, Mantas Mazeika, Dawn
  Song, and Jacob Steinhardt.
\newblock Measuring massive multitask language understanding.
\newblock \emph{Proceedings of the International Conference on Learning
  Representations (ICLR)}, 2021.

\bibitem[Kim \& Lee(2024)Kim and Lee]{kim2024strategic}
Jisu Kim and Juhwan Lee.
\newblock Strategic data ordering: Enhancing large language model performance
  through curriculum learning.
\newblock \emph{arXiv preprint arXiv:2405.07490}, 2024.

\bibitem[Kim et~al.(2019)Kim, Petrov, Petrushkov, Khadivi, and
  Ney]{kim-etal-2019-pivot}
Yunsu Kim, Petre Petrov, Pavel Petrushkov, Shahram Khadivi, and Hermann Ney.
\newblock Pivot-based transfer learning for neural machine translation between
  non-{E}nglish languages.
\newblock In Kentaro Inui, Jing Jiang, Vincent Ng, and Xiaojun Wan (eds.),
  \emph{Proceedings of the 2019 Conference on Empirical Methods in Natural
  Language Processing and the 9th International Joint Conference on Natural
  Language Processing (EMNLP-IJCNLP)}, pp.\  866--876, Hong Kong, China,
  November 2019. Association for Computational Linguistics.
\newblock \doi{10.18653/v1/D19-1080}.
\newblock URL \url{https://aclanthology.org/D19-1080/}.

\bibitem[Kumar et~al.(2021)Kumar, Koehn, and Khudanpur]{kumar2021learning}
Gaurav Kumar, Philipp Koehn, and Sanjeev Khudanpur.
\newblock Learning policies for multilingual training of neural machine
  translation systems.
\newblock \emph{arXiv preprint arXiv:2103.06964}, 2021.

\bibitem[Li et~al.(2021)Li, Zhang, and He]{li2021curriculum}
Conglong Li, Minjia Zhang, and Yuxiong He.
\newblock Curriculum learning: A regularization method for efficient and stable
  billion-scale gpt model pre-training.
\newblock \emph{arXiv preprint arXiv:2108.06084}, 8:\penalty0 13, 2021.

\bibitem[Lin et~al.(2021{\natexlab{a}})Lin, Lee, Qiao, and
  Ren]{lin-etal-2021-common}
Bill~Yuchen Lin, Seyeon Lee, Xiaoyang Qiao, and Xiang Ren.
\newblock Common sense beyond {E}nglish: Evaluating and improving multilingual
  language models for commonsense reasoning.
\newblock In Chengqing Zong, Fei Xia, Wenjie Li, and Roberto Navigli (eds.),
  \emph{Proceedings of the 59th Annual Meeting of the Association for
  Computational Linguistics and the 11th International Joint Conference on
  Natural Language Processing (Volume 1: Long Papers)}, pp.\  1274--1287,
  Online, August 2021{\natexlab{a}}. Association for Computational Linguistics.
\newblock \doi{10.18653/v1/2021.acl-long.102}.
\newblock URL \url{https://aclanthology.org/2021.acl-long.102/}.

\bibitem[Lin et~al.(2021{\natexlab{b}})Lin, Mihaylov, Artetxe, Wang, Chen,
  Simig, Ott, Goyal, Bhosale, Du, Pasunuru, Shleifer, Koura, Chaudhary, O'Horo,
  Wang, Zettlemoyer, Kozareva, Diab, Stoyanov, and Li]{xi2022xstorycloze}
Xi~Victoria Lin, Todor Mihaylov, Mikel Artetxe, Tianlu Wang, Shuohui Chen,
  Daniel Simig, Myle Ott, Naman Goyal, Shruti Bhosale, Jingfei Du, Ramakanth
  Pasunuru, Sam Shleifer, Punit~Singh Koura, Vishrav Chaudhary, Brian O'Horo,
  Jeff Wang, Luke Zettlemoyer, Zornitsa Kozareva, Mona~T. Diab, Veselin
  Stoyanov, and Xian Li.
\newblock Few-shot learning with multilingual language models.
\newblock \emph{CoRR}, abs/2112.10668, 2021{\natexlab{b}}.
\newblock URL \url{https://arxiv.org/abs/2112.10668}.

\bibitem[Liu et~al.(2024)Liu, Zheng, Muennighoff, Zeng, Dou, Pang, Jiang, and
  Lin]{liu2024regmix}
Qian Liu, Xiaosen Zheng, Niklas Muennighoff, Guangtao Zeng, Longxu Dou, Tianyu
  Pang, Jing Jiang, and Min Lin.
\newblock Regmix: Data mixture as regression for language model pre-training.
\newblock \emph{arXiv preprint arXiv:2407.01492}, 2024.

\bibitem[Loshchilov(2017)]{loshchilov2017decoupled}
I~Loshchilov.
\newblock Decoupled weight decay regularization.
\newblock \emph{arXiv preprint arXiv:1711.05101}, 2017.

\bibitem[{Meta AI}(2025)]{meta2024llama4}
{Meta AI}.
\newblock Introducing llama 4: Advancing multimodal intelligence, 2025.
\newblock URL \url{https://ai.meta.com/blog/llama-4-multimodal-intelligence/}.

\bibitem[Mohammadshahi et~al.(2024)Mohammadshahi, Vamvas, and
  Sennrich]{mohammadshahi-etal-2024-investigating}
Alireza Mohammadshahi, Jannis Vamvas, and Rico Sennrich.
\newblock Investigating multi-pivot ensembling with massively multilingual
  machine translation models.
\newblock In Shabnam Tafreshi, Arjun Akula, Jo{\~a}o Sedoc, Aleksandr Drozd,
  Anna Rogers, and Anna Rumshisky (eds.), \emph{Proceedings of the Fifth
  Workshop on Insights from Negative Results in NLP}, pp.\  169--180, Mexico
  City, Mexico, June 2024. Association for Computational Linguistics.
\newblock \doi{10.18653/v1/2024.insights-1.19}.
\newblock URL \url{https://aclanthology.org/2024.insights-1.19/}.

\bibitem[Mostafazadeh et~al.(2017)Mostafazadeh, Roth, Louis, Chambers, and
  Allen]{mostafazadeh2017lsdsem}
Nasrin Mostafazadeh, Michael Roth, Annie Louis, Nathanael Chambers, and James
  Allen.
\newblock Lsdsem 2017 shared task: The story cloze test.
\newblock In \emph{Proceedings of the 2nd Workshop on Linking Models of
  Lexical, Sentential and Discourse-level Semantics}, pp.\  46--51, 2017.

\bibitem[Muennighoff et~al.(2022)Muennighoff, Wang, Sutawika, Roberts,
  Biderman, Scao, Bari, Shen, Yong, Schoelkopf, Tang, Radev, Aji, Almubarak,
  Albanie, Alyafeai, Webson, Raff, and Raffel]{muennighoff2022crosslingual}
Niklas Muennighoff, Thomas Wang, Lintang Sutawika, Adam Roberts, Stella
  Biderman, Teven~Le Scao, M~Saiful Bari, Sheng Shen, Zheng-Xin Yong, Hailey
  Schoelkopf, Xiangru Tang, Dragomir Radev, Alham~Fikri Aji, Khalid Almubarak,
  Samuel Albanie, Zaid Alyafeai, Albert Webson, Edward Raff, and Colin Raffel.
\newblock Crosslingual generalization through multitask finetuning, 2022.

\bibitem[Na{\"i}r et~al.(2024)Na{\"i}r, Yamani, Lhadj, and
  Baghdadi]{nair-etal-2024-curriculum}
Marwa Na{\"i}r, Kamel Yamani, Lynda Lhadj, and Riyadh Baghdadi.
\newblock Curriculum learning for small code language models.
\newblock In Xiyan Fu and Eve Fleisig (eds.), \emph{Proceedings of the 62nd
  Annual Meeting of the Association for Computational Linguistics (Volume 4:
  Student Research Workshop)}, pp.\  390--401, Bangkok, Thailand, August 2024.
  Association for Computational Linguistics.
\newblock ISBN 979-8-89176-097-4.
\newblock \doi{10.18653/v1/2024.acl-srw.44}.
\newblock URL \url{https://aclanthology.org/2024.acl-srw.44/}.

\bibitem[Penedo et~al.(2025)Penedo, Kydl{\'\i}{\v{c}}ek, Sabol{\v{c}}ec,
  Messmer, Foroutan, Kargaran, Raffel, Jaggi, Von~Werra, and
  Wolf]{penedo2024fineweb-2}
Guilherme Penedo, Hynek Kydl{\'\i}{\v{c}}ek, Vinko Sabol{\v{c}}ec, Bettina
  Messmer, Negar Foroutan, Amir~Hossein Kargaran, Colin Raffel, Martin Jaggi,
  Leandro Von~Werra, and Thomas Wolf.
\newblock Fineweb2: One pipeline to scale them all--adapting pre-training data
  processing to every language.
\newblock \emph{arXiv preprint arXiv:2506.20920}, 2025.

\bibitem[Pfeiffer et~al.(2022)Pfeiffer, Goyal, Lin, Li, Cross, Riedel, and
  Artetxe]{pfeiffer-etal-2022-lifting}
Jonas Pfeiffer, Naman Goyal, Xi~Lin, Xian Li, James Cross, Sebastian Riedel,
  and Mikel Artetxe.
\newblock Lifting the curse of multilinguality by pre-training modular
  transformers.
\newblock In Marine Carpuat, Marie-Catherine de~Marneffe, and Ivan~Vladimir
  Meza~Ruiz (eds.), \emph{Proceedings of the 2022 Conference of the North
  American Chapter of the Association for Computational Linguistics: Human
  Language Technologies}, pp.\  3479--3495, Seattle, United States, July 2022.
  Association for Computational Linguistics.
\newblock \doi{10.18653/v1/2022.naacl-main.255}.
\newblock URL \url{https://aclanthology.org/2022.naacl-main.255/}.

\bibitem[Ponti et~al.(2020)Ponti, Glavaš, Majewska, Liu, Vulić, and
  Korhonen]{ponti2020xcopamultilingualdatasetcausal}
Edoardo~Maria Ponti, Goran Glavaš, Olga Majewska, Qianchu Liu, Ivan Vulić,
  and Anna Korhonen.
\newblock Xcopa: A multilingual dataset for causal commonsense reasoning, 2020.
\newblock URL \url{https://arxiv.org/abs/2005.00333}.

\bibitem[Raffel et~al.(2019)Raffel, Shazeer, Roberts, Lee, Narang, Matena,
  Zhou, Li, and Liu]{2019t5}
Colin Raffel, Noam Shazeer, Adam Roberts, Katherine Lee, Sharan Narang, Michael
  Matena, Yanqi Zhou, Wei Li, and Peter~J. Liu.
\newblock Exploring the limits of transfer learning with a unified text-to-text
  transformer.
\newblock \emph{arXiv e-prints}, 2019.

\bibitem[Ranaldi et~al.(2024)Ranaldi, Pucci, and
  Freitas]{ranaldi-etal-2024-language}
Leonardo Ranaldi, Giulia Pucci, and Andr{\`e} Freitas.
\newblock Does the \textit{Order} matter? {C}urriculum learning over languages.
\newblock In Nicoletta Calzolari, Min-Yen Kan, Veronique Hoste, Alessandro
  Lenci, Sakriani Sakti, and Nianwen Xue (eds.), \emph{Proceedings of the 2024
  Joint International Conference on Computational Linguistics, Language
  Resources and Evaluation (LREC-COLING 2024)}, pp.\  5212--5220, Torino,
  Italia, May 2024. ELRA and ICCL.
\newblock URL \url{https://aclanthology.org/2024.lrec-main.464/}.

\bibitem[Romanou et~al.(2024)Romanou, Foroutan, Sotnikova, Chen, Nelaturu,
  Singh, Maheshwary, Altomare, Haggag, Amayuelas, et~al.]{romanou2024include}
Angelika Romanou, Negar Foroutan, Anna Sotnikova, Zeming Chen, Sree~Harsha
  Nelaturu, Shivalika Singh, Rishabh Maheshwary, Micol Altomare, Mohamed~A
  Haggag, Alfonso Amayuelas, et~al.
\newblock Include: Evaluating multilingual language understanding with regional
  knowledge.
\newblock \emph{arXiv preprint arXiv:2411.19799}, 2024.

\bibitem[Sakaguchi et~al.(2021)Sakaguchi, Bras, Bhagavatula, and
  Choi]{ai2:winogrande}
Keisuke Sakaguchi, Ronan~Le Bras, Chandra Bhagavatula, and Yejin Choi.
\newblock Winogrande: An adversarial winograd schema challenge at scale.
\newblock \emph{Communications of the ACM}, 64\penalty0 (9):\penalty0 99--106,
  2021.

\bibitem[Talmor et~al.(2019)Talmor, Herzig, Lourie, and
  Berant]{Talmor2019commonsenseqaaq}
Alon Talmor, Jonathan Herzig, Nicholas Lourie, and Jonathan Berant.
\newblock Commonsenseqa: A question answering challenge targeting commonsense
  knowledge.
\newblock In \emph{Proceedings of the 2019 Conference of the North {A}merican
  Chapter of the Association for Computational Linguistics: Human Language
  Technologies, Volume 1 (Long and Short Papers)}, pp.\  4149--4158,
  Minneapolis, Minnesota, 2019. Association for Computational Linguistics.
\newblock \doi{10.18653/v1/N19-1421}.
\newblock URL \url{https://www.aclweb.org/anthology/N19-1421}.

\bibitem[Team et~al.(2025)Team, Kamath, Ferret, Pathak, Vieillard, Merhej,
  Perrin, Matejovicova, Ramé, Rivière, Rouillard, Mesnard, Cideron, bastien
  Grill, Ramos, Yvinec, Casbon, Pot, Penchev, Liu, Visin, Kenealy, Beyer, Zhai,
  Tsitsulin, Busa-Fekete, Feng, Sachdeva, Coleman, Gao, Mustafa, Barr,
  Parisotto, Tian, Eyal, Cherry, Peter, Sinopalnikov, Bhupatiraju, Agarwal,
  Kazemi, Malkin, Kumar, Vilar, Brusilovsky, Luo, Steiner, Friesen, Sharma,
  Sharma, Gilady, Goedeckemeyer, Saade, Feng, Kolesnikov, Bendebury, Abdagic,
  Vadi, György, Pinto, Das, Bapna, Miech, Yang, Paterson, Shenoy, Chakrabarti,
  Piot, Wu, Shahriari, Petrini, Chen, Lan, Choquette-Choo, Carey, Brick,
  Deutsch, Eisenbud, Cattle, Cheng, Paparas, Sreepathihalli, Reid, Tran, Zelle,
  Noland, Huizenga, Kharitonov, Liu, Amirkhanyan, Cameron, Hashemi,
  Klimczak-Plucińska, Singh, Mehta, Lehri, Hazimeh, Ballantyne, Szpektor,
  Nardini, Pouget-Abadie, Chan, Stanton, Wieting, Lai, Orbay, Fernandez,
  Newlan, yeong Ji, Singh, Black, Yu, Hui, Vodrahalli, Greff, Qiu, Valentine,
  Coelho, Ritter, Hoffman, Watson, Chaturvedi, Moynihan, Ma, Babar, Noy, Byrd,
  Roy, Momchev, Chauhan, Sachdeva, Bunyan, Botarda, Caron, Rubenstein,
  Culliton, Schmid, Sessa, Xu, Stanczyk, Tafti, Shivanna, Wu, Pan, Rokni,
  Willoughby, Vallu, Mullins, Jerome, Smoot, Girgin, Iqbal, Reddy, Sheth,
  Põder, Bhatnagar, Panyam, Eiger, Zhang, Liu, Yacovone, Liechty, Kalra, Evci,
  Misra, Roseberry, Feinberg, Kolesnikov, Han, Kwon, Chen, Chow, Zhu, Wei,
  Egyed, Cotruta, Giang, Kirk, Rao, Black, Babar, Lo, Moreira, Martins,
  Sanseviero, Gonzalez, Gleicher, Warkentin, Mirrokni, Senter, Collins, Barral,
  Ghahramani, Hadsell, Matias, Sculley, Petrov, Fiedel, Shazeer, Vinyals, Dean,
  Hassabis, Kavukcuoglu, Farabet, Buchatskaya, Alayrac, Anil, Dmitry, Lepikhin,
  Borgeaud, Bachem, Joulin, Andreev, Hardin, Dadashi, and
  Hussenot]{gemmateam2025gemma3technicalreport}
Gemma Team, Aishwarya Kamath, Johan Ferret, Shreya Pathak, Nino Vieillard,
  Ramona Merhej, Sarah Perrin, Tatiana Matejovicova, Alexandre Ramé, Morgane
  Rivière, Louis Rouillard, Thomas Mesnard, Geoffrey Cideron, Jean bastien
  Grill, Sabela Ramos, Edouard Yvinec, Michelle Casbon, Etienne Pot, Ivo
  Penchev, Gaël Liu, Francesco Visin, Kathleen Kenealy, Lucas Beyer, Xiaohai
  Zhai, Anton Tsitsulin, Robert Busa-Fekete, Alex Feng, Noveen Sachdeva,
  Benjamin Coleman, Yi~Gao, Basil Mustafa, Iain Barr, Emilio Parisotto, David
  Tian, Matan Eyal, Colin Cherry, Jan-Thorsten Peter, Danila Sinopalnikov,
  Surya Bhupatiraju, Rishabh Agarwal, Mehran Kazemi, Dan Malkin, Ravin Kumar,
  David Vilar, Idan Brusilovsky, Jiaming Luo, Andreas Steiner, Abe Friesen,
  Abhanshu Sharma, Abheesht Sharma, Adi~Mayrav Gilady, Adrian Goedeckemeyer,
  Alaa Saade, Alex Feng, Alexander Kolesnikov, Alexei Bendebury, Alvin Abdagic,
  Amit Vadi, András György, André~Susano Pinto, Anil Das, Ankur Bapna,
  Antoine Miech, Antoine Yang, Antonia Paterson, Ashish Shenoy, Ayan
  Chakrabarti, Bilal Piot, Bo~Wu, Bobak Shahriari, Bryce Petrini, Charlie Chen,
  Charline~Le Lan, Christopher~A. Choquette-Choo, CJ~Carey, Cormac Brick,
  Daniel Deutsch, Danielle Eisenbud, Dee Cattle, Derek Cheng, Dimitris Paparas,
  Divyashree~Shivakumar Sreepathihalli, Doug Reid, Dustin Tran, Dustin Zelle,
  Eric Noland, Erwin Huizenga, Eugene Kharitonov, Frederick Liu, Gagik
  Amirkhanyan, Glenn Cameron, Hadi Hashemi, Hanna Klimczak-Plucińska, Harman
  Singh, Harsh Mehta, Harshal~Tushar Lehri, Hussein Hazimeh, Ian Ballantyne,
  Idan Szpektor, Ivan Nardini, Jean Pouget-Abadie, Jetha Chan, Joe Stanton,
  John Wieting, Jonathan Lai, Jordi Orbay, Joseph Fernandez, Josh Newlan,
  Ju~yeong Ji, Jyotinder Singh, Kat Black, Kathy Yu, Kevin Hui, Kiran
  Vodrahalli, Klaus Greff, Linhai Qiu, Marcella Valentine, Marina Coelho,
  Marvin Ritter, Matt Hoffman, Matthew Watson, Mayank Chaturvedi, Michael
  Moynihan, Min Ma, Nabila Babar, Natasha Noy, Nathan Byrd, Nick Roy, Nikola
  Momchev, Nilay Chauhan, Noveen Sachdeva, Oskar Bunyan, Pankil Botarda, Paul
  Caron, Paul~Kishan Rubenstein, Phil Culliton, Philipp Schmid, Pier~Giuseppe
  Sessa, Pingmei Xu, Piotr Stanczyk, Pouya Tafti, Rakesh Shivanna, Renjie Wu,
  Renke Pan, Reza Rokni, Rob Willoughby, Rohith Vallu, Ryan Mullins, Sammy
  Jerome, Sara Smoot, Sertan Girgin, Shariq Iqbal, Shashir Reddy, Shruti Sheth,
  Siim Põder, Sijal Bhatnagar, Sindhu~Raghuram Panyam, Sivan Eiger, Susan
  Zhang, Tianqi Liu, Trevor Yacovone, Tyler Liechty, Uday Kalra, Utku Evci,
  Vedant Misra, Vincent Roseberry, Vlad Feinberg, Vlad Kolesnikov, Woohyun Han,
  Woosuk Kwon, Xi~Chen, Yinlam Chow, Yuvein Zhu, Zichuan Wei, Zoltan Egyed,
  Victor Cotruta, Minh Giang, Phoebe Kirk, Anand Rao, Kat Black, Nabila Babar,
  Jessica Lo, Erica Moreira, Luiz~Gustavo Martins, Omar Sanseviero, Lucas
  Gonzalez, Zach Gleicher, Tris Warkentin, Vahab Mirrokni, Evan Senter, Eli
  Collins, Joelle Barral, Zoubin Ghahramani, Raia Hadsell, Yossi Matias,
  D.~Sculley, Slav Petrov, Noah Fiedel, Noam Shazeer, Oriol Vinyals, Jeff Dean,
  Demis Hassabis, Koray Kavukcuoglu, Clement Farabet, Elena Buchatskaya,
  Jean-Baptiste Alayrac, Rohan Anil, Dmitry, Lepikhin, Sebastian Borgeaud,
  Olivier Bachem, Armand Joulin, Alek Andreev, Cassidy Hardin, Robert Dadashi,
  and Léonard Hussenot.
\newblock Gemma 3 technical report, 2025.
\newblock URL \url{https://arxiv.org/abs/2503.19786}.

\bibitem[Tikhonov \& Ryabinin(2021)Tikhonov and Ryabinin]{tikhonov2021heads}
Alexey Tikhonov and Max Ryabinin.
\newblock It's all in the heads: Using attention heads as a baseline for
  cross-lingual transfer in commonsense reasoning, 2021.

\bibitem[Touvron et~al.(2023)Touvron, Lavril, Izacard, Martinet, Lachaux,
  Lacroix, Rozi{\`e}re, Goyal, Hambro, Azhar, et~al.]{touvron2023llama}
Hugo Touvron, Thibaut Lavril, Gautier Izacard, Xavier Martinet, Marie-Anne
  Lachaux, Timoth{\'e}e Lacroix, Baptiste Rozi{\`e}re, Naman Goyal, Eric
  Hambro, Faisal Azhar, et~al.
\newblock Llama: Open and efficient foundation language models.
\newblock \emph{arXiv preprint arXiv:2302.13971}, 2023.

\bibitem[{\"U}st{\"u}n et~al.(2024){\"U}st{\"u}n, Aryabumi, Yong, Ko,
  D{'}souza, Onilude, Bhandari, Singh, Ooi, Kayid, Vargus, Blunsom, Longpre,
  Muennighoff, Fadaee, Kreutzer, and Hooker]{ustun-etal-2024-aya}
Ahmet {\"U}st{\"u}n, Viraat Aryabumi, Zheng Yong, Wei-Yin Ko, Daniel D{'}souza,
  Gbemileke Onilude, Neel Bhandari, Shivalika Singh, Hui-Lee Ooi, Amr Kayid,
  Freddie Vargus, Phil Blunsom, Shayne Longpre, Niklas Muennighoff, Marzieh
  Fadaee, Julia Kreutzer, and Sara Hooker.
\newblock Aya model: An instruction finetuned open-access multilingual language
  model.
\newblock In Lun-Wei Ku, Andre Martins, and Vivek Srikumar (eds.),
  \emph{Proceedings of the 62nd Annual Meeting of the Association for
  Computational Linguistics (Volume 1: Long Papers)}, pp.\  15894--15939,
  Bangkok, Thailand, August 2024. Association for Computational Linguistics.
\newblock \doi{10.18653/v1/2024.acl-long.845}.
\newblock URL \url{https://aclanthology.org/2024.acl-long.845/}.

\bibitem[Vaswani(2017)]{vaswani2017attention}
A~Vaswani.
\newblock Attention is all you need.
\newblock \emph{Advances in Neural Information Processing Systems}, 2017.

\bibitem[Wang et~al.(2020)Wang, Lipton, and Tsvetkov]{wang-etal-2020-negative}
Zirui Wang, Zachary~C. Lipton, and Yulia Tsvetkov.
\newblock On negative interference in multilingual models: Findings and a
  meta-learning treatment.
\newblock In Bonnie Webber, Trevor Cohn, Yulan He, and Yang Liu (eds.),
  \emph{Proceedings of the 2020 Conference on Empirical Methods in Natural
  Language Processing (EMNLP)}, pp.\  4438--4450, Online, November 2020.
  Association for Computational Linguistics.
\newblock \doi{10.18653/v1/2020.emnlp-main.359}.
\newblock URL \url{https://aclanthology.org/2020.emnlp-main.359/}.

\bibitem[Xie et~al.(2023)Xie, Santurkar, Ma, and Liang]{xie2023data}
Sang~Michael Xie, Shibani Santurkar, Tengyu Ma, and Percy~S Liang.
\newblock Data selection for language models via importance resampling.
\newblock \emph{Advances in Neural Information Processing Systems},
  36:\penalty0 34201--34227, 2023.

\bibitem[Xie et~al.(2024)Xie, Pham, Dong, Du, Liu, Lu, Liang, Le, Ma, and
  Yu]{xie2024doremi}
Sang~Michael Xie, Hieu Pham, Xuanyi Dong, Nan Du, Hanxiao Liu, Yifeng Lu,
  Percy~S Liang, Quoc~V Le, Tengyu Ma, and Adams~Wei Yu.
\newblock Doremi: Optimizing data mixtures speeds up language model
  pretraining.
\newblock \emph{Advances in Neural Information Processing Systems}, 36, 2024.

\bibitem[Xue et~al.(2021)Xue, Constant, Roberts, Kale, Al-Rfou, Siddhant,
  Barua, and Raffel]{xue-etal-2021-mt5}
Linting Xue, Noah Constant, Adam Roberts, Mihir Kale, Rami Al-Rfou, Aditya
  Siddhant, Aditya Barua, and Colin Raffel.
\newblock m{T}5: A massively multilingual pre-trained text-to-text transformer.
\newblock In Kristina Toutanova, Anna Rumshisky, Luke Zettlemoyer, Dilek
  Hakkani-Tur, Iz~Beltagy, Steven Bethard, Ryan Cotterell, Tanmoy Chakraborty,
  and Yichao Zhou (eds.), \emph{Proceedings of the 2021 Conference of the North
  American Chapter of the Association for Computational Linguistics: Human
  Language Technologies}, pp.\  483--498, Online, June 2021. Association for
  Computational Linguistics.
\newblock \doi{10.18653/v1/2021.naacl-main.41}.
\newblock URL \url{https://aclanthology.org/2021.naacl-main.41/}.

\bibitem[Yang et~al.(2025)Yang, Li, Yang, Zhang, Hui, Zheng, Yu, Gao, Huang,
  Lv, et~al.]{yang2025qwen3}
An~Yang, Anfeng Li, Baosong Yang, Beichen Zhang, Binyuan Hui, Bo~Zheng, Bowen
  Yu, Chang Gao, Chengen Huang, Chenxu Lv, et~al.
\newblock Qwen3 technical report.
\newblock \emph{arXiv preprint arXiv:2505.09388}, 2025.

\bibitem[Yoo et~al.(2024)Yoo, Park, Yun, Oh, and Lee]{yoo2024code}
Haneul Yoo, Cheonbok Park, Sangdoo Yun, Alice Oh, and Hwaran Lee.
\newblock Code-switching curriculum learning for multilingual transfer in llms.
\newblock \emph{arXiv preprint arXiv:2411.02460}, 2024.

\bibitem[Zhang et~al.(2021)Zhang, Meng, Tong, and
  Zhou]{zhang-etal-2021-competence-based}
Mingliang Zhang, Fandong Meng, Yunhai Tong, and Jie Zhou.
\newblock Competence-based curriculum learning for multilingual machine
  translation.
\newblock In Marie-Francine Moens, Xuanjing Huang, Lucia Specia, and Scott
  Wen-tau Yih (eds.), \emph{Findings of the Association for Computational
  Linguistics: EMNLP 2021}, pp.\  2481--2493, Punta Cana, Dominican Republic,
  November 2021. Association for Computational Linguistics.
\newblock \doi{10.18653/v1/2021.findings-emnlp.212}.
\newblock URL \url{https://aclanthology.org/2021.findings-emnlp.212/}.

\bibitem[Zhang et~al.(2023)Zhang, Aljunied, Gao, Chia, and
  Bing]{zhang2023m3exam}
Wenxuan Zhang, Mahani Aljunied, Chang Gao, Yew~Ken Chia, and Lidong Bing.
\newblock M3exam: A multilingual, multimodal, multilevel benchmark for
  examining large language models.
\newblock \emph{Advances in Neural Information Processing Systems},
  36:\penalty0 5484--5505, 2023.

\bibitem[Zhang et~al.(2024)Zhang, Lee, Fang, Yu, Jia, Jiang, and
  Barbieri]{zhang-etal-2024-plug}
Zhihan Zhang, Dong-Ho Lee, Yuwei Fang, Wenhao Yu, Mengzhao Jia, Meng Jiang, and
  Francesco Barbieri.
\newblock {PLUG}: Leveraging pivot language in cross-lingual instruction
  tuning.
\newblock In Lun-Wei Ku, Andre Martins, and Vivek Srikumar (eds.),
  \emph{Proceedings of the 62nd Annual Meeting of the Association for
  Computational Linguistics (Volume 1: Long Papers)}, pp.\  7025--7046,
  Bangkok, Thailand, August 2024. Association for Computational Linguistics.
\newblock \doi{10.18653/v1/2024.acl-long.379}.
\newblock URL \url{https://aclanthology.org/2024.acl-long.379/}.

\bibitem[Zhao et~al.(2024{\natexlab{a}})Zhao, Hu, Guo, Sui, Wu, Deng, Zhao,
  Qin, Che, and Liu]{zhao2024lens}
Weixiang Zhao, Yulin Hu, Jiahe Guo, Xingyu Sui, Tongtong Wu, Yang Deng, Yanyan
  Zhao, Bing Qin, Wanxiang Che, and Ting Liu.
\newblock Lens: Rethinking multilingual enhancement for large language models.
\newblock \emph{arXiv preprint arXiv:2410.04407}, 2024{\natexlab{a}}.

\bibitem[Zhao et~al.(2024{\natexlab{b}})Zhao, Du, Ding, Xiong, Sun, Jun, Liu,
  and Qin]{zhao-etal-2024-deciphering}
Yang Zhao, Li~Du, Xiao Ding, Kai Xiong, Zhouhao Sun, Shi Jun, Ting Liu, and
  Bing Qin.
\newblock Deciphering the impact of pretraining data on large language models
  through machine unlearning.
\newblock In Lun-Wei Ku, Andre Martins, and Vivek Srikumar (eds.),
  \emph{Findings of the Association for Computational Linguistics: ACL 2024},
  pp.\  9386--9406, Bangkok, Thailand, August 2024{\natexlab{b}}. Association
  for Computational Linguistics.
\newblock \doi{10.18653/v1/2024.findings-acl.559}.
\newblock URL \url{https://aclanthology.org/2024.findings-acl.559/}.

\bibitem[Zhou et~al.(2021)Zhou, Ding, Duh, Watanabe, Sasano, and
  Takeda]{zhou-etal-2021-self}
Lei Zhou, Liang Ding, Kevin Duh, Shinji Watanabe, Ryohei Sasano, and Koichi
  Takeda.
\newblock Self-guided curriculum learning for neural machine translation.
\newblock In Marcello Federico, Alex Waibel, Marta~R. Costa-juss{\`a}, Jan
  Niehues, Sebastian Stuker, and Elizabeth Salesky (eds.), \emph{Proceedings of
  the 18th International Conference on Spoken Language Translation (IWSLT
  2021)}, pp.\  206--214, Bangkok, Thailand (online), August 2021. Association
  for Computational Linguistics.
\newblock \doi{10.18653/v1/2021.iwslt-1.25}.
\newblock URL \url{https://aclanthology.org/2021.iwslt-1.25/}.

\bibitem[Zou et~al.(2022)Zou, Saeedi, and Carl]{zou2022investigating}
Longhui Zou, Ali Saeedi, and Michael Carl.
\newblock Investigating the impact of different pivot languages on translation
  quality.
\newblock In \emph{Proceedings of the 15th biennial conference of the
  Association for Machine Translation in the Americas (Workshop 1: Empirical
  Translation Process Research)}, pp.\  15--28, 2022.

\end{thebibliography}
\bibliographystyle{iclr2026_conference}

\appendix

\section{Language Model Training}
\label{appendix:training_setup}
We train our models using SwissAI's fork of Hugging Face's Nanotron repository.\footnote{The codebase: \href{https://github.com/swiss-ai/nanotron-multilingual}{https://github.com/swiss-ai/nanotron-multilingual}}
Here we provide details about training the language models used in our experiments.

\begin{table}[h]
\centering
\setlength{\tabcolsep}{4pt} 
\begin{tabular}{lcccccc}
\toprule
\textbf{Model} & \textbf{Arch.} & \textbf{Layers} & \textbf{Hidden} & \textbf{Attn. Heads} & \textbf{RoPE $\theta$} & \textbf{Vocab} \\
\midrule
$1.1$B & LLaMA & 24 & 1536 & 16 & 500{,}000 & 131{,}000 \\
$3$B   & LLaMA & 28 & 2496 & 24 & 500{,}000 & 131{,}000 \\
\bottomrule
\end{tabular}
\caption{Overview of the architectural configurations for different model sizes.}
\label{tab:model-configs}
\end{table}

Our experiments focus on models with $1.1$ and 3 billion parameters ($1.1$B and $3$B).
All models follow the LLaMA architecture~\cite{touvron2023llama}.
The model size is determined by adjusting the number of layers,
hidden sizes, and the number of attention heads (Details in Table~\ref{tab:model-configs}).

\subsection{Training Hyperparameters}
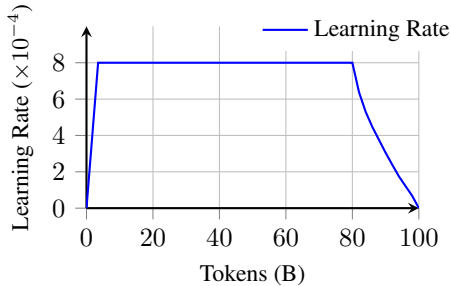
\begin{figure}[htbp]
    \centering
    \begin{tikzpicture}
    \begin{axis}[
        width=6cm,
        height=4cm,
        xlabel={Tokens (B)},
        ylabel={Learning Rate (\( \times 10^{-4} \))},
        ymin=0, ymax=10,
        xmin=0, xmax=100,
        legend style={draw=none, at={(0.5,1.1)}, anchor=north west, font=\footnotesize},
        grid=both,
        thick,
        enlargelimits=false,
        xtick={0,20,...,100},
        ytick={0,2,...,8},
        xlabel style={font=\small},
        ylabel style={font=\small},
        axis lines=left,
        tick align=outside
    ]
    
    \addplot[
        blue,
        thick
    ] coordinates {
    (0,   0)
    (3.5,   8)
    (80,  8)
    (82,  6.38)
    (84,  5.30)
    (86,  4.47)
    (88,  3.76)
    (90,  3.05)
    (92,  2.39)
    (94,  1.75)
    (96,  1.21)
    (98,  0.68)
    (100, 0)
    };
    
    \addlegendentry{Learning Rate}
    
    \end{axis}
    \end{tikzpicture}
    \caption{Learning rate schedule over tokens with warmup and decay.}
    \label{fig:lr_schedule}
\end{figure}

We train our models using HuggingFace's Nanotron trainer.\footnote{\href{https://github.com/huggingface/nanotron}{https://github.com/huggingface/nanotron}} 
The key training hyperparameters are as follows:
\begin{itemize}
    \item \textbf{Learning Rate.} We use a learning rate of $8 \times 10^{-4}$ with linear warmup over the first 4\% of training. A ``1-sqrt'' decay schedule~\citep{hagele2024scaling} is applied during the final 20\%, as shown in Figure~\ref{fig:lr_schedule}.
    \item \textbf{Optimizer.} All experiments use AdamW with $\beta = (0.9, 0.95)$~\citep{loshchilov2017decoupled}.
    \item \textbf{Weight Decay.} We set the weight decay parameter to $\lambda = 0.1$ for regularization.
    \item \textbf{Batch Size.} The micro-batch size is fixed at 5 across all runs.
\end{itemize}

\subsection{Hardware Setup}
We train our models on CSCS's Alps cluster~\footnote{More information on the cluster here: \href{https://www.cscs.ch/computers/alps}{https://www.cscs.ch/computers/alps}} containing nodes with 4 NVIDIA Grace-Hopper H100 GPUs (96 GB memory each).
\begin{itemize}
    \item \textbf{$1.1$B models.} We train our $1.1$B models on 22 nodes (or 88 GPUs) over around 15h per 100B tokens. This gives a global batch size of 440 examples.
    \item \textbf{3B models.} We train our $3$B models on 64 nodes (or 256 GPUs) for around 18h per $100$B tokens. Therefore our runs have a global batch size of 640 examples.
\end{itemize}

\subsection{Sampling Methods}
\label{appendix:sampling_methods}
Let $\mathcal{L}$ be the set of languages in the dataset, and let $\pi^{\text{natural}} \in \Delta_{|\mathcal{L}|}$ represent the natural distribution of these languages, defined as:
\[
\pi_l^{\text{natural}} = \frac{\omega_l}{\sum_{l' \in \mathcal{L}} \omega_{l'}}
\]
where $\omega_l$ denotes the number of words (or tokens) for language $l$ in the dataset. In this work, we use the number of words as a proxy for language frequency, a common practice when presenting statistics for highly multilingual datasets~\cite{penedo2024fineweb-2}.
We implement the following sampling strategies:
\begin{itemize}
    \item \textbf{Natural Sampling.} This method samples according to the natural distribution $\pi^{\text{natural}}$, directly reflecting language frequencies in the dataset. Typically, this distribution is highly imbalanced, with a few languages dominating the cumulative share of data.

    \item \textbf{Temperature Sampling.} This method adjusts the natural distribution using a temperature parameter $\tau$ to create a less skewed distribution:
    \[
    \pi_l^{\text{temp}, \tau} = \frac{\omega_l^{1/\tau}}{\sum_{l' \in \mathcal{L}} \omega_{l'}^{1/\tau}}
    \]
    By tuning $\tau$, the distribution can be shifted towards uniformity, thereby reducing imbalance among languages.
\end{itemize}

Figures~\ref{tab:token_distribution_en_xglm_fixed_total} and \ref{tab:token_distribution_en_xglm_fixed_multilingual} present the training data distribution for experiments described in Section~\ref{sec:english}.

\section{Benchmark Setup}
\label{appendix:benchmark_setup}
We evaluate our models using HuggingFace's Lighteval codebase~\citep{lighteval}.\footnote{\href{https://huggingface.co/docs/lighteval/en/index}{https://huggingface.co/docs/lighteval/en/index}}

\subsection{Benchmarks}
We select 10 standard multilingual benchmarks to evaluate our models on various downstream tasks.

\begin{itemize}
    \item \textbf{Belebele}: A multilingual reading comprehension dataset containing passages and corresponding questions in many languages. It evaluates models’ ability to understand text and answer related questions~\citep{bandarkar-etal-2024-belebele}.
    \item \textbf{XCodah}: A multilingual adaptation of CODAH for adversarially-authored commonsense reasoning tasks, testing robustness in natural language understanding~\citep{lin-etal-2021-common, Chen2019CODAHAA}.
    \item \textbf{XCSQA}: A multilingual version of CommonsenseQA, consisting of multiple-choice questions that require reasoning about everyday concepts and their relations~\citep{lin-etal-2021-common, Talmor2019commonsenseqaaq}.
    \item \textbf{XCOPA}: A multilingual adaptation of the COPA dataset for evaluating cross-lingual causal commonsense reasoning, covering multiple languages to test reasoning transfer across linguistic boundaries~\citep{ponti2020xcopamultilingualdatasetcausal}.
    \item \textbf{XStoryCloze}: A multilingual extension of the StoryCloze Test, where models must choose the most coherent ending to short narratives, testing story comprehension and commonsense reasoning~\citep{mostafazadeh2017lsdsem, xi2022xstorycloze}.
    \item \textbf{XWinogrande}: A multilingual version of WinoGrande, containing sentences with ambiguous pronouns. It measures models’ ability to resolve coreference using contextual and commonsense cues~\citep{ai2:winogrande, muennighoff2022crosslingual, tikhonov2021heads}.
    \item \textbf{MMMLU}: A multilingual adaptation of MMLU, evaluating model performance across a wide spectrum of tasks and domains~\citep{hendryckstest2021, dac2023okapi}.
    \item \textbf{INCLUDE}: A large-scale benchmark covering 44 languages, designed to evaluate multilingual LLMs in realistic language environments with a focus on knowledge and reasoning~\citep{romanou2024include}.
    \item \textbf{Exams}: A benchmark of standardized test questions across subjects and educational levels, used to assess reasoning and problem-solving abilities in exam-like conditions~\citep{hardalov-etal-2020-exams}.
    \item \textbf{M3Exams}: A multilingual exam-style benchmark that extends Exams across different languages, subjects, and difficulty levels~\citep{zhang2023m3exam}.
\end{itemize}

\subsection{Aggregations}
We aggregate benchmark results to compute a language-specific score for each model. Let $\mathcal{T}_l$ be the set of benchmarks (or tasks) containing a split for language $l$. The aggregated score for a model $m$ per language $l$ is defined as:
\[
s_l^m = \frac{1}{|\mathcal{T}_l|}\sum_{t \in \mathcal{T}_l} s_{t,l}^m
\]
where $s_l^m$ is the score of a model $m$ on the split $l$ of a task $t$.
To mitigate biases arising from varying numbers of benchmarks per language, we compute a language-specific random baseline $\zeta_l$. This baseline helps assess whether a given aggregated score significantly outperforms random predictions. Specifically, we calculate the random baseline for each language as the average of the individual random baselines across all tasks that include language $l$:
\[
\zeta_l = \frac{1}{|\mathcal{T}_l|}\sum_{t \in \mathcal{T}_l} \zeta_t
\]

\section{Pivot Ablation}
\label{appendix:pivot_ablation}

Table~\ref{tab:en_pivot_lang_metadata} presents the languages included in the experiments described in Section~\ref{sec:english}.
The set of languages analyzed in the experiments of Section~\ref{sec:family} is listed in Table~\ref{tab:cyrillic_pivot_lang_metadata}.

\begin{table}[htbp]
\centering
\resizebox{\linewidth}{!}{
\begin{tabular}{lll}
\toprule
\textbf{Language} & \textbf{Language Family} & \textbf{Script}\\
\midrule
Arabic           & Afro-Asiatic (Semitic) & Perso-Arabic  \\
Bulgarian        & Indo-European (Slavic) & Cyrillic\\
Bengali          & Indo-European (Indo-Aryan) & Bengali\\
Catalan          & Indo-European (Romance) & Latin\\
German           & Indo-European (Germanic)  & Latin \\
Greek            & Indo-European (Hellenic) & Greek\\
English          & Indo-European (Germanic) & Latin\\
Spanish          & Indo-European (Romance)  & Latin \\
Estonian         & Uralic (Finnic) & Latin\\
Basque           & Language Isolate & Latin\\
Persian (Farsi)  & Indo-European (Iranian) & Perso-Arabic  \\
Finnish          & Uralic (Finnic)  & Latin \\
French           & Indo-European (Romance)  & Latin \\
Hindi            & Indo-European (Indo-Aryan) & Devanagari\\
Haitian Creole   & Creole (French-based) & Latin\\
Indonesian       & Austronesian & Latin\\
Italian          & Indo-European (Romance)  & Latin \\
Japanese         & Japonic & Kanji \& Kana (CJK)\\
Korean           & Koreanic & Hangugeo (CJK) \\
Burmese          & Sino-Tibetan & Burmese\\
Portuguese       & Indo-European (Romance)  & Latin \\
Russian          & Indo-European (Slavic) & Cyrillic\\
Swahili          & Niger-Congo (Bantu)  & Latin \\
Tamil            & Dravidian & Tamil\\
Telugu           & Dravidian & Telugu (Brahmic)\\
Thai             & Kra–Dai (Tai) & Thai\\
Turkish          & Turkic  & Latin \\
Urdu             & Indo-European (Indo-Aryan) & Perso-Arabic   \\
Vietnamese       & Austroasiatic &  Vietnamese (Latin-based)\\
Chinese (Mandarin) & Sino-Tibetan & Hanzi (CJK)\\
\bottomrule
\end{tabular}
}
\caption{Languages used in experiments discussed in Section~\ref{sec:english}.}
\label{tab:en_pivot_lang_metadata}
\end{table}
\begin{table}
\centering
\resizebox{0.75\linewidth}{!}{
\begin{tabular}{lll}
\toprule
\textbf{Language} & \textbf{Language Family} & \textbf{Script}\\
\midrule
English          & Indo-European (Germanic) & Latin \\
Russian          & Indo-European (Slavic) & Cyrillic \\
Ukrainian        & Indo-European (Slavic) & Cyrillic \\
Belarusian       & Indo-European (Slavic) & Cyrillic \\
Serbian          & Indo-European (Slavic) & Cyrillic \\
Macedonian       & Indo-European (Slavic) & Cyrillic \\
Bulgarian        & Indo-European (Slavic) & Cyrillic \\
Polish           & Indo-European (Slavic) & Latin \\
Czech            & Indo-European (Slavic) & Latin\\
Slovak           & Indo-European (Slavic) & Latin \\
Tajik            & Indo-European (Iranian)& Cyrillic \\
Uzbek            & Turkic    & Cyrillic \\
Kyrgyz           & Turkic    & Cyrillic\\
Kazakh           & Turkic    & Cyrillic\\
Mongolian        &  Mongolic & Cyrillic\\

\bottomrule
\end{tabular}
}
\caption{Languages used in experiments discussed in Section~\ref{sec:family}.}
\label{tab:cyrillic_pivot_lang_metadata}
\end{table}

\begin{figure*}[ht]
    \centering
    \begin{subfigure}[b]{0.49\textwidth}
        \includegraphics[width=\textwidth]{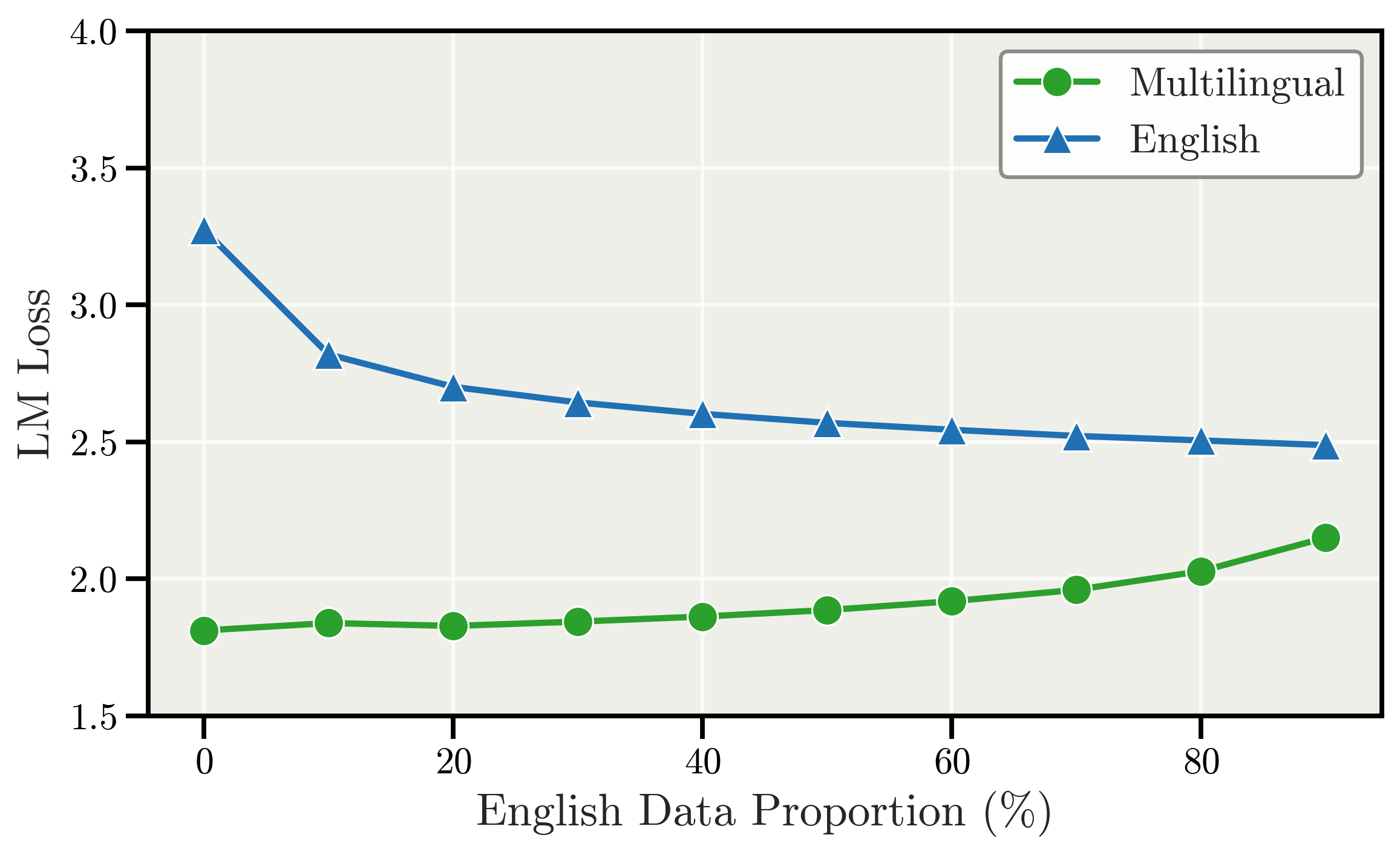}
        \caption{Fixed Total Budget}
        \label{fig:fixed_total_3B}
    \end{subfigure}
    \begin{subfigure}[b]{0.49\textwidth}
        \includegraphics[width=\textwidth]{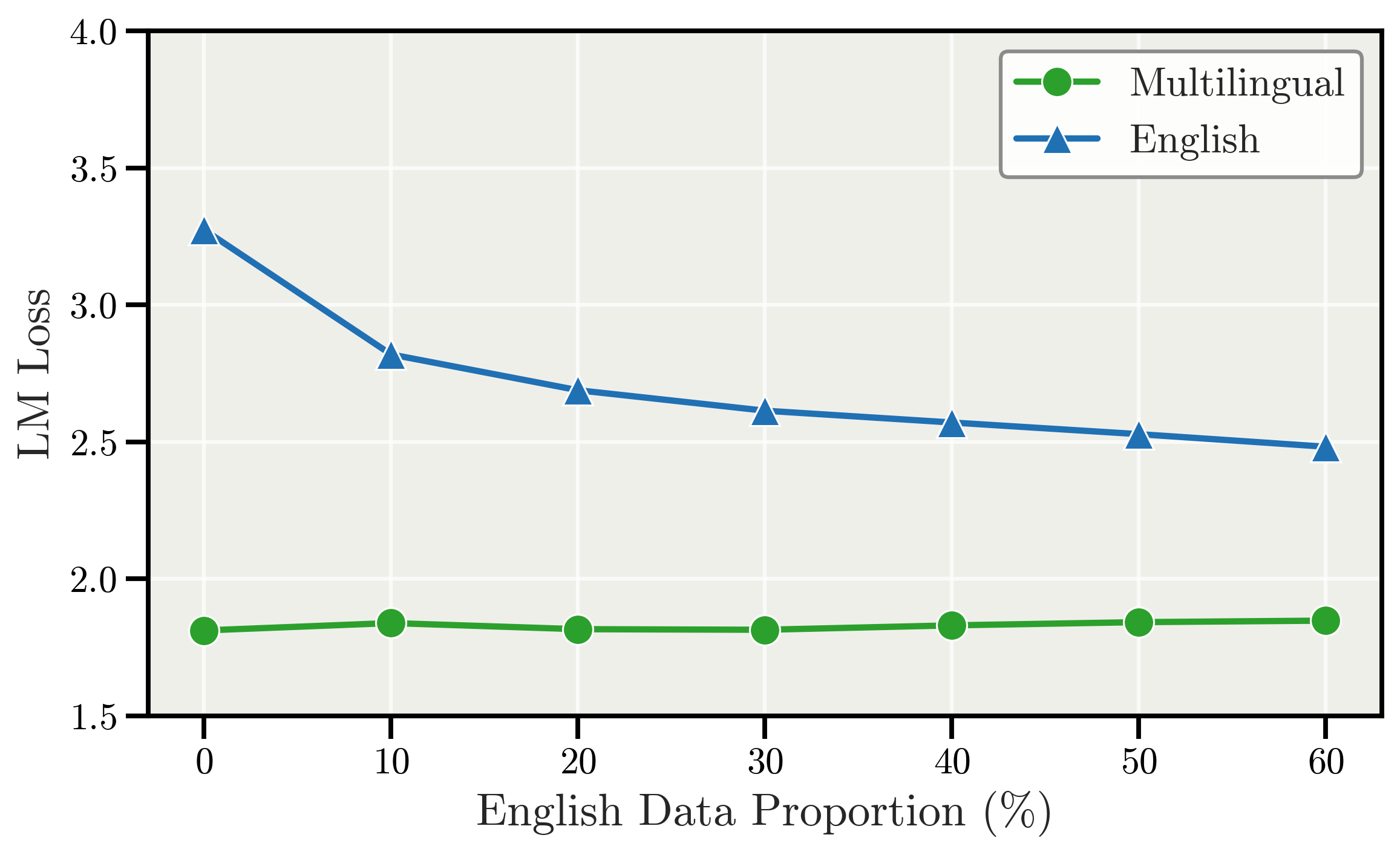}
        \caption{Fixed Multilingual Budget}
        \label{fig:fixed_multilingual_3B}
    \end{subfigure}
    \caption{Validation LM loss for \textbf{English} and weighted average LM loss of non-English (\textbf{Multilingual}) across different proportions of English in the training data for \textbf{$3$B} models. (a) In a \textbf{Fixed Total Budget}, increasing English data ($\geq$50\%), leads to a performance drop in other languages.  (b) In a \textbf{Fixed Multilingual Budget}, increasing English data (up to 60\%) does not have a negative effect on other languages.}
    \label{fig:multilingual_effect_3B}
\end{figure*}

\begin{figure*}[t]
    \centering
    \begin{subfigure}[b]{0.49\textwidth}
        \includegraphics[width=\textwidth]{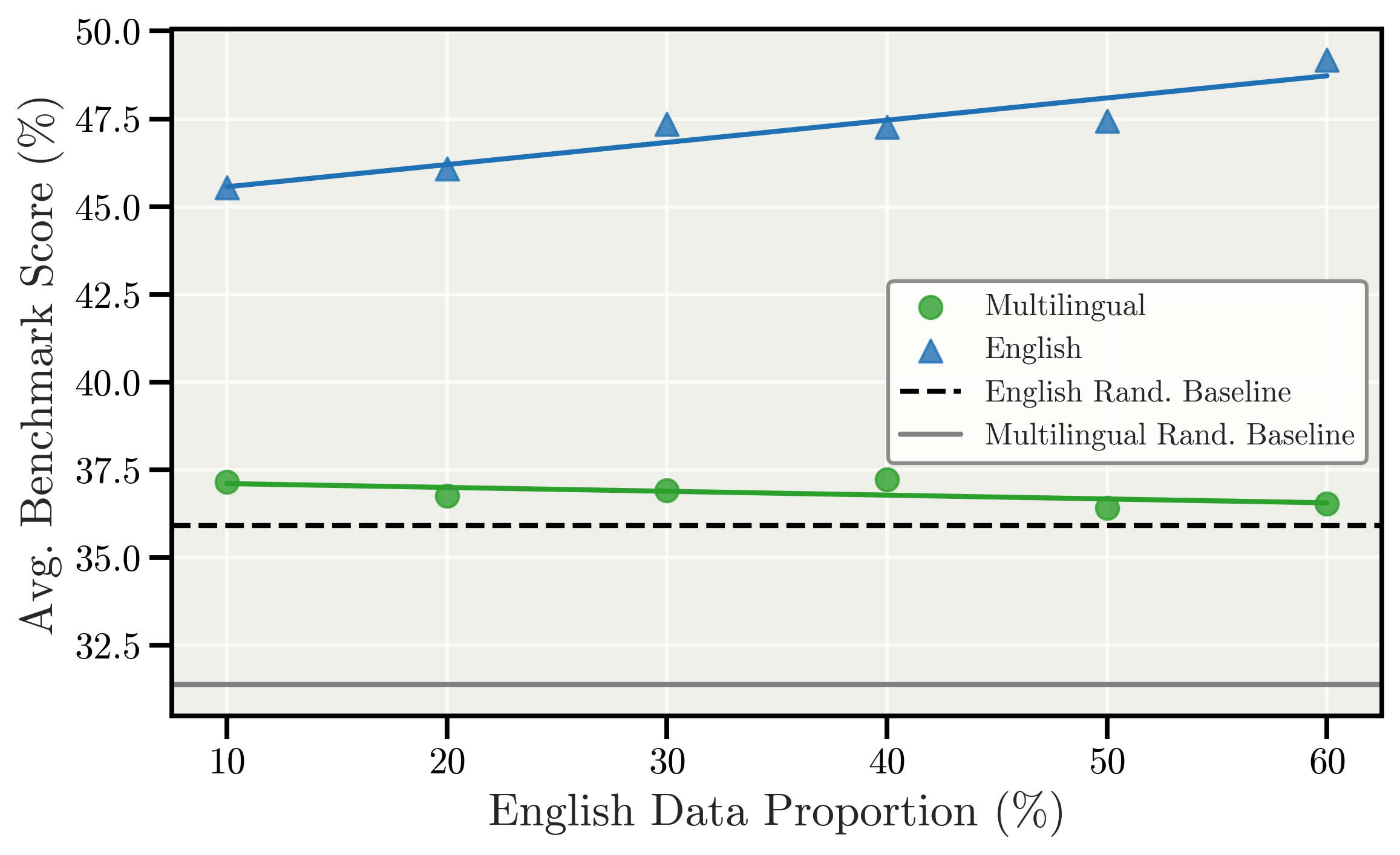}
        \caption{Fixed Total Budget}
        \label{fig:fixed_total_benchmark_3B}
    \end{subfigure}
    \begin{subfigure}[b]{0.49\textwidth}
        \includegraphics[width=\textwidth]{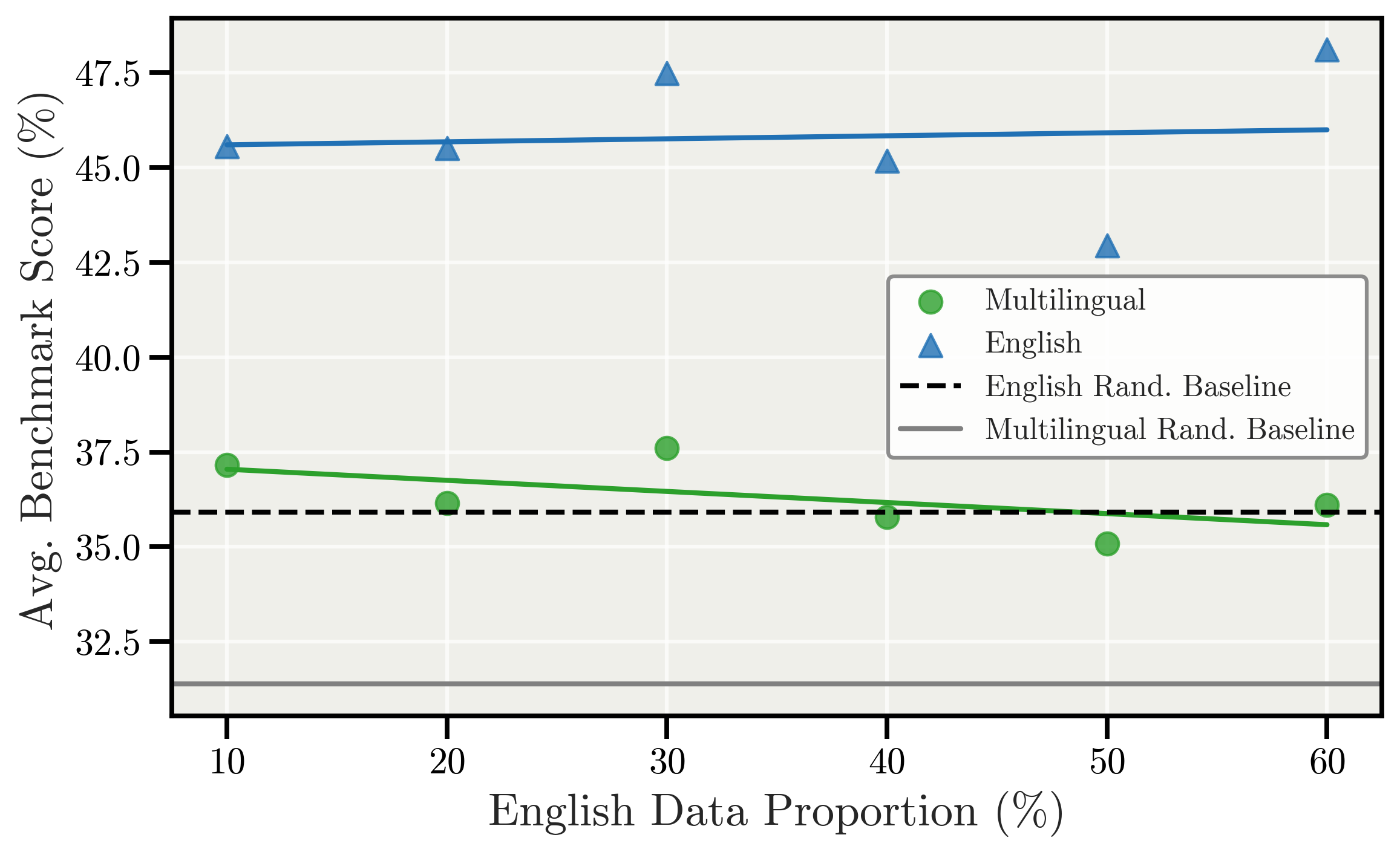}
        \caption{Fixed Multilingual Budget}
        \label{fig:fixed_multilingual_benchmark}
    \end{subfigure}
    \caption{Aggregated benchmark performance for \textbf{English} and weighted average of non-English (\textbf{Multilingual}) across different proportions of English in the training data for \textbf{$3$B} models. 
    The dashed lines represents the random baselines for each language group. 
    \textbf{(a)} In a \textbf{Fixed Total Budget}, increasing English data ($\geq$50\%), does not hurt downstream performance on \textit{other} group. \textbf{(b)} In a \textbf{Fixed Multilingual Budget}, we see that increasing English data has a negligible impact on the \textit{other} group's performance.}
    \label{fig:multilingual_effect_benchmark_3B}
\end{figure*}

Figures~\ref{fig:multilingual_effect_3B} and \ref{fig:multilingual_effect_benchmark_3B} present the validation loss and average benchmark scores for English and non-English (``Multilingual'') languages for \textbf{3B} models. Consistent with our observations for the $1.1$B models, we find that under the Fixed Total Budget setting, increasing the proportion of English data ($\geq50\%$), leads to a decline in performance for other languages. In contrast, under the Fixed Multilingual Budget setting, increasing the share of English data (up to 60\%) does not adversely affect the performance of non-English languages.


\section{Cross-lingual Transfer}
\label{sec:cross_lingual_transfer}

To examine how non-English languages influence English performance under the \textit{Fixed Total Budget} setting, we train models on data spanning 1,834 languages while systematically varying the share of data allocated to each. Specifically, we partition the languages from the \finewebtwo{} dataset into two groups:

\textbf{\textit{Target Languages}. } A set of 45 high- and mid-resource languages that we aim for the model to perform well on.

\textbf{\textit{Tail Languages}. } The remaining 1,789 low-resource languages, which the model is expected to support only as a secondary objective.

The full lists of target and tail languages are provided in Appendix~\ref{target_tail_langs}. Importantly, we exclude English from the training data to neutralize its dominant influence and allow for a clearer analysis of cross-linguistic interactions.
We train $3$B-parameter models by varying the proportion of tail-language data in the training mix, ranging from 6\% to 33\%, and evaluate the impact on performance across the target language set. 

\begin{figure}
    \centering
    \begin{subfigure}[b]{0.49\linewidth}
        \includegraphics[width=\textwidth]{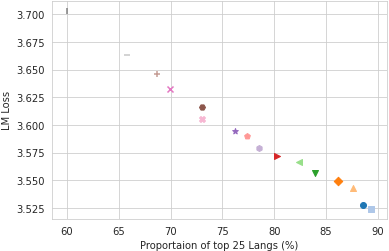}
        \caption{English validation loss }
        \label{fig:com_english_vs_other_loss}
    \end{subfigure}
    \begin{subfigure}[b]{0.49\linewidth}
        \includegraphics[width=\textwidth]{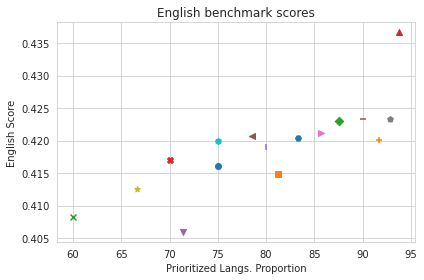}
        \caption{English average benchmark score}
        \label{fig:com_english_vs_other_benchmark}
    \end{subfigure}

    \caption{English (a) validation LM loss and (b) average benchmark score across different proportions of the top 25 languages (model size: $3$B). Increasing token allocation for tail languages reduces validation loss in English and improves English accuracy. 
    }
    \label{fig:com_english_vs_other}
\end{figure}

Figure~\ref{fig:com_english_vs_other_loss} presents the effect of adjusting the balance between the top-25 high-resource languages (in \finewebtwo) and the remaining languages on English validation loss. Although English is not part of the training data, we observe that its validation loss decreases as more tokens from high-resource languages are included, and increases when more tokens from lower-resource languages are introduced. This effect is likely due to the close linguistic proximity of several high-resource languages (\eg, German, French) to English, which provides beneficial transfer.

Supporting this interpretation, we find that English performance is most strongly correlated with Romance, Slavic, and Germanic languages, with Pearson correlation coefficients of 0.78, 0.85, and 0.80, respectively (Table~\ref{tab:english_correlation_loss}). Figure~\ref{fig:com_english_vs_other_benchmark} shows the same pattern in benchmark results: English benefits from the presence of related high-resource languages.
Together, these findings highlight a positive interaction between English and typologically related high-resource languages, which enhances English performance even when it is excluded from training.

\subsection{Target and Tail Languages}
\label{target_tail_langs}
The target languages used in the \curse{} experiments are as follows: 
German, Russian, French, Japanese, Spanish, Mandarin Chinese, Italian, Dutch, Polish, Portuguese, Czech, Vietnamese, Indonesian, Turkish, Swedish, Persian (Farsi), Korean, Hungarian, Arabic, Greek, Romanian, Danish, Finnish, Thai, Ukrainian, Slovak, Norwegian Bokmål, Bulgarian, Catalan, Croatian, Latin, Serbian, Hindi, Slovenian, Lithuanian, Estonian, Hebrew, Latvian, Tosk Albanian, Icelandic, Macedonian, Galician, Basque, Malayalam, Romansh, Swiss German.
Tail languages contain the rest of the languages from the \finewebtwo{} corpus.

Tables~\ref{tab:scrip_stats} and \ref{tab:lang_family_stat} present detailed information about the language families and scripts included in the \finewebtwo{} dataset.

\begin{figure}
    \centering
    \begin{subfigure}[b]{0.32\textwidth}
        \includegraphics[width=\textwidth]{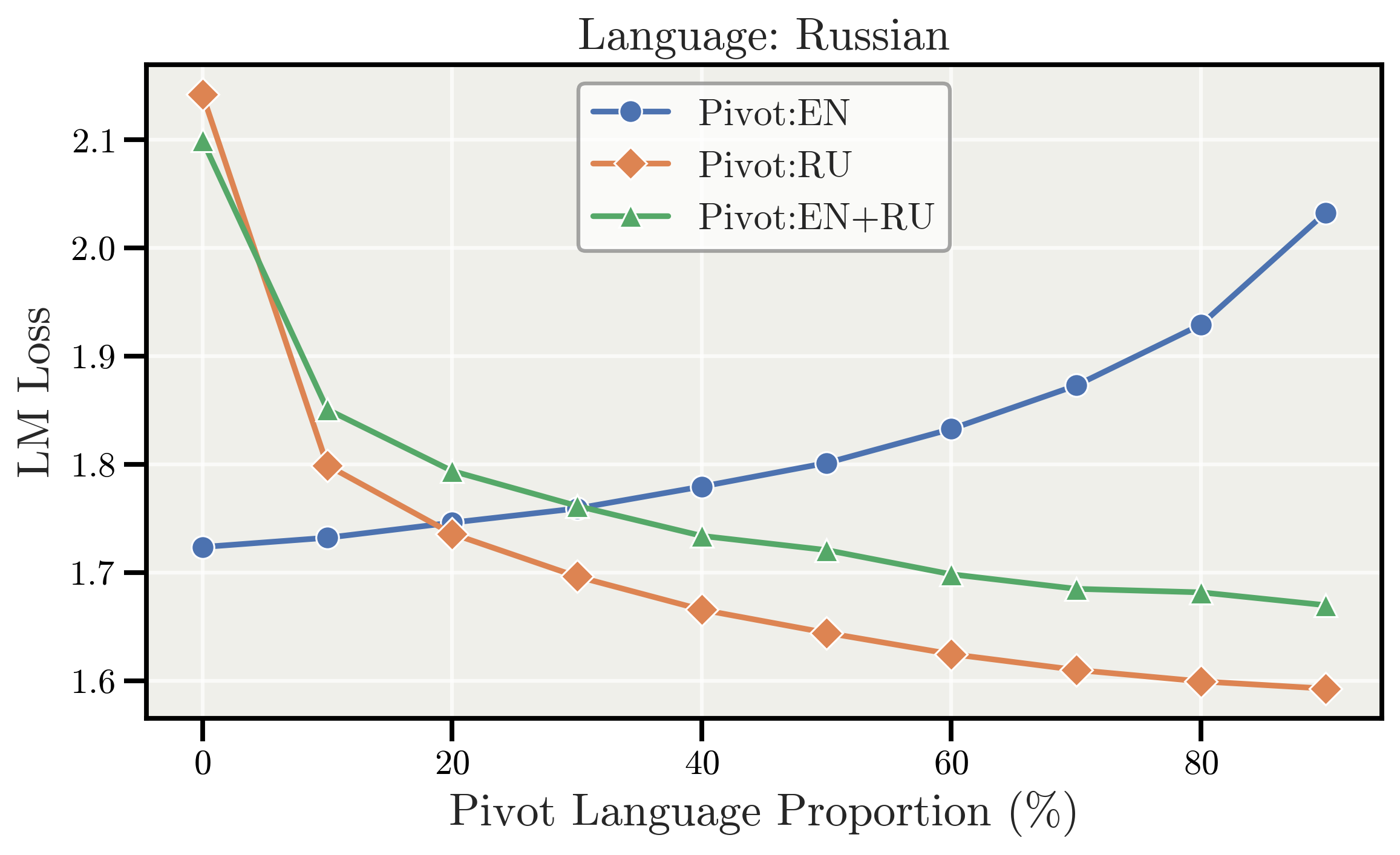}
    \end{subfigure}
    \begin{subfigure}[b]{0.32\textwidth}
        \includegraphics[width=\textwidth]{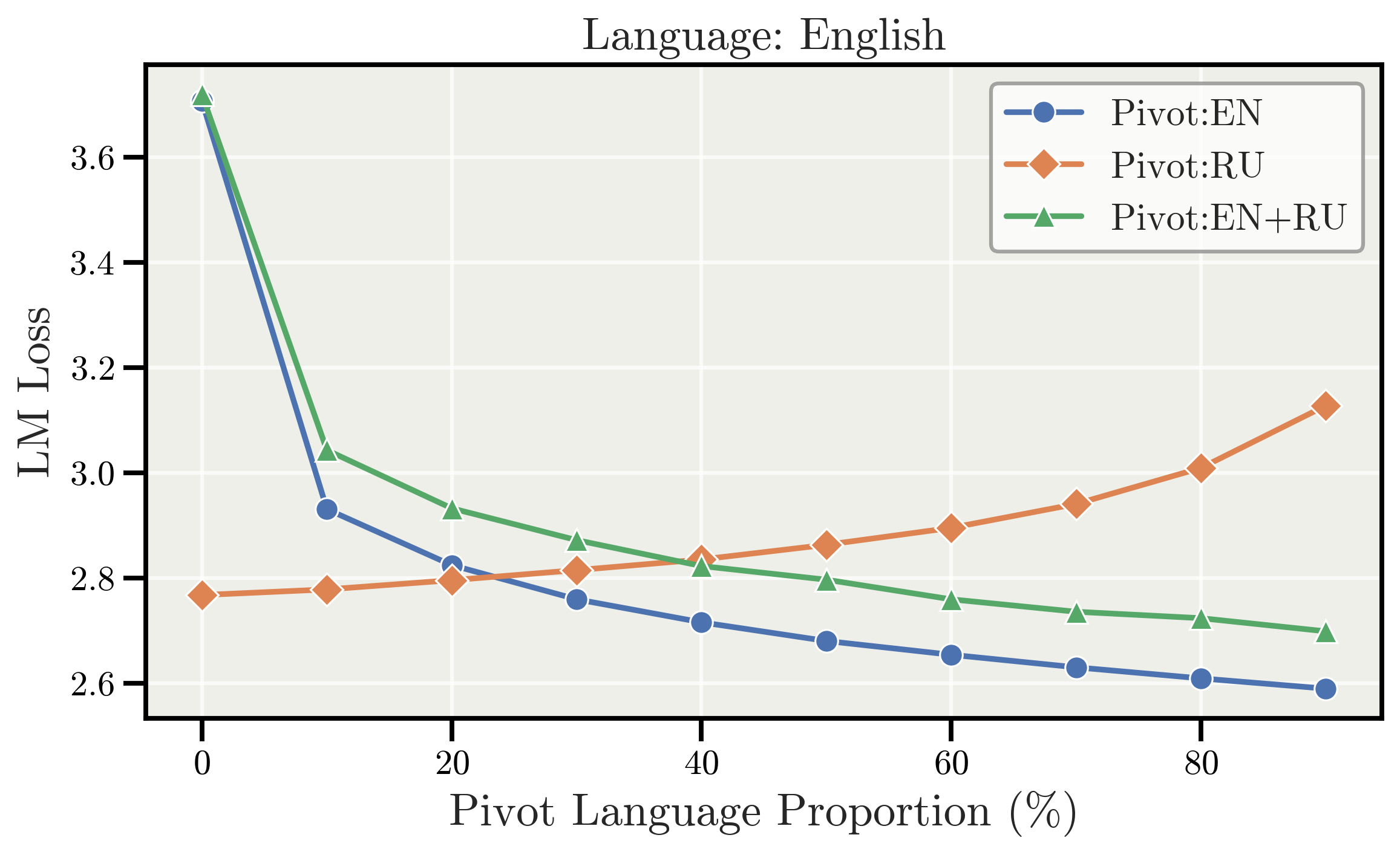}
    \end{subfigure}
    
    \begin{subfigure}[b]{0.32\textwidth}
        \includegraphics[width=\textwidth]{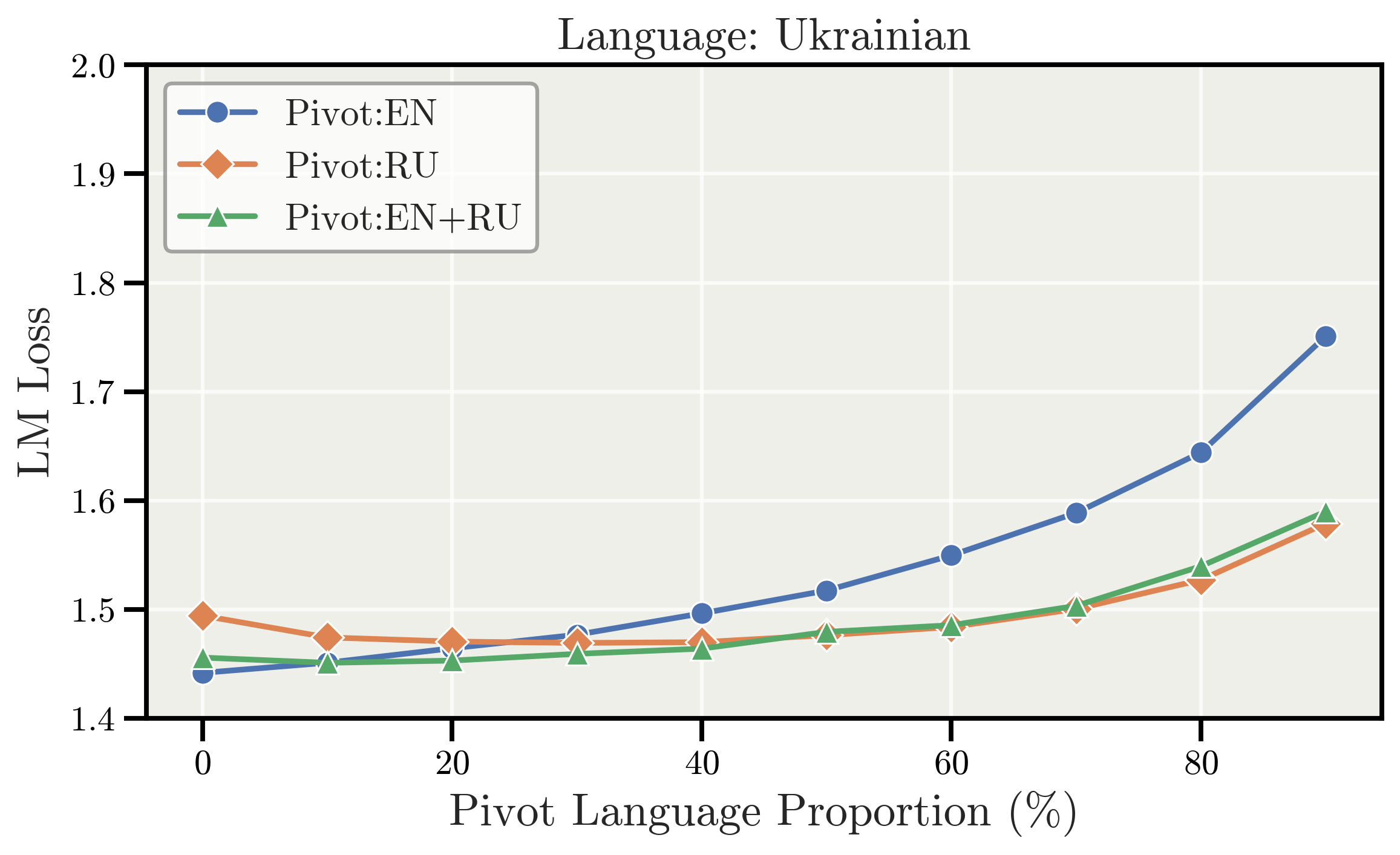}
    \end{subfigure}
    \begin{subfigure}[b]{0.32\textwidth}
        \includegraphics[width=\textwidth]{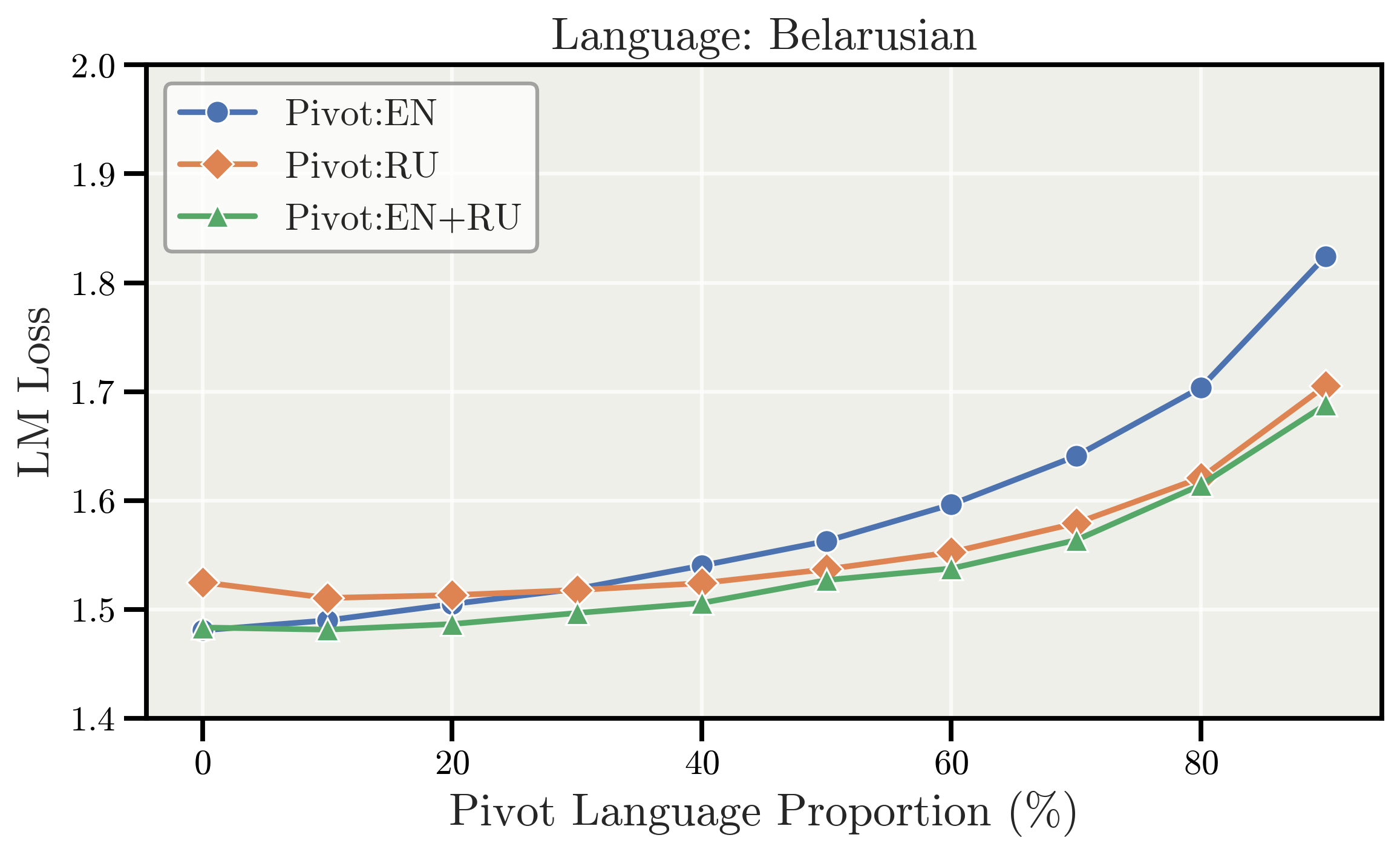}
    \end{subfigure}
    \begin{subfigure}[b]{0.32\textwidth}
        \includegraphics[width=\textwidth]{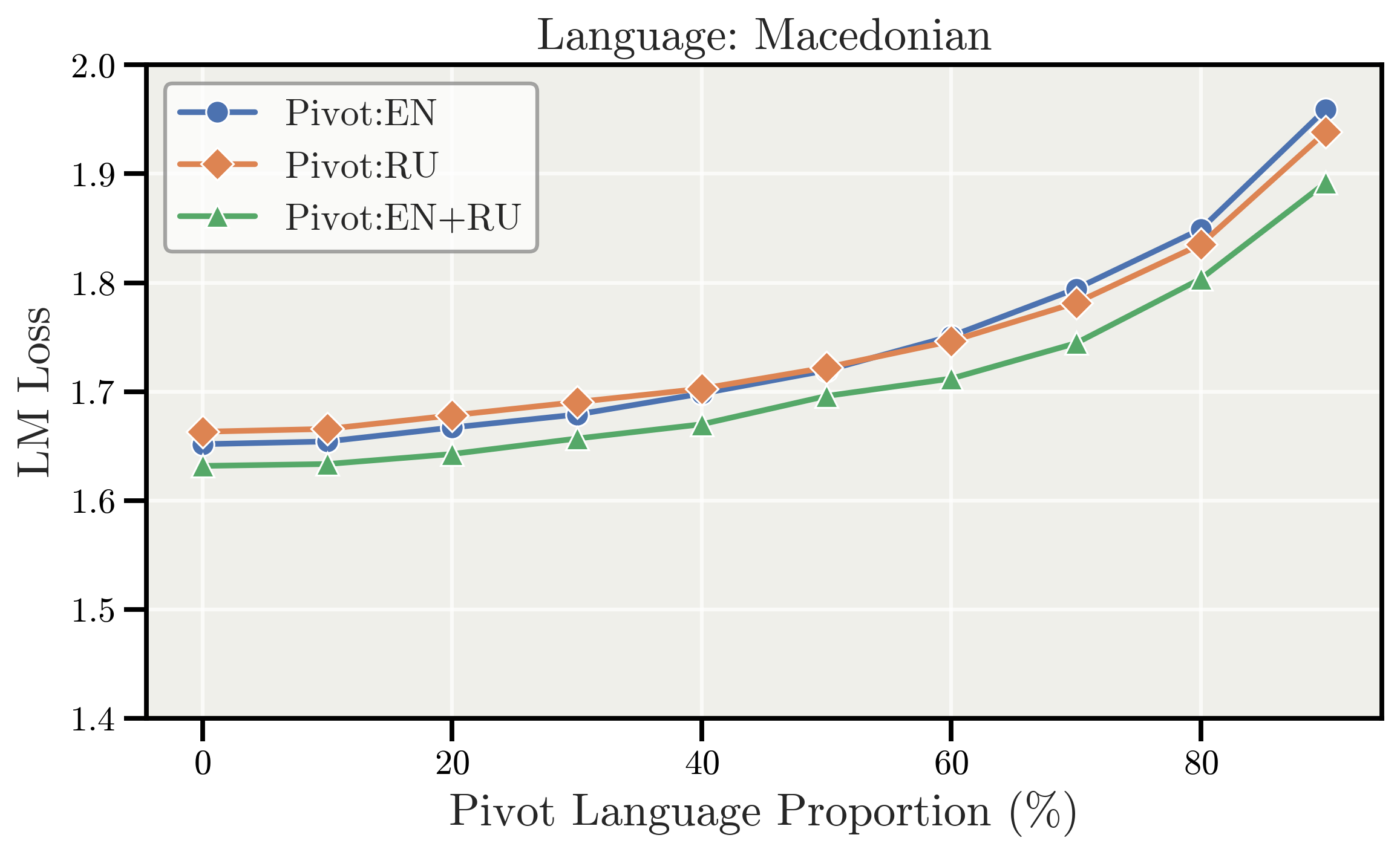}
    \end{subfigure}

    \begin{subfigure}[b]{0.32\textwidth}
        \includegraphics[width=\textwidth]{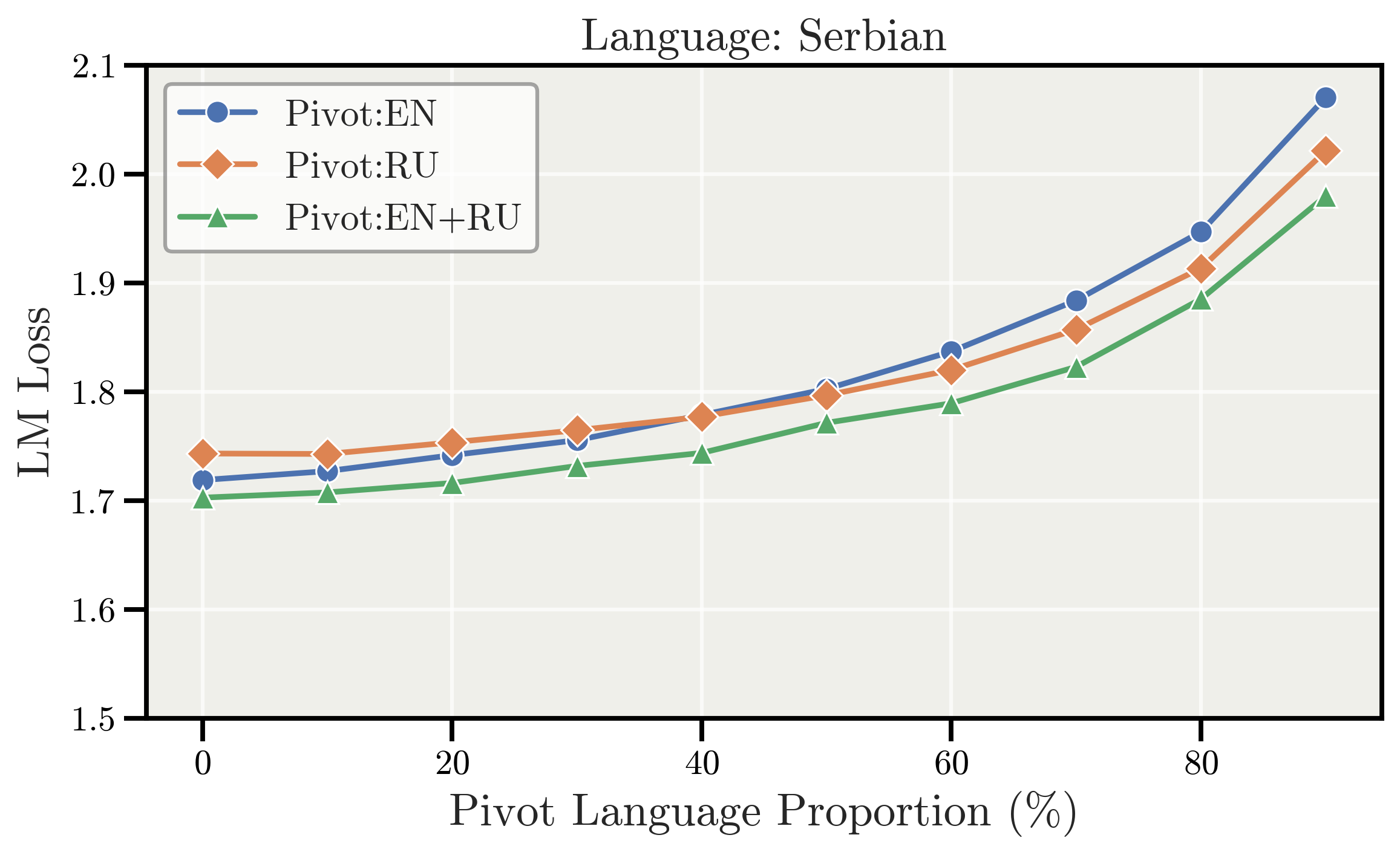}
    \end{subfigure}
    \begin{subfigure}[b]{0.32\textwidth}
        \includegraphics[width=\textwidth]{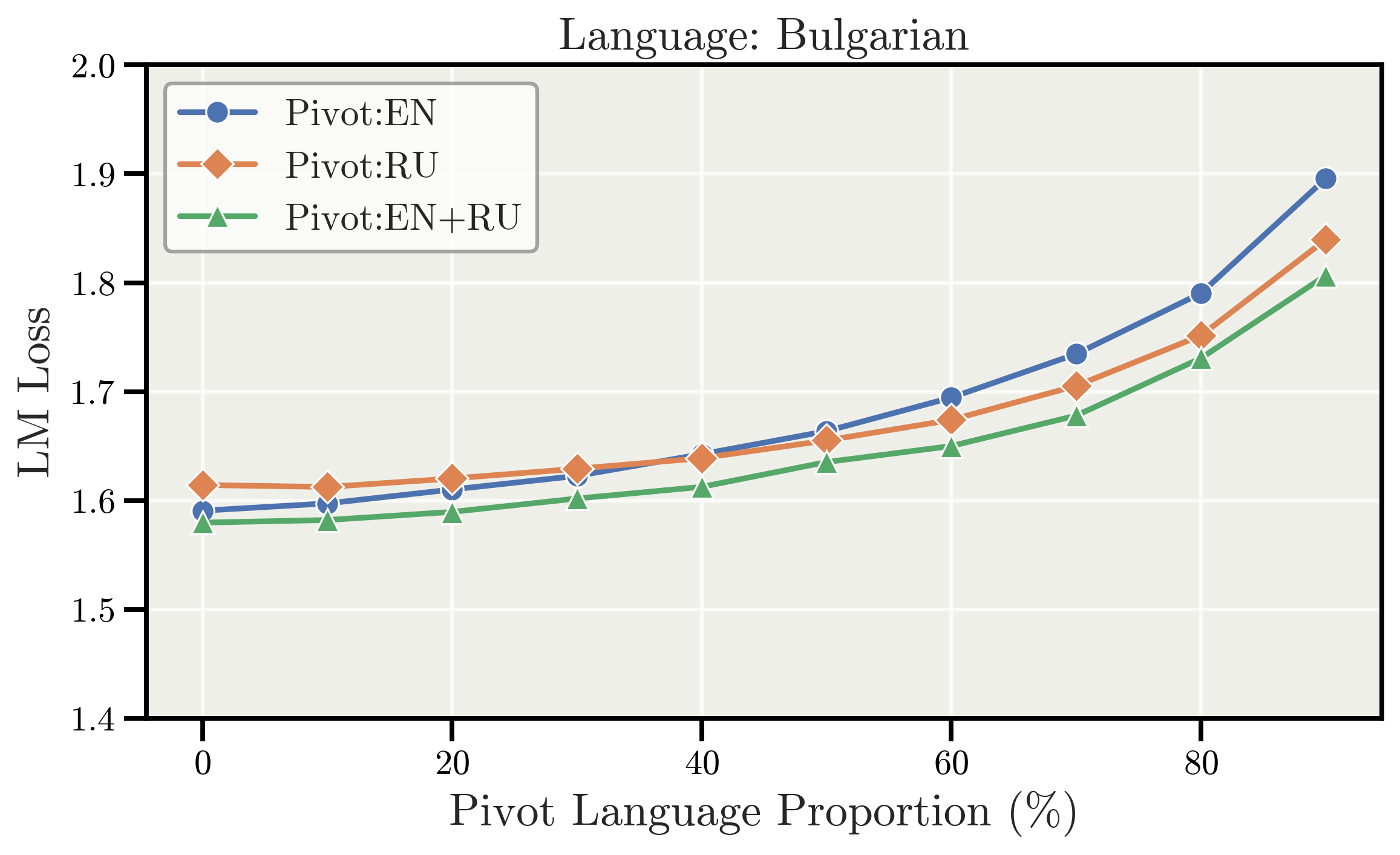}
    \end{subfigure}
    \begin{subfigure}[b]{0.32\textwidth}
        \includegraphics[width=\textwidth]{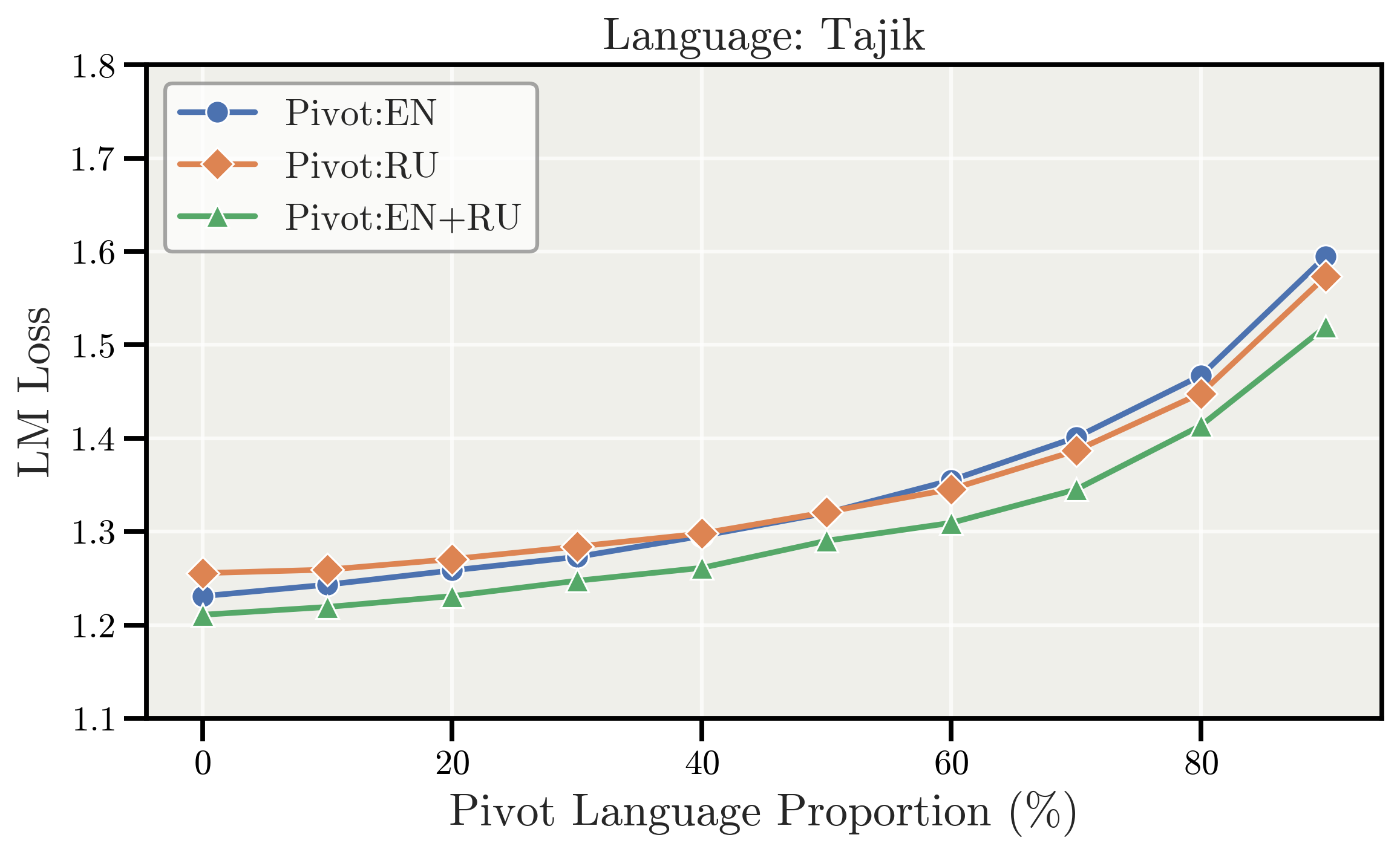}
    \end{subfigure}
    
    \begin{subfigure}[b]{0.32\textwidth}
        \includegraphics[width=\textwidth]{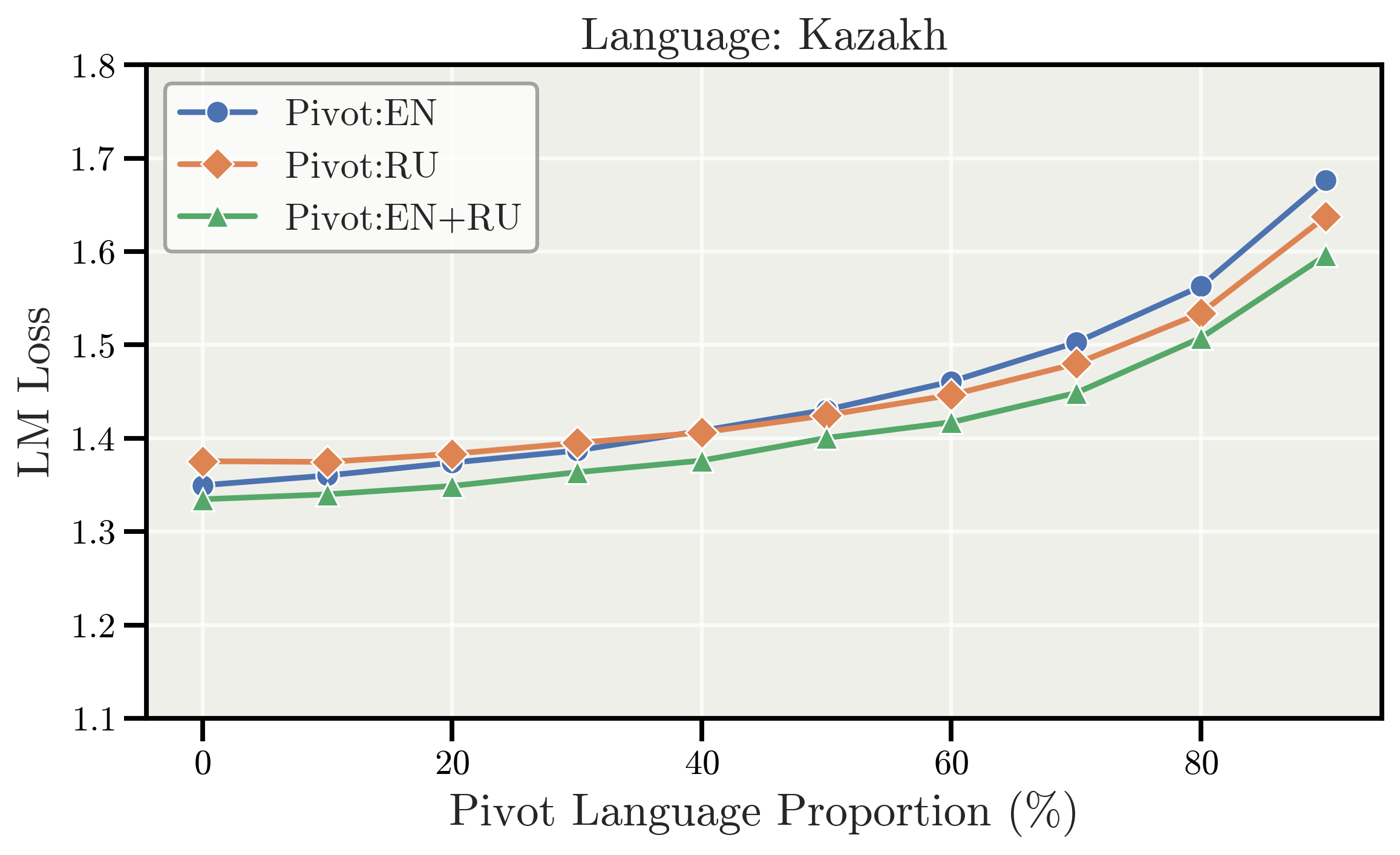}
    \end{subfigure}
    \begin{subfigure}[b]{0.32\textwidth}
        \includegraphics[width=\textwidth]{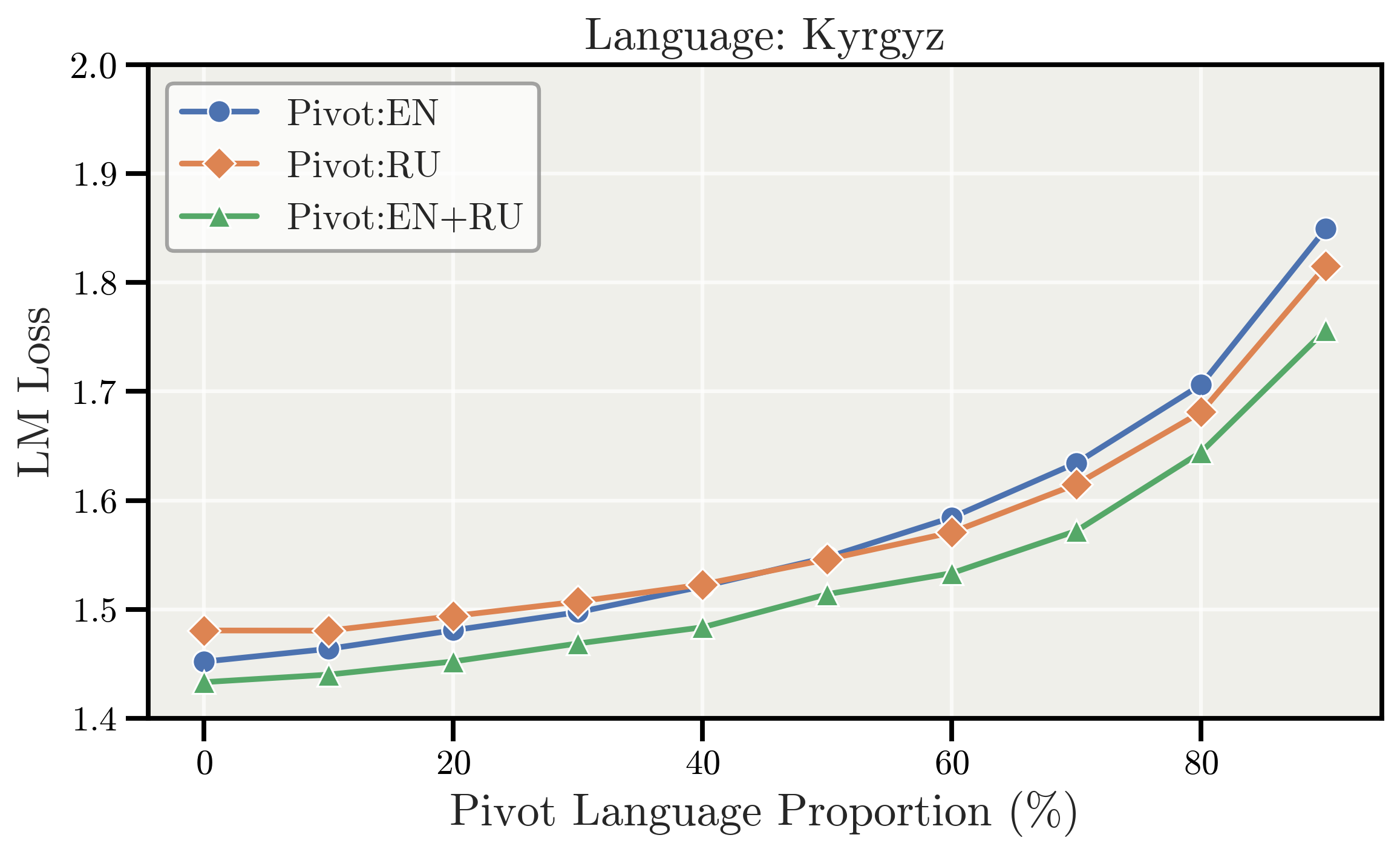}
    \end{subfigure}
    \begin{subfigure}[b]{0.32\textwidth}
        \includegraphics[width=\textwidth]{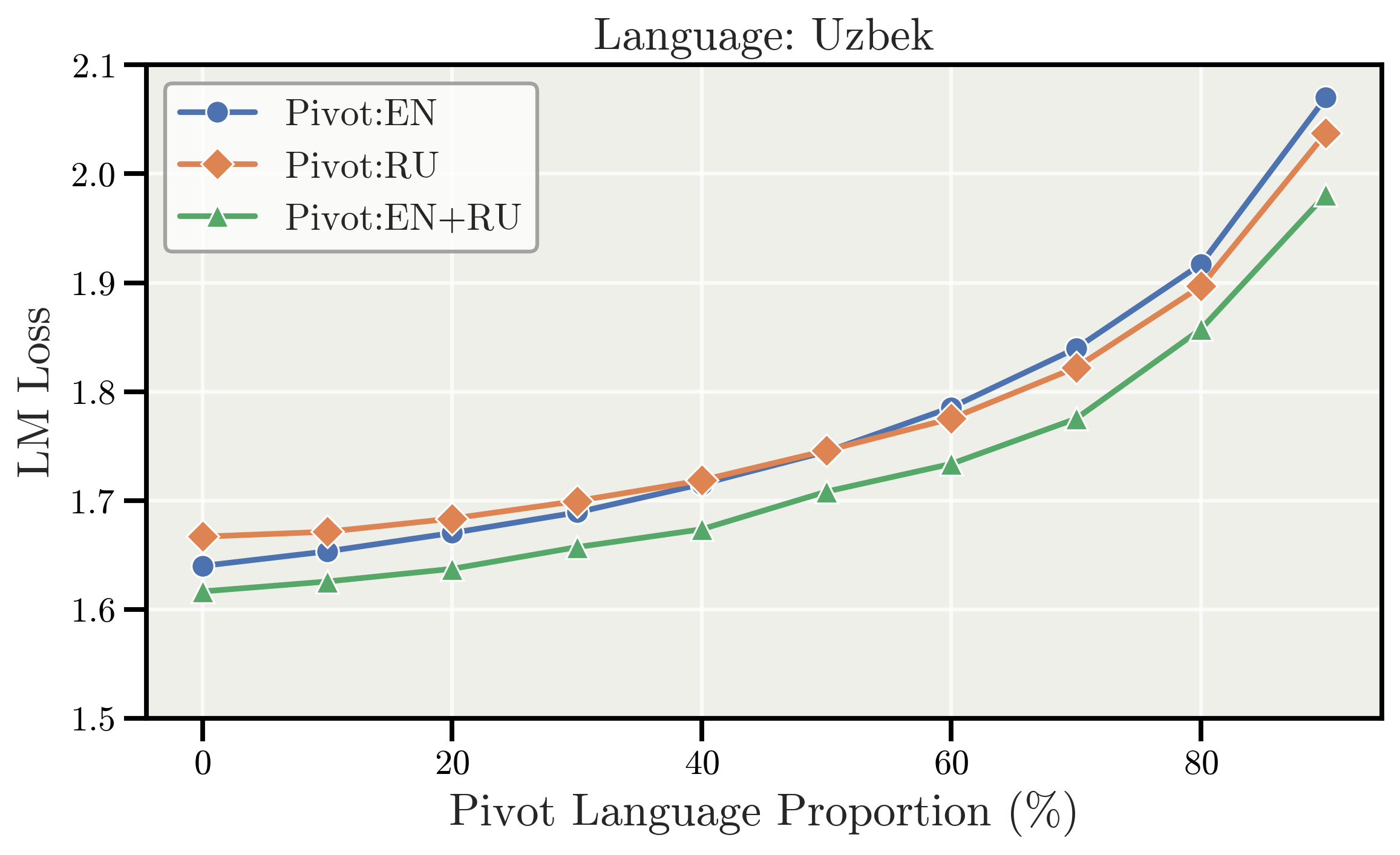}
    \end{subfigure}

    \begin{subfigure}[b]{0.32\textwidth}
        \includegraphics[width=\textwidth]{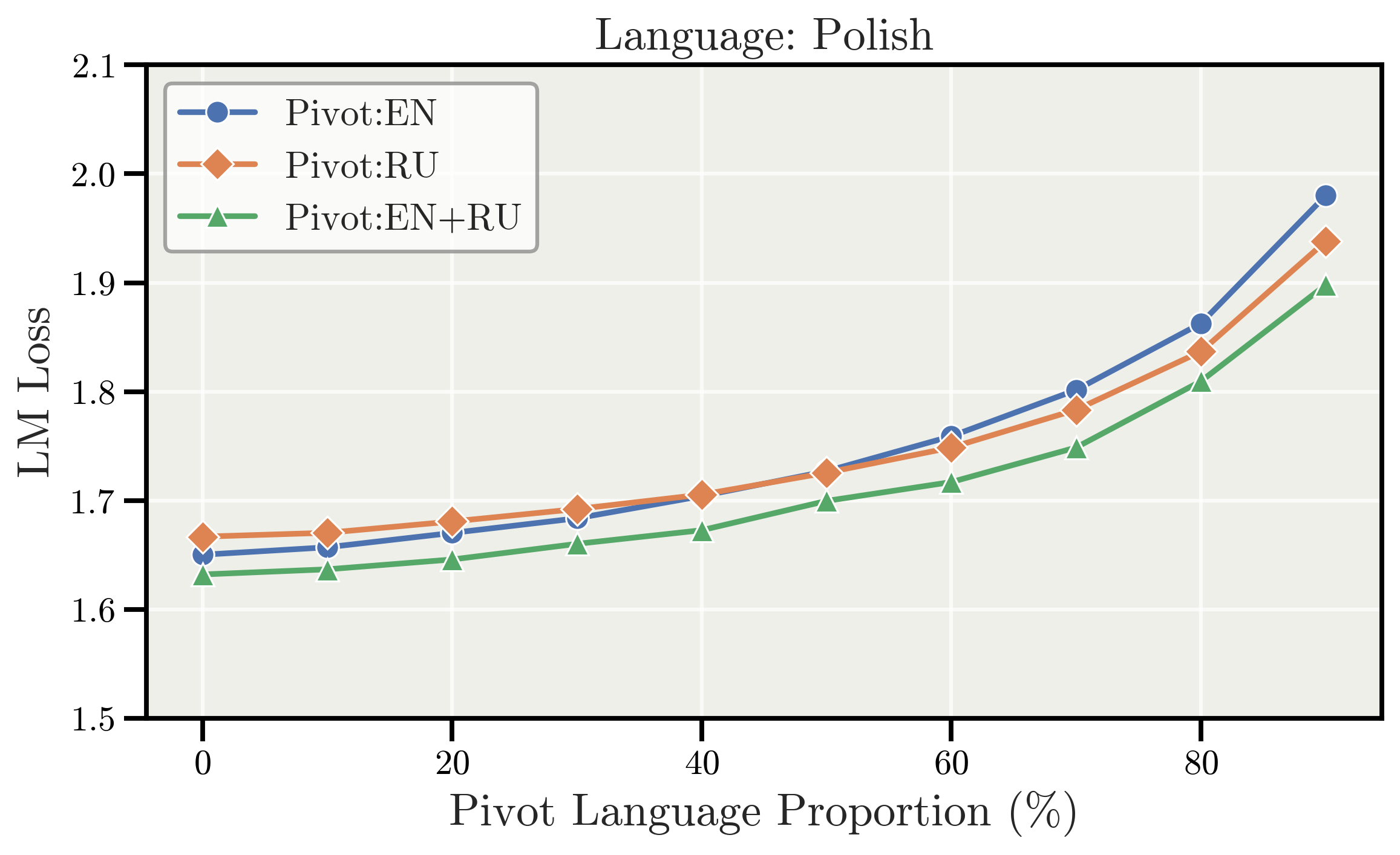}
    \end{subfigure}
    \begin{subfigure}[b]{0.32\textwidth}
        \includegraphics[width=\textwidth]{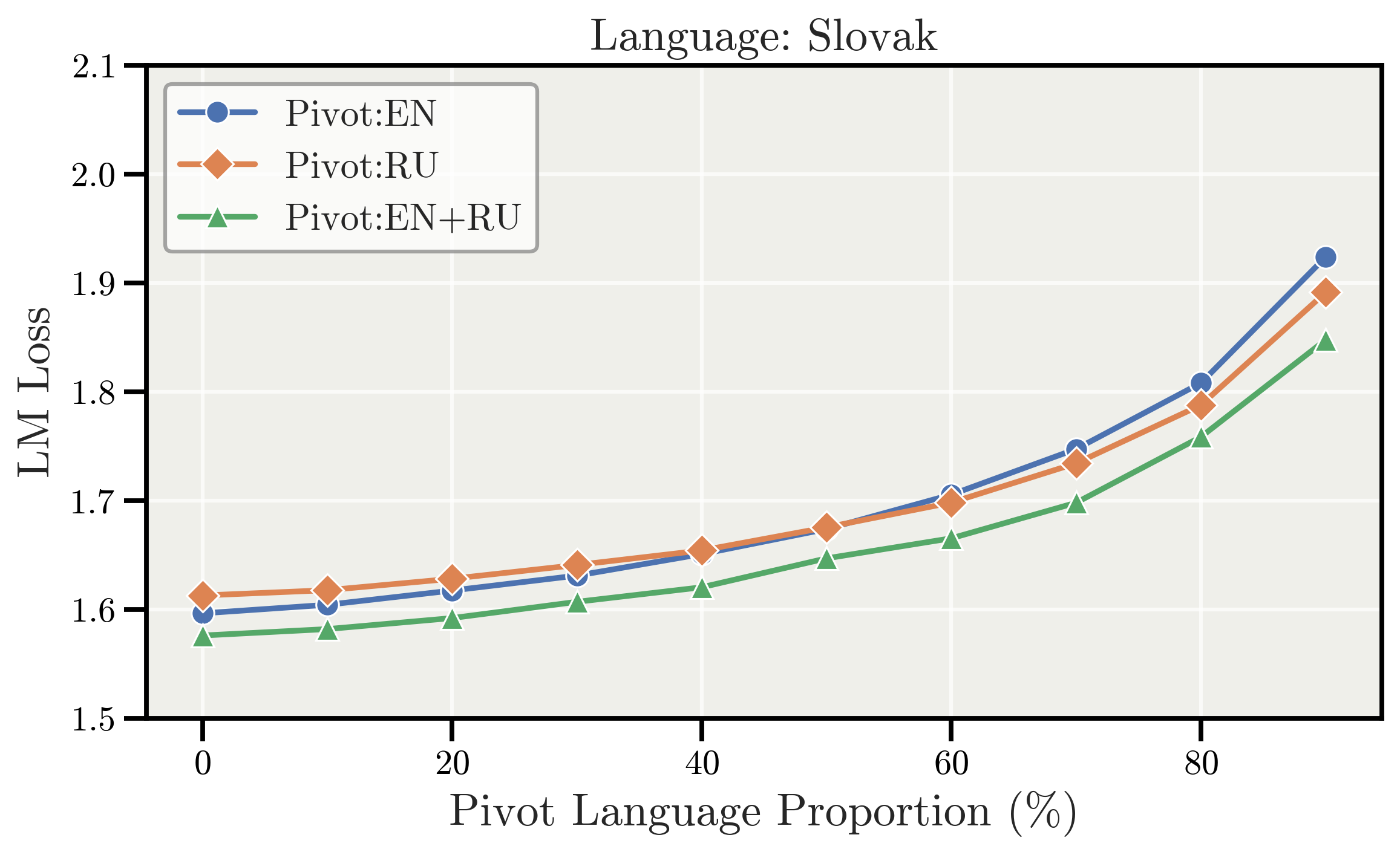}
    \end{subfigure}
    \begin{subfigure}[b]{0.32\textwidth}
        \includegraphics[width=\textwidth]{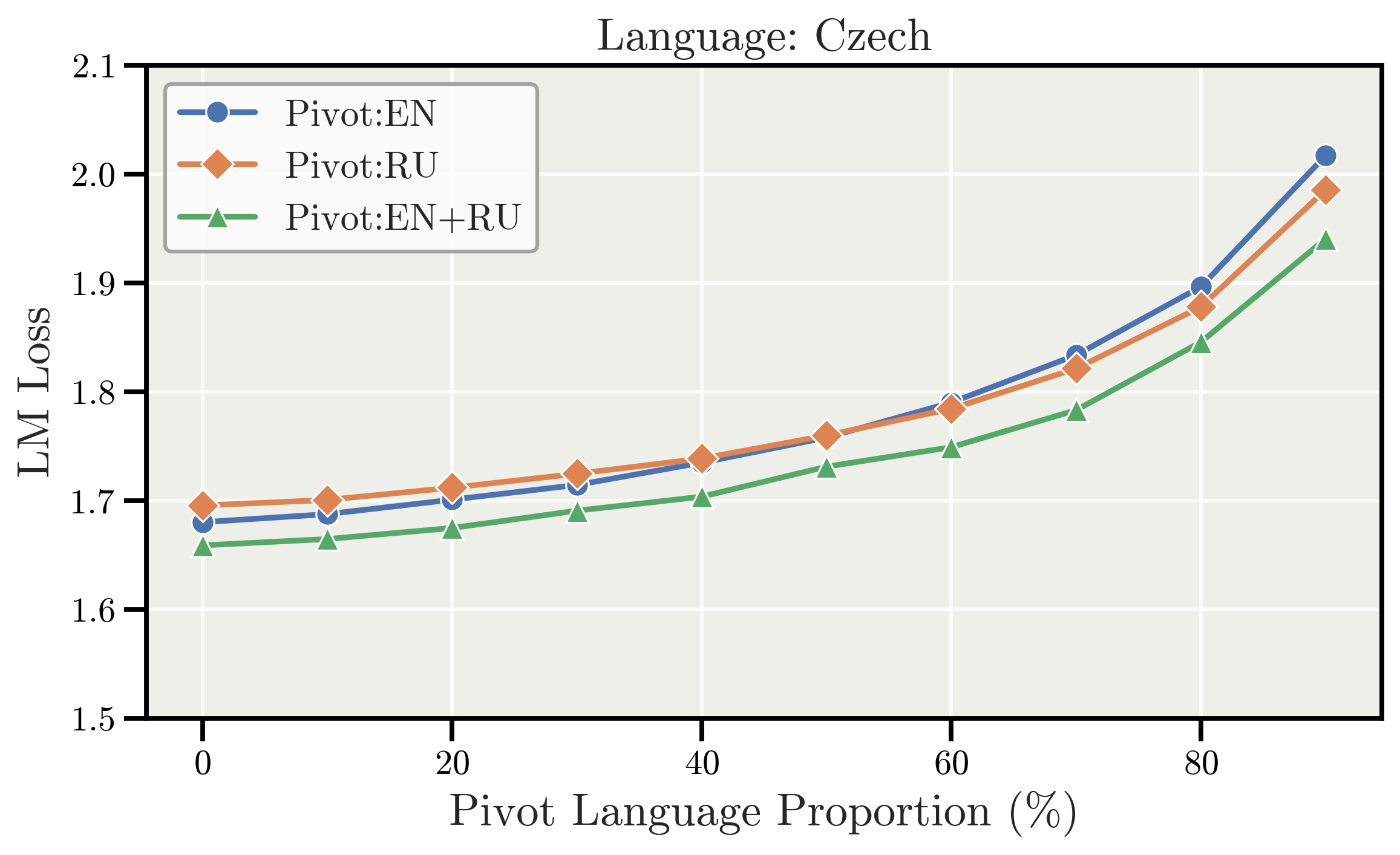}
    \end{subfigure}
    \caption{Validation LM loss for each language in the experiments described in Section~\ref{sec:family}, using English, Russian, or a combination of English and Russian as the pivot language in the training mix. The combination of English and Russian yields the best performance for most languages (model size: $1.1$B).}
    \label{fig:pivot_ablation_per_lang_loss}
\end{figure}

\begin{table}
\centering
\resizebox{0.8\textwidth}{!}{
\begin{tabular}{lc}
\toprule
\textbf{Language Family} & \textbf{Pearson Correlation} \\
\midrule
Slavic                             &  0.853 \\
Germanic                           &  0.808 \\
Romance                            &  0.785 \\
Malayo-Sumbawan               &  0.683 \\
Semitic                            &  0.521 \\
Creoles and Pidgins                & –0.578 \\
Kuki-Chin                     & –0.759 \\
Bantu                              & –0.776 \\
Greater Central Philippine         & –0.827 \\
Mixtec                             & –0.832 \\
Celebic                            & –0.872 \\
Cariban                            & –0.895 \\
Panoan                             & –0.897 \\
Oti-Volta                     & –0.899 \\
Western Mande                      & –0.915 \\
Zapotecan                          & –0.919 \\
Mayan                              & –0.925 \\
Brahmaputran                       & –0.939 \\
Northern Luzon                     & –0.945 \\
Chinantecan                        & –0.952 \\
Oceanic                            & –0.962 \\
Algonquian                         & –0.963 \\
Quechuan                           & –0.980 \\
Central Malayo-Polynesian     & –0.985 \\
Maweti-Guarani                & –0.985 \\
Tucanoan                           & –0.992 \\
\bottomrule
\end{tabular}
}
\caption{Pearson correlation ($r$) between English validation loss and each language family,  
retaining only results with $p\!<\!0.05$ and sorted in descending order of $r$.}
\label{tab:english_correlation_loss}
\end{table}

\begin{table}[htbp]
  \centering
  \begin{minipage}{0.48\linewidth}
  \captionsetup{width=\linewidth}
  \rowcolors{2}{rowgray}{rowgray}  
  \centering
  \begin{tabular}{lr}
    \toprule
    \rowcolor{white}        
    \textbf{Script} & \textbf{\# Languages} \\
    \toprule
    Latn & 1639 \\
    Cyrl & 56  \\
    Arab & 30  \\
    Deva & 29  \\
    Ethi & 9   \\
    Thai & 7   \\
    Cans & 6   \\
    Beng & 5   \\
    Mymr & 5   \\
    Hani & 5   \\
    Telu & 3   \\
    Hebr & 3   \\
    Grek & 3   \\
    Tibt & 3   \\
    Tfng & 2   \\
    Armn & 2   \\
    Orya & 2   \\
    Geor & 2   \\
    Syrc & 2   \\
    Laoo & 2   \\
    Knda & 2   \\
    \bottomrule
  \end{tabular}
  \caption{Scripts and the number of languages each one supports. Sixteen other scripts are present in the \finewebtwo{} dataset, each supporting one language.}
\label{tab:scrip_stats}
  \end{minipage}\hfill
    \begin{minipage}{0.48\linewidth}
    \rowcolors{2}{rowgray}{rowgray}
  \begin{tabular}{lr}
    \toprule
    \rowcolor{white}\textbf{Language Family} & \textbf{\# Languages}\\\toprule
    Bantu & 73\\
    Oceanic & 67\\
    Mayan & 24\\
    Turkic & 22\\
    Indic & 22\\
    Creoles and Pidgins & 20\\
    Germanic & 17\\
    Tucanoan & 16\\
    Greater Central Philippine & 15\\
    Romance & 15\\
    Semitic & 14\\
    Mixtec & 13\\
    Slavic & 13\\
    Zapotecan & 12\\
    Central Malayo-Polynesian & 12\\
    Iranian & 11\\
    Oti-Volta & 11\\
    Malayo-Sumbawan & 11\\
    Kuki-Chin & 10\\
    Northern Luzon & 10\\
    Celebic & 9\\
    Quechuan & 9\\
    Maweti-Guarani & 9\\
    Dravidian & 8\\
    Brahmaputran & 8\\
    Panoan & 8\\
    Western Mande & 8\\
    Cariban & 8\\
    Algonquian & 8\\
    Chinantecan & 7\\
    \bottomrule
  \end{tabular}
  \caption{Top language sub-families in \finewebtwo{} and their number of associated languages. The classification is according to \citet{wals}. Labels for 768 languages in \finewebtwo{} were not available.}
  \label{tab:lang_family_stat}

    \end{minipage}
\end{table}
\begin{table}
\centering
\resizebox{\textwidth}{!}{%
\centering
\begin{tabular}{llcccc}
\toprule
\makecell{\textbf{Num PT}\\\textbf{Langs}}  & \textbf{Variant} &
\makecell{\textbf{Top 25 lang}\\\textbf{$B$ Tokens (Prop.)}} &
\makecell{\textbf{Top 50 lang}\\\textbf{$B$ Tokens (Prop.)}} &
\makecell{\textbf{Top 100 lang}\\\textbf{$B$ Tokens (Prop.)}} &
\makecell{\textbf{Top 200 lang}\\\textbf{$B$ Tokens (Prop.)}} \\
\midrule

\multirow{4}{*}{50}
& Natural             & 55.77 (0.56) & -            & -            & - \\
& Temp.               & 40.15 (0.40) & -            & -            & - \\
& Natural -- C & 60.08 (0.56) & -            & -            & - \\
& Temp. -- C   & 60.08 (0.40) & -            & -            & - \\
\midrule

\multirow{4}{*}{100}
& Natural             & 55.07 (0.55) & 59.33 (0.59) & -            & - \\
& Temp.               & 30.51 (0.30) & 45.65 (0.46) & -            & - \\
& Natural -- C & 55.77 (0.55) & 60.08 (0.59) & -            & - \\
& Temp. -- C   & 40.15 (0.30) & 60.08 (0.46) & -            & - \\
\midrule

\multirow{4}{*}{200}
& Natural             & 54.98 (0.55) & 59.23 (0.59) & 59.99 (0.60) & - \\
& Temp.               & 25.28 (0.25) & 37.83 (0.38) & 49.79 (0.50) & - \\
& Natural -- C & 55.07 (0.55) & 59.33 (0.59) & 60.08 (0.60) & - \\
& Temp. -- C   & 30.51 (0.25) & 45.65 (0.38) & 60.08 (0.50) & - \\
\midrule

\multirow{4}{*}{400}
& Natural             & 54.97 (0.55) & 59.22 (0.59) & 59.98 (0.60) & 60.07 (0.60) \\
& Temp.               & 22.07 (0.22) & 33.03 (0.33) & 43.47 (0.43) & 52.46 (0.52) \\
& Natural -- C & 54.98 (0.55) & 59.23 (0.59) & 59.99 (0.60) & 60.08 (0.60) \\
& Temp. -- C   & 25.28 (0.22)             &  37.83 (0.33)            & 49.79 (0.43)             & 60.08 (0.52) \\
\bottomrule
\end{tabular}
}
\caption{
Total number of tokens (in billions) and the corresponding proportions contributed by the top-25, 50, 100, and 200 languages. \textit{Num PT Langs} refers to the total number of languages included during pretraining. \textit{Natural} and \textit{Temp.} represent natural sampling and temperature-based sampling, respectively, both conducted with a fixed token budget of $100$B tokens. \textit{Natural-C} and \textit{Temp.-C} denote the same sampling strategies applied under the Controlled Growth setting, which uses a total of $90$B tokens. English is excluded from the token counts and proportions.
}
\label{tab:token_distribution}
\end{table}

\begin{table}
\centering
\small
\resizebox{\textwidth}{!}{
\begin{tabular}{llccccccccc}
\toprule
\textbf{\makecell{\textbf{Num PT}\\\textbf{Langs}}} & \textbf{Variant} &
\textbf{BB} & \textbf{M3E} & \textbf{MMMLU} & \textbf{PAWS-X} &
\textbf{XCSQA} & \textbf{XCodah} & \textbf{XCopa} &
\textbf{XSC} & \textbf{XWG}\\
\midrule

\multirow{2}{*}{25}
& Natural & 38.22 & 38.70 & 30.75 & 49.60 & 35.70 & 51.67 & 66.80 & 75.10 & 65.60\\
& Temp.   & 33.67 & 33.20 & 27.52 & 45.70 & 31.20 & 37.33 & 62.00 & 63.60 & 54.90\\
\midrule

\multirow{2}{*}{50}
& Natural & 37.44 & 38.60 & 30.91 & 49.00 & 33.00 & 51.67 & 67.00 & 73.60 & 66.50\\
& Temp.   & 32.33 & 33.50 & 27.51 & 55.90 & 31.60 & 38.00 & 61.80 & 65.30 & 55.80\\
\midrule

\multirow{2}{*}{100}
& Natural & 37.67 & 37.60 & 30.72 & 50.40 & 34.10 & 51.67 & 69.40 & 74.80 & 65.10\\
& Temp.   & 32.22 & 33.90 & 26.75 & 55.20 & 30.90 & 38.67 & 61.20 & 63.20 & 54.60\\
\midrule

\multirow{2}{*}{200}
& Natural & 37.44 & 37.20 & 30.54 & 54.00 & 31.80 & 52.33 & 66.20 & 75.00 & 65.40\\
& Temp.   & 31.67 & 32.80 & 27.07 & 43.70 & 28.80 & 37.33 & 62.00 & 62.40 & 55.40\\
\midrule

\multirow{2}{*}{400}
& Natural & 37.33 & 38.60 & 30.61 & 55.20 & 35.30 & 53.33 & 68.60 & 73.80 & 64.20\\
& Temp.   & 31.22 & 29.70 & 26.78 & 55.40 & 24.70 & 34.67 & 57.20 & 62.30 & 56.50\\
\bottomrule
\end{tabular}
}
\caption{Benchmark scores (\%) for English with varying number and sampling of 25--400 languages during pretraining. \textit{Num PT Langs} refers to the total number of languages included during pretraining. \textit{Natural} and \textit{Temp.} represent natural sampling and temperature-based sampling, respectively, both conducted with a fixed token budget of $100$B tokens. \textit{BB}, \textit{M3E}, \textit{XSC}, and \textit{XWG} denote the results for BeleBele, M3Exams, XStoryCloze, and XWinogrande respectively.}
\label{tab:english_benchmark}
\end{table}

\begin{table}[ht]
    \centering
    \scriptsize
    \setlength{\tabcolsep}{4pt}

    \resizebox{\textwidth}{!}{
    \begin{tabular}{lccccccccccc}
    \toprule
    Language & en=00\% & en=10\% & en=20\% & en=30\% & en=40\% & en=50\% & en=60\% & en=70\% & en=80\% & en=90\% & en=100\% \\
    \midrule
    ar & 3810.73 (3.8\%) & 3360.93 (3.4\%) & 2987.49 (3.0\%) & 2614.06 (2.6\%) & 2286.44 (2.3\%) & 1867.18 (1.9\%) & 1434.00 (1.5\%) & 1143.22 (1.1\%) & 762.15 (0.8\%) & 381.07 (0.4\%) & 0.00 (0.0\%) \\
bg & 2855.74 (2.9\%) & 2609.73 (2.6\%) & 2319.76 (2.3\%) & 2029.79 (2.0\%) & 1713.45 (1.7\%) & 1449.85 (1.4\%) & 1113.49 (1.2\%) & 856.72 (0.9\%) & 571.15 (0.6\%) & 285.57 (0.3\%) & 0.00 (0.0\%) \\
bn & 2044.27 (2.0\%) & 1850.78 (1.9\%) & 1645.14 (1.6\%) & 1439.50 (1.4\%) & 1226.56 (1.2\%) & 1028.21 (1.0\%) & 789.67 (0.8\%) & 613.28 (0.6\%) & 408.85 (0.4\%) & 204.43 (0.2\%) & 0.00 (0.0\%) \\
ca & 2434.91 (2.4\%) & 2245.24 (2.2\%) & 1995.77 (2.0\%) & 1746.30 (1.7\%) & 1460.95 (1.5\%) & 1247.36 (1.2\%) & 957.97 (1.0\%) & 730.47 (0.7\%) & 486.98 (0.5\%) & 243.49 (0.2\%) & 0.00 (0.0\%) \\
de & 6587.51 (6.6\%) & 6186.77 (6.2\%) & 5499.35 (5.5\%) & 4811.93 (4.8\%) & 3952.51 (4.0\%) & 3437.09 (3.4\%) & 2639.69 (2.7\%) & 1976.25 (2.0\%) & 1317.50 (1.3\%) & 658.75 (0.7\%) & 0.00 (0.0\%) \\
el & 3498.77 (3.5\%) & 3132.17 (3.1\%) & 2784.15 (2.8\%) & 2436.13 (2.4\%) & 2099.26 (2.1\%) & 1740.09 (1.7\%) & 1336.39 (1.4\%) & 1049.63 (1.0\%) & 699.75 (0.7\%) & 349.88 (0.3\%) & 0.00 (0.0\%) \\
en & 0.00 (0.0\%) & 10002.43 (10.0\%) & 20004.86 (20.0\%) & 30007.30 (30.0\%) & 40009.73 (40.0\%) & 50012.16 (50.0\%) & 57614.01 (60.0\%) & 70017.02 (70.0\%) & 80019.46 (80.0\%) & 90021.89 (90.0\%) & 100024.32 (100.0\%) \\
es & 7044.66 (7.0\%) & 6275.04 (6.3\%) & 5577.81 (5.6\%) & 4880.58 (4.9\%) & 4226.80 (4.2\%) & 3486.13 (3.5\%) & 2677.35 (2.8\%) & 2113.40 (2.1\%) & 1408.93 (1.4\%) & 704.47 (0.7\%) & 0.00 (0.0\%) \\
et & 2009.65 (2.0\%) & 1811.96 (1.8\%) & 1610.63 (1.6\%) & 1409.30 (1.4\%) & 1205.79 (1.2\%) & 1006.64 (1.0\%) & 773.10 (0.8\%) & 602.90 (0.6\%) & 401.93 (0.4\%) & 200.97 (0.2\%) & 0.00 (0.0\%) \\
eu & 1239.38 (1.2\%) & 1163.61 (1.2\%) & 1034.32 (1.0\%) & 905.03 (0.9\%) & 743.63 (0.7\%) & 646.45 (0.6\%) & 496.47 (0.5\%) & 371.82 (0.4\%) & 247.88 (0.2\%) & 123.94 (0.1\%) & 0.00 (0.0\%) \\
fa & 3706.17 (3.7\%) & 3380.02 (3.4\%) & 3004.46 (3.0\%) & 2628.91 (2.6\%) & 2223.70 (2.2\%) & 1877.79 (1.9\%) & 1442.14 (1.5\%) & 1111.85 (1.1\%) & 741.23 (0.7\%) & 370.62 (0.4\%) & 0.00 (0.0\%) \\
fi & 2968.54 (3.0\%) & 2739.67 (2.7\%) & 2435.26 (2.4\%) & 2130.85 (2.1\%) & 1781.12 (1.8\%) & 1522.04 (1.5\%) & 1168.93 (1.2\%) & 890.56 (0.9\%) & 593.71 (0.6\%) & 296.85 (0.3\%) & 0.00 (0.0\%) \\
fr & 6415.58 (6.4\%) & 5865.82 (5.9\%) & 5214.06 (5.2\%) & 4562.30 (4.6\%) & 3849.35 (3.8\%) & 3258.79 (3.3\%) & 2502.75 (2.6\%) & 1924.67 (1.9\%) & 1283.12 (1.3\%) & 641.56 (0.6\%) & 0.00 (0.0\%) \\
hi & 2932.04 (2.9\%) & 2462.93 (2.5\%) & 2189.27 (2.2\%) & 1915.61 (1.9\%) & 1759.23 (1.8\%) & 1368.30 (1.4\%) & 1050.85 (1.1\%) & 879.61 (0.9\%) & 586.41 (0.6\%) & 293.20 (0.3\%) & 0.00 (0.0\%) \\
ht & 687.25 (0.7\%) & 700.65 (0.7\%) & 622.80 (0.6\%) & 544.95 (0.5\%) & 412.35 (0.4\%) & 389.25 (0.4\%) & 298.95 (0.3\%) & 206.18 (0.2\%) & 137.45 (0.1\%) & 68.73 (0.1\%) & 0.00 (0.0\%) \\
id & 4037.86 (4.0\%) & 3656.56 (3.7\%) & 3250.27 (3.2\%) & 2843.99 (2.8\%) & 2422.72 (2.4\%) & 2031.42 (2.0\%) & 1560.13 (1.6\%) & 1211.36 (1.2\%) & 807.57 (0.8\%) & 403.79 (0.4\%) & 0.00 (0.0\%) \\
it & 5229.67 (5.2\%) & 4916.80 (4.9\%) & 4370.49 (4.4\%) & 3824.18 (3.8\%) & 3137.80 (3.1\%) & 2731.56 (2.7\%) & 2097.84 (2.2\%) & 1568.90 (1.6\%) & 1045.93 (1.0\%) & 522.97 (0.5\%) & 0.00 (0.0\%) \\
ja & 5249.15 (5.2\%) & 3905.57 (3.9\%) & 3471.62 (3.5\%) & 3037.67 (3.0\%) & 3149.49 (3.1\%) & 2169.76 (2.2\%) & 1666.38 (1.7\%) & 1574.75 (1.6\%) & 1049.83 (1.0\%) & 524.92 (0.5\%) & 0.00 (0.0\%) \\
ko & 3004.03 (3.0\%) & 2337.96 (2.3\%) & 2078.18 (2.1\%) & 1818.41 (1.8\%) & 1802.42 (1.8\%) & 1298.86 (1.3\%) & 997.53 (1.0\%) & 901.21 (0.9\%) & 600.81 (0.6\%) & 300.40 (0.3\%) & 0.00 (0.0\%) \\
my & 1084.07 (1.1\%) & 943.16 (0.9\%) & 838.36 (0.8\%) & 733.57 (0.7\%) & 650.44 (0.7\%) & 523.98 (0.5\%) & 402.41 (0.4\%) & 325.22 (0.3\%) & 216.81 (0.2\%) & 108.41 (0.1\%) & 0.00 (0.0\%) \\
pt & 5067.45 (5.1\%) & 4776.05 (4.8\%) & 4245.38 (4.2\%) & 3714.70 (3.7\%) & 3040.47 (3.0\%) & 2653.36 (2.7\%) & 2037.78 (2.1\%) & 1520.23 (1.5\%) & 1013.49 (1.0\%) & 506.74 (0.5\%) & 0.00 (0.0\%) \\
ru & 8194.02 (8.2\%) & 7520.23 (7.5\%) & 6684.65 (6.7\%) & 5849.07 (5.8\%) & 4916.41 (4.9\%) & 4177.90 (4.2\%) & 3208.63 (3.3\%) & 2458.21 (2.5\%) & 1638.80 (1.6\%) & 819.40 (0.8\%) & 0.00 (0.0\%) \\
sw & 1119.24 (1.1\%) & 1009.14 (1.0\%) & 897.01 (0.9\%) & 784.89 (0.8\%) & 671.55 (0.7\%) & 560.63 (0.6\%) & 430.57 (0.4\%) & 335.77 (0.3\%) & 223.85 (0.2\%) & 111.92 (0.1\%) & 0.00 (0.0\%) \\
ta & 1621.73 (1.6\%) & 1475.10 (1.5\%) & 1311.20 (1.3\%) & 1147.30 (1.1\%) & 973.04 (1.0\%) & 819.50 (0.8\%) & 629.37 (0.7\%) & 486.52 (0.5\%) & 324.35 (0.3\%) & 162.17 (0.2\%) & 0.00 (0.0\%) \\
te & 1211.86 (1.2\%) & 1066.46 (1.1\%) & 947.97 (0.9\%) & 829.47 (0.8\%) & 727.12 (0.7\%) & 592.48 (0.6\%) & 455.02 (0.5\%) & 363.56 (0.4\%) & 242.37 (0.2\%) & 121.19 (0.1\%) & 0.00 (0.0\%) \\
th & 2314.72 (2.3\%) & 2292.68 (2.3\%) & 2037.94 (2.0\%) & 1783.19 (1.8\%) & 1388.83 (1.4\%) & 1273.71 (1.3\%) & 978.21 (1.0\%) & 694.42 (0.7\%) & 462.94 (0.5\%) & 231.47 (0.2\%) & 0.00 (0.0\%) \\
tr & 4072.98 (4.1\%) & 3919.12 (3.9\%) & 3483.66 (3.5\%) & 3048.21 (3.0\%) & 2443.79 (2.4\%) & 2177.29 (2.2\%) & 1672.16 (1.7\%) & 1221.89 (1.2\%) & 814.60 (0.8\%) & 407.30 (0.4\%) & 0.00 (0.0\%) \\
ur & 1459.28 (1.5\%) & 1225.81 (1.2\%) & 1089.61 (1.1\%) & 953.41 (1.0\%) & 875.57 (0.9\%) & 681.00 (0.7\%) & 523.01 (0.5\%) & 437.79 (0.4\%) & 291.86 (0.3\%) & 145.93 (0.1\%) & 0.00 (0.0\%) \\
vi & 4726.26 (4.7\%) & 3793.06 (3.8\%) & 3371.61 (3.4\%) & 2950.16 (2.9\%) & 2835.76 (2.8\%) & 2107.26 (2.1\%) & 1618.37 (1.7\%) & 1417.88 (1.4\%) & 945.25 (0.9\%) & 472.63 (0.5\%) & 0.00 (0.0\%) \\
zh & 3396.76 (3.4\%) & 3398.87 (3.4\%) & 3021.22 (3.0\%) & 2643.56 (2.6\%) & 2038.06 (2.0\%) & 1888.26 (1.9\%) & 1450.18 (1.5\%) & 1019.03 (1.0\%) & 679.35 (0.7\%) & 339.68 (0.3\%) & 0.00 (0.0\%) \\
    \bottomrule
    \end{tabular}
    }
    \caption{Token counts (in millions) and their total proportions (\%) for the \textit{Fixed Total Budget} experiments described in Section~\ref{sec:english}. Total number of tokens is 100B.}
    \label{tab:token_distribution_en_xglm_fixed_total}
\end{table}
\begin{table}[ht]
    \centering
    \scriptsize
    \setlength{\tabcolsep}{4pt}

    \resizebox{\textwidth}{!}{
    \begin{tabular}{lccccccccccc}
    \toprule
    Language & en=20\% & en=30\% & en=40\% & en=50\% & en=60\% \\
    \midrule
    ar & 3345.99 (3.0\%) & 3345.99 (2.6\%) & 3316.12 (2.2\%) & 3286.24 (1.9\%) & 3345.99 (1.5\%) \\
bg & 2598.14 (2.3\%) & 2598.14 (2.0\%) & 2574.94 (1.7\%) & 2551.74 (1.4\%) & 2598.14 (1.2\%) \\
bn & 1842.56 (1.6\%) & 1842.56 (1.4\%) & 1826.10 (1.2\%) & 1809.65 (1.0\%) & 1842.56 (0.8\%) \\
ca & 2235.26 (2.0\%) & 2235.26 (1.7\%) & 2215.31 (1.5\%) & 2195.35 (1.2\%) & 2235.26 (1.0\%) \\
de & 6159.27 (5.5\%) & 6159.27 (4.8\%) & 6104.28 (4.1\%) & 6049.28 (3.4\%) & 6159.27 (2.7\%) \\
el & 3118.25 (2.8\%) & 3118.25 (2.4\%) & 3090.41 (2.1\%) & 3062.57 (1.7\%) & 3118.25 (1.4\%) \\
en & 22405.45 (20.0\%) & 38409.34 (30.0\%) & 59214.40 (40.0\%) & 88021.40 (50.0\%) & 134432.69 (60.0\%) \\
es & 6247.15 (5.6\%) & 6247.15 (4.9\%) & 6191.37 (4.2\%) & 6135.59 (3.5\%) & 6247.15 (2.8\%) \\
et & 1803.91 (1.6\%) & 1803.91 (1.4\%) & 1787.80 (1.2\%) & 1771.69 (1.0\%) & 1803.91 (0.8\%) \\
eu & 1158.43 (1.0\%) & 1158.43 (0.9\%) & 1148.09 (0.8\%) & 1137.75 (0.6\%) & 1158.43 (0.5\%) \\
fa & 3365.00 (3.0\%) & 3365.00 (2.6\%) & 3334.95 (2.3\%) & 3304.91 (1.9\%) & 3365.00 (1.5\%) \\
fi & 2727.49 (2.4\%) & 2727.49 (2.1\%) & 2703.14 (1.8\%) & 2678.79 (1.5\%) & 2727.49 (1.2\%) \\
fr & 5839.75 (5.2\%) & 5839.75 (4.6\%) & 5787.61 (3.9\%) & 5735.47 (3.3\%) & 5839.75 (2.6\%) \\
hi & 2451.99 (2.2\%) & 2451.99 (1.9\%) & 2430.09 (1.6\%) & 2408.20 (1.4\%) & 2451.99 (1.1\%) \\
ht & 697.54 (0.6\%) & 697.54 (0.5\%) & 691.31 (0.5\%) & 685.08 (0.4\%) & 697.54 (0.3\%) \\
id & 3640.30 (3.2\%) & 3640.30 (2.8\%) & 3607.80 (2.4\%) & 3575.30 (2.0\%) & 3640.30 (1.6\%) \\
it & 4894.95 (4.4\%) & 4894.95 (3.8\%) & 4851.25 (3.3\%) & 4807.54 (2.7\%) & 4894.95 (2.2\%) \\
ja & 3888.21 (3.5\%) & 3888.21 (3.0\%) & 3853.50 (2.6\%) & 3818.78 (2.2\%) & 3888.21 (1.7\%) \\
ko & 2327.57 (2.1\%) & 2327.57 (1.8\%) & 2306.78 (1.6\%) & 2286.00 (1.3\%) & 2327.57 (1.0\%) \\
my & 938.97 (0.8\%) & 938.97 (0.7\%) & 930.58 (0.6\%) & 922.20 (0.5\%) & 938.97 (0.4\%) \\
pt & 4754.82 (4.2\%) & 4754.82 (3.7\%) & 4712.37 (3.2\%) & 4669.91 (2.7\%) & 4754.82 (2.1\%) \\
ru & 7486.80 (6.7\%) & 7486.80 (5.8\%) & 7419.96 (5.0\%) & 7353.11 (4.2\%) & 7486.80 (3.3\%) \\
sw & 1004.66 (0.9\%) & 1004.66 (0.8\%) & 995.69 (0.7\%) & 986.72 (0.6\%) & 1004.66 (0.4\%) \\
ta & 1468.54 (1.3\%) & 1468.54 (1.1\%) & 1455.43 (1.0\%) & 1442.32 (0.8\%) & 1468.54 (0.7\%) \\
te & 1061.72 (0.9\%) & 1061.72 (0.8\%) & 1052.24 (0.7\%) & 1042.76 (0.6\%) & 1061.72 (0.5\%) \\
th & 2282.49 (2.0\%) & 2282.49 (1.8\%) & 2262.11 (1.5\%) & 2241.73 (1.3\%) & 2282.49 (1.0\%) \\
tr & 3901.70 (3.5\%) & 3901.70 (3.0\%) & 3866.87 (2.6\%) & 3832.03 (2.2\%) & 3901.70 (1.7\%) \\
ur & 1220.36 (1.1\%) & 1220.36 (1.0\%) & 1209.46 (0.8\%) & 1198.57 (0.7\%) & 1220.36 (0.5\%) \\
vi & 3776.21 (3.4\%) & 3776.21 (2.9\%) & 3742.49 (2.5\%) & 3708.77 (2.1\%) & 3776.21 (1.7\%) \\
zh & 3383.76 (3.0\%) & 3383.76 (2.6\%) & 3353.55 (2.3\%) & 3323.34 (1.9\%) & 3383.76 (1.5\%) \\
    \midrule
    \textbf{Total} & 112027.20 (100.0\%) & 128031.99 (100.0\%) & 148037.75 (100.0\%) & 176043.52 (100.0\%) & 224054.06 (100.0\%) \\
    \bottomrule
    \end{tabular}
    }
    \caption{Token counts (in millions) and their total proportions (\%) for the \textit{Fixed Multilingual Budget} experiments described in Section~\ref{sec:english}.}
    \label{tab:token_distribution_en_xglm_fixed_multilingual}
\end{table}

\begin{table}[ht]
        \centering
        \scriptsize
        \setlength{\tabcolsep}{4pt}
        \resizebox{\textwidth}{!}{
        \begin{tabular}{lccccccccccc}
        \toprule
        Language & en=00\% & en=10\% & en=20\% & en=30\% & en=40\% & en=50\% & en=60\% & en=70\% & en=80\% & en=90\% & en=100\% \\
        \midrule
        be & 3733.16 (3.7\%) & 3359.85 (3.4\%) & 2986.53 (3.0\%) & 2613.21 (2.6\%) & 2239.90 (2.2\%) & 1866.58 (1.9\%) & 1493.27 (1.5\%) & 1119.95 (1.1\%) & 746.63 (0.7\%) & 373.32 (0.4\%) & 0.00 (0.0\%) \\
bg & 7720.59 (7.7\%) & 6948.53 (6.9\%) & 6176.47 (6.2\%) & 5404.41 (5.4\%) & 4632.36 (4.6\%) & 3860.30 (3.9\%) & 3088.24 (3.1\%) & 2316.18 (2.3\%) & 1544.12 (1.5\%) & 772.06 (0.8\%) & 0.00 (0.0\%) \\
cs & 10619.66 (10.6\%) & 9557.69 (9.6\%) & 8495.73 (8.5\%) & 7433.76 (7.4\%) & 6371.80 (6.4\%) & 5309.83 (5.3\%) & 4247.86 (4.2\%) & 3185.90 (3.2\%) & 2123.93 (2.1\%) & 1061.97 (1.1\%) & 0.00 (0.0\%) \\
en & 0.00 (0.0\%) & 10002.43 (10.0\%) & 20004.86 (20.0\%) & 30007.30 (30.0\%) & 40009.73 (40.0\%) & 50012.16 (50.0\%) & 60014.59 (60.0\%) & 70017.02 (70.0\%) & 80019.46 (80.0\%) & 90021.89 (90.0\%) & 100024.32 (100.0\%) \\
kk & 4263.37 (4.3\%) & 3837.04 (3.8\%) & 3410.70 (3.4\%) & 2984.36 (3.0\%) & 2558.02 (2.6\%) & 2131.69 (2.1\%) & 1705.35 (1.7\%) & 1279.01 (1.3\%) & 852.67 (0.9\%) & 426.34 (0.4\%) & 0.00 (0.0\%) \\
ky & 3025.91 (3.0\%) & 2723.32 (2.7\%) & 2420.73 (2.4\%) & 2118.14 (2.1\%) & 1815.55 (1.8\%) & 1512.95 (1.5\%) & 1210.36 (1.2\%) & 907.77 (0.9\%) & 605.18 (0.6\%) & 302.59 (0.3\%) & 0.00 (0.0\%) \\
mk & 3615.86 (3.6\%) & 3254.27 (3.3\%) & 2892.69 (2.9\%) & 2531.10 (2.5\%) & 2169.51 (2.2\%) & 1807.93 (1.8\%) & 1446.34 (1.4\%) & 1084.76 (1.1\%) & 723.17 (0.7\%) & 361.59 (0.4\%) & 0.00 (0.0\%) \\
mn & 4088.58 (4.1\%) & 3679.72 (3.7\%) & 3270.86 (3.3\%) & 2862.00 (2.9\%) & 2453.15 (2.5\%) & 2044.29 (2.0\%) & 1635.43 (1.6\%) & 1226.57 (1.2\%) & 817.72 (0.8\%) & 408.86 (0.4\%) & 0.00 (0.0\%) \\
pl & 13226.50 (13.2\%) & 11903.85 (11.9\%) & 10581.20 (10.6\%) & 9258.55 (9.3\%) & 7935.90 (7.9\%) & 6613.25 (6.6\%) & 5290.60 (5.3\%) & 3967.95 (4.0\%) & 2645.30 (2.6\%) & 1322.65 (1.3\%) & 0.00 (0.0\%) \\
ru & 22152.79 (22.1\%) & 19937.51 (19.9\%) & 17722.23 (17.7\%) & 15506.95 (15.5\%) & 13291.68 (13.3\%) & 11076.40 (11.1\%) & 8861.12 (8.9\%) & 6645.84 (6.6\%) & 4430.56 (4.4\%) & 2215.28 (2.2\%) & 0.00 (0.0\%) \\
sk & 7265.10 (7.3\%) & 6538.59 (6.5\%) & 5812.08 (5.8\%) & 5085.57 (5.1\%) & 4359.06 (4.4\%) & 3632.55 (3.6\%) & 2906.04 (2.9\%) & 2179.53 (2.2\%) & 1453.02 (1.5\%) & 726.51 (0.7\%) & 0.00 (0.0\%) \\
sr & 4707.78 (4.7\%) & 4237.00 (4.2\%) & 3766.22 (3.8\%) & 3295.44 (3.3\%) & 2824.67 (2.8\%) & 2353.89 (2.4\%) & 1883.11 (1.9\%) & 1412.33 (1.4\%) & 941.56 (0.9\%) & 470.78 (0.5\%) & 0.00 (0.0\%) \\
tg & 3350.71 (3.3\%) & 3015.64 (3.0\%) & 2680.57 (2.7\%) & 2345.50 (2.3\%) & 2010.43 (2.0\%) & 1675.36 (1.7\%) & 1340.28 (1.3\%) & 1005.21 (1.0\%) & 670.14 (0.7\%) & 335.07 (0.3\%) & 0.00 (0.0\%) \\
uk & 9323.48 (9.3\%) & 8391.14 (8.4\%) & 7458.79 (7.5\%) & 6526.44 (6.5\%) & 5594.09 (5.6\%) & 4661.74 (4.7\%) & 3729.39 (3.7\%) & 2797.05 (2.8\%) & 1864.70 (1.9\%) & 932.35 (0.9\%) & 0.00 (0.0\%) \\
uz & 2930.83 (2.9\%) & 2637.74 (2.6\%) & 2344.66 (2.3\%) & 2051.58 (2.1\%) & 1758.50 (1.8\%) & 1465.41 (1.5\%) & 1172.33 (1.2\%) & 879.25 (0.9\%) & 586.17 (0.6\%) & 293.08 (0.3\%) & 0.00 (0.0\%) \\
        \bottomrule
        \end{tabular}
        }
        \caption{Token counts (in millions) and their total proportions (\%) for the \textit{English} as pivot runs described in Section~\ref{sec:family}. Total number of tokens is 100B.}
    \end{table}
\begin{table}[ht]
        \centering
        \scriptsize
        \setlength{\tabcolsep}{4pt}
        \resizebox{\textwidth}{!}{
        \begin{tabular}{lccccccccccc}
        \toprule
        Language & ru=00\% & ru=10\% & ru=20\% & ru=30\% & ru=40\% & ru=50\% & ru=60\% & ru=70\% & ru=80\% & ru=90\% & ru=100\% \\
        \midrule
        be & 3359.26 (3.4\%) & 3023.33 (3.0\%) & 2687.41 (2.7\%) & 2351.48 (2.4\%) & 2015.56 (2.0\%) & 1679.63 (1.7\%) & 1343.70 (1.3\%) & 1007.78 (1.0\%) & 671.85 (0.7\%) & 335.93 (0.3\%) & 0.00 (0.0\%) \\
bg & 6947.32 (6.9\%) & 6252.59 (6.3\%) & 5557.86 (5.6\%) & 4863.12 (4.9\%) & 4168.39 (4.2\%) & 3473.66 (3.5\%) & 2778.93 (2.8\%) & 2084.20 (2.1\%) & 1389.46 (1.4\%) & 694.73 (0.7\%) & 0.00 (0.0\%) \\
cs & 9556.02 (9.6\%) & 8600.42 (8.6\%) & 7644.82 (7.6\%) & 6689.22 (6.7\%) & 5733.61 (5.7\%) & 4778.01 (4.8\%) & 3822.41 (3.8\%) & 2866.81 (2.9\%) & 1911.20 (1.9\%) & 955.60 (1.0\%) & 0.00 (0.0\%) \\
en & 29952.18 (29.9\%) & 26956.97 (27.0\%) & 23961.75 (24.0\%) & 20966.53 (21.0\%) & 17971.31 (18.0\%) & 14976.09 (15.0\%) & 11980.87 (12.0\%) & 8985.66 (9.0\%) & 5990.44 (6.0\%) & 2995.22 (3.0\%) & 0.00 (0.0\%) \\
kk & 3836.37 (3.8\%) & 3452.73 (3.5\%) & 3069.09 (3.1\%) & 2685.46 (2.7\%) & 2301.82 (2.3\%) & 1918.18 (1.9\%) & 1534.55 (1.5\%) & 1150.91 (1.2\%) & 767.27 (0.8\%) & 383.64 (0.4\%) & 0.00 (0.0\%) \\
ky & 2722.84 (2.7\%) & 2450.56 (2.4\%) & 2178.27 (2.2\%) & 1905.99 (1.9\%) & 1633.71 (1.6\%) & 1361.42 (1.4\%) & 1089.14 (1.1\%) & 816.85 (0.8\%) & 544.57 (0.5\%) & 272.28 (0.3\%) & 0.00 (0.0\%) \\
mk & 3253.70 (3.3\%) & 2928.33 (2.9\%) & 2602.96 (2.6\%) & 2277.59 (2.3\%) & 1952.22 (2.0\%) & 1626.85 (1.6\%) & 1301.48 (1.3\%) & 976.11 (1.0\%) & 650.74 (0.7\%) & 325.37 (0.3\%) & 0.00 (0.0\%) \\
mn & 3679.08 (3.7\%) & 3311.17 (3.3\%) & 2943.26 (2.9\%) & 2575.35 (2.6\%) & 2207.45 (2.2\%) & 1839.54 (1.8\%) & 1471.63 (1.5\%) & 1103.72 (1.1\%) & 735.82 (0.7\%) & 367.91 (0.4\%) & 0.00 (0.0\%) \\
pl & 11901.77 (11.9\%) & 10711.59 (10.7\%) & 9521.41 (9.5\%) & 8331.24 (8.3\%) & 7141.06 (7.1\%) & 5950.88 (5.9\%) & 4760.71 (4.8\%) & 3570.53 (3.6\%) & 2380.35 (2.4\%) & 1190.18 (1.2\%) & 0.00 (0.0\%) \\
ru & 0.00 (0.0\%) & 10002.43 (10.0\%) & 20004.86 (20.0\%) & 30007.30 (30.0\%) & 40009.73 (40.0\%) & 50012.16 (50.0\%) & 60014.59 (60.0\%) & 70017.02 (70.0\%) & 80019.46 (80.0\%) & 90021.89 (90.0\%) & 100024.32 (100.0\%) \\
sk & 6537.45 (6.5\%) & 5883.70 (5.9\%) & 5229.96 (5.2\%) & 4576.21 (4.6\%) & 3922.47 (3.9\%) & 3268.72 (3.3\%) & 2614.98 (2.6\%) & 1961.23 (2.0\%) & 1307.49 (1.3\%) & 653.74 (0.7\%) & 0.00 (0.0\%) \\
sr & 4236.26 (4.2\%) & 3812.63 (3.8\%) & 3389.01 (3.4\%) & 2965.38 (3.0\%) & 2541.76 (2.5\%) & 2118.13 (2.1\%) & 1694.50 (1.7\%) & 1270.88 (1.3\%) & 847.25 (0.8\%) & 423.63 (0.4\%) & 0.00 (0.0\%) \\
tg & 3015.11 (3.0\%) & 2713.60 (2.7\%) & 2412.09 (2.4\%) & 2110.58 (2.1\%) & 1809.07 (1.8\%) & 1507.56 (1.5\%) & 1206.05 (1.2\%) & 904.53 (0.9\%) & 603.02 (0.6\%) & 301.51 (0.3\%) & 0.00 (0.0\%) \\
uk & 8389.67 (8.4\%) & 7550.70 (7.5\%) & 6711.74 (6.7\%) & 5872.77 (5.9\%) & 5033.80 (5.0\%) & 4194.84 (4.2\%) & 3355.87 (3.4\%) & 2516.90 (2.5\%) & 1677.93 (1.7\%) & 838.97 (0.8\%) & 0.00 (0.0\%) \\
uz & 2637.28 (2.6\%) & 2373.55 (2.4\%) & 2109.83 (2.1\%) & 1846.10 (1.8\%) & 1582.37 (1.6\%) & 1318.64 (1.3\%) & 1054.91 (1.1\%) & 791.18 (0.8\%) & 527.46 (0.5\%) & 263.73 (0.3\%) & 0.00 (0.0\%) \\
        \bottomrule
        \end{tabular}
        }
        \caption{Token counts (in millions) and their total proportions (\%) for the \textit{Russian} as pivot runs described in Section~\ref{sec:family}. Total number of tokens is 100B.}
    \end{table}
\begin{table}
        \centering
        \scriptsize
        \setlength{\tabcolsep}{4pt}
        \resizebox{\textwidth}{!}{
        \begin{tabular}{lccccccccccc}
        \toprule
        Language & en=00\%,ru=00\% & en=05\%,ru=05\% & en=10\%,ru=10\% & en=15\%,ru=15\% & en=20\%,ru=20\% & en=25\%,ru=25\% & en=30\%,ru=30\% & en=35\%,ru=35\% & en=40\%,ru=40\% & en=45\%,ru=45\% & en=50\%,ru=50\% \\
        \midrule
        be & 4795.17 (4.8\%) & 4315.65 (4.3\%) & 3836.14 (3.8\%) & 3356.62 (3.4\%) & 2877.10 (2.9\%) & 2301.68 (2.4\%) & 1918.07 (1.9\%) & 1438.55 (1.4\%) & 920.67 (1.0\%) & 479.52 (0.5\%) & 0.00 (0.0\%) \\
bg & 9916.94 (9.9\%) & 8925.24 (8.9\%) & 7933.55 (7.9\%) & 6941.86 (6.9\%) & 5950.16 (5.9\%) & 4760.13 (5.0\%) & 3966.77 (4.0\%) & 2975.08 (3.0\%) & 1904.05 (2.0\%) & 991.69 (1.0\%) & 0.00 (0.0\%) \\
cs & 13640.73 (13.6\%) & 12276.65 (12.3\%) & 10912.58 (10.9\%) & 9548.51 (9.5\%) & 8184.44 (8.2\%) & 6547.55 (6.8\%) & 5456.29 (5.5\%) & 4092.22 (4.1\%) & 2619.02 (2.7\%) & 1364.07 (1.4\%) & 0.00 (0.0\%) \\
en & 0.00 (0.0\%) & 5001.22 (5.0\%) & 10002.43 (10.0\%) & 15003.65 (15.0\%) & 20004.86 (20.0\%) & 24005.84 (25.0\%) & 30007.30 (30.0\%) & 35008.51 (35.0\%) & 38409.34 (40.0\%) & 45010.94 (45.0\%) & 50012.16 (50.0\%) \\
kk & 5476.21 (5.5\%) & 4928.59 (4.9\%) & 4380.97 (4.4\%) & 3833.35 (3.8\%) & 3285.73 (3.3\%) & 2628.58 (2.7\%) & 2190.48 (2.2\%) & 1642.86 (1.6\%) & 1051.43 (1.1\%) & 547.62 (0.5\%) & 0.00 (0.0\%) \\
ky & 3886.72 (3.9\%) & 3498.04 (3.5\%) & 3109.37 (3.1\%) & 2720.70 (2.7\%) & 2332.03 (2.3\%) & 1865.62 (1.9\%) & 1554.69 (1.6\%) & 1166.01 (1.2\%) & 746.25 (0.8\%) & 388.67 (0.4\%) & 0.00 (0.0\%) \\
mk & 4644.49 (4.6\%) & 4180.04 (4.2\%) & 3715.59 (3.7\%) & 3251.14 (3.3\%) & 2786.69 (2.8\%) & 2229.36 (2.3\%) & 1857.80 (1.9\%) & 1393.35 (1.4\%) & 891.74 (0.9\%) & 464.45 (0.5\%) & 0.00 (0.0\%) \\
mn & 5251.69 (5.3\%) & 4726.52 (4.7\%) & 4201.35 (4.2\%) & 3676.18 (3.7\%) & 3151.01 (3.2\%) & 2520.81 (2.6\%) & 2100.68 (2.1\%) & 1575.51 (1.6\%) & 1008.32 (1.1\%) & 525.17 (0.5\%) & 0.00 (0.0\%) \\
pl & 16989.15 (17.0\%) & 15290.24 (15.3\%) & 13591.32 (13.6\%) & 11892.41 (11.9\%) & 10193.49 (10.2\%) & 8154.79 (8.5\%) & 6795.66 (6.8\%) & 5096.75 (5.1\%) & 3261.92 (3.4\%) & 1698.92 (1.7\%) & 0.00 (0.0\%) \\
ru & 0.00 (0.0\%) & 5001.22 (5.0\%) & 10002.43 (10.0\%) & 15003.65 (15.0\%) & 20004.86 (20.0\%) & 24005.84 (25.0\%) & 30007.30 (30.0\%) & 35008.51 (35.0\%) & 38409.34 (40.0\%) & 45010.94 (45.0\%) & 50012.16 (50.0\%) \\
sk & 9331.86 (9.3\%) & 8398.68 (8.4\%) & 7465.49 (7.5\%) & 6532.30 (6.5\%) & 5599.12 (5.6\%) & 4479.29 (4.7\%) & 3732.75 (3.7\%) & 2799.56 (2.8\%) & 1791.72 (1.9\%) & 933.19 (0.9\%) & 0.00 (0.0\%) \\
sr & 6047.04 (6.0\%) & 5442.34 (5.4\%) & 4837.63 (4.8\%) & 4232.93 (4.2\%) & 3628.22 (3.6\%) & 2902.58 (3.0\%) & 2418.82 (2.4\%) & 1814.11 (1.8\%) & 1161.03 (1.2\%) & 604.70 (0.6\%) & 0.00 (0.0\%) \\
tg & 4303.92 (4.3\%) & 3873.53 (3.9\%) & 3443.13 (3.4\%) & 3012.74 (3.0\%) & 2582.35 (2.6\%) & 2065.88 (2.2\%) & 1721.57 (1.7\%) & 1291.18 (1.3\%) & 826.35 (0.9\%) & 430.39 (0.4\%) & 0.00 (0.0\%) \\
uk & 11975.82 (12.0\%) & 10778.24 (10.8\%) & 9580.65 (9.6\%) & 8383.07 (8.4\%) & 7185.49 (7.2\%) & 5748.39 (6.0\%) & 4790.33 (4.8\%) & 3592.75 (3.6\%) & 2299.36 (2.4\%) & 1197.58 (1.2\%) & 0.00 (0.0\%) \\
uz & 3764.58 (3.8\%) & 3388.12 (3.4\%) & 3011.67 (3.0\%) & 2635.21 (2.6\%) & 2258.75 (2.3\%) & 1807.00 (1.9\%) & 1505.83 (1.5\%) & 1129.37 (1.1\%) & 722.80 (0.8\%) & 376.46 (0.4\%) & 0.00 (0.0\%) \\
        \bottomrule
        \end{tabular}
        }
        \caption{Token counts (in millions) and their total proportions (\%) for the \textit{English and Russian} as pivots runs described in Section~\ref{sec:family}. Total number of tokens is 100B.}
    \end{table}

\end{document}